\documentclass{article}

\usepackage{iclr2020}

\usepackage{graphicx}
\usepackage{subcaption}
\usepackage{float}
\usepackage[justification=raggedright]{caption}	% makes captions ragged right - thanks to Bryce Lobdell
\usepackage{lscape}                                         % Useful for wide tables or figures.
\usepackage{wrapfig}

\usepackage[lined,ruled,linesnumbered]{algorithm2e}
\usepackage{animate} %% convert frames to video

\usepackage{booktabs}                   % Publication quality tables
\usepackage{multirow}
\usepackage{color,soul}
\usepackage{tabu,tabularx,,siunitx}

\usepackage{makecell}

\usepackage{paralist}
\usepackage{enumitem}

\usepackage{bm}                          % Make bold, italic math symbols
\usepackage{epsfig}                      % for figures
\usepackage{graphicx}                  % another package that works for figures
\usepackage{times}
\usepackage{mathptmx}
\usepackage{mathtools}
\usepackage{amssymb,amsmath}   % Short math guide for LaTeX ftp://ftp.ams.org/pub/tex/doc/amsmath/short-math-guide.pdf

\usepackage{units}
\usepackage{color}
\usepackage[T1]{fontenc}    % use 8-bit T1 fonts
\usepackage{amsfonts}       % blackboard math symbols
\usepackage[utf8]{inputenc} % allow utf-8 input
\usepackage{nicefrac}       % compact symbols for 1/2, etc.
\usepackage{microtype}      % microtypography
\usepackage{comment}

\usepackage{url}  % Hyphenation of URLs.

\usepackage[pagebackref=false,breaklinks=true,letterpaper=true,colorlinks,bookmarks=false,citecolor=blue]{hyperref}
\usepackage{xspace}
\usepackage{grfext}
\PrependGraphicsExtensions*{.jpg,.png,.PNG}
\usepackage{paralist}
\def\ie{i.e.,~}               % that is, in other words
\newlength\paramargin
\newlength\figmargin

\newlength\secmargin
\newlength\figcapmargin
\newlength\tabcapmargin

\newcommand{\mpage}[2]
{
\begin{minipage}{#1\linewidth}\centering
#2
\end{minipage}
}

\newcommand{\topic}[1]
{
\vspace{1.5mm}\noindent\textbf{#1}
}

\newcommand{\figref}[1]{Figure~\ref{fig:#1}}

\long\def\ignorethis#1{}

\newcommand{\tb}[1]{\textbf{#1}}

\newbox\jsavebox%

\def\xi{\mathbf{x}_i}

\graphicspath{{figure}, {example}}

\usepackage{amsmath}
\usepackage{mathtools}
\usepackage{wrapfig}
\usepackage{tikz}
\usepackage{pgfplots}

\newcommand {\jiabin}[1]{{\color{cyan}\textbf{Jia-Bin: }#1}\normalfont}
\newcommand {\vincent}[1]{{\color{blue}\textbf{Vincent: }#1}\normalfont}
\newcommand {\abhi}[1]{{\color{red}{Abhishek: }#1}\normalfont}
\newcommand {\esther}[1]{{\color{pink}\textbf{Esther: }#1}\normalfont}
\newcommand {\jiarui}[1]{{\color{teal}\textbf{Jiarui: }#1}\normalfont}

\newcommand{\final}{0}
\ifthenelse{\equal{\final}{1}}{
\renewcommand {\jiabin}[1]{}
\renewcommand {\vincent}[1]{}
\renewcommand {\abhi}[1]{}
\renewcommand {\esther}[1]{}
\renewcommand {\jiarui}[1]{}
}

\title{Few-Shot Adaptation of Generative \\ Adversarial Networks}

\author{
{Esther Robb$^1$ \quad
Wen-Sheng Chu$^2$ \quad
Abhishek Kumar$^2$ \quad
Jia-Bin Huang$^1$ 
} \\
\\
$^1$Virginia Tech \quad\quad
$^2$Google Research
}

\begin{document}

\maketitle
\vspace{-1ex}

%%%%%%%%%% Abstract %%%%%%%%%%%%%%
\begin{abstract}
Generative Adversarial Networks (GANs) have shown remarkable performance in image synthesis tasks, but typically require a large number of training samples to achieve high-quality synthesis.
This paper proposes a simple and effective method, Few-Shot GAN (FSGAN), for adapting GANs in few-shot settings (less than 100 images). 
FSGAN repurposes component analysis techniques and learns to adapt the singular values of the pre-trained weights while freezing the corresponding singular vectors. 
This provides a highly expressive parameter space for adaptation while constraining changes to the pretrained weights. 
We validate our method in a challenging few-shot setting of 5-100 images in the target domain. 
We show that our method has significant visual quality gains compared with existing GAN adaptation methods. 
We report qualitative and quantitative results showing the effectiveness of our method. 
We additionally highlight a problem for few-shot synthesis in the standard quantitative metric used by data-efficient image synthesis works.
Code and additional results are available at \hyperlink{http://e-271.github.io/few-shot-gan}{http://e-271.github.io/few-shot-gan}.
\begin{comment}
We present Few-Shot GAN (FSGAN), a simple and effective method for adapting Generative Adversarial Networks (GANs) in a few-shot setting, {\em i.e.}, no more than 100 training images.
%Unlike conventional data-efficient GANs that exploit domain adaptation in the regime of 100-1000 images, 
FSGAN repurposes component analysis techniques and learns to adapt the expressive space of singular values computed from pre-trained weights while freezing the corresponding singular vectors. 
Compared to alternative GAN adaptation methods, FSGAN offers a significantly fewer amount of learnable parameters, and meanwhile generating diverse and high-quality images using as few as only 5-100 training images.
We validate FSGAN in challenging settings of both close-domain and far-domain adaptation,
and report extensive qualitative and quantitative results showing the effectiveness of FSGAN.
We additionally reveal a problem of using the standard FID as a few-shot metric, showing FID could be biased towards overfitting. 
\end{comment}
\end{abstract}

%%%%%%%%%% Introduction %%%%%%%%%%%%%%
\section{Introduction}
\label{sec:intro}
Recent years have witnessed rapid progress in Generative Adversarial Networks (GAN) ~\citep{goodfellow2014neurips} with improvements in architecture designs \citep{radford2015unsupervised,karras2018iclr,zhang2018self,karras2019style}, training techniques~\citep{salimans2016improved,karras2018iclr,miyato2018spectral}, and loss functions~\citep{arjovsky2017wasserstein,gulrajani2017improved}. 
Training these models, however, typically requires large, diverse datasets in a target visual domain. 
While there have been significant advancements in improving training stability~\citep{karras2018iclr,miyato2018spectral}, adversarial optimization remains challenging because the optimal solutions lie at saddle points rather than a minimum of a loss function \citep{yadav2017stabilizing}.
Additionally, GAN-based models may suffer from the inadequate generation of rare modes in the training data because they optimize a mode-seeking loss rather than the mode-covering loss of standard likelihood maximization \citep{poole2016improved}. 
These difficulties of training GANs become even more severe when the number of training examples is scarce. 
In the low-data regime (e.g., less than 1,000 samples), GANs frequently suffer from memorization or instability, leading to a lack of diversity or poor visual quality.

Several recent efforts have been devoted to improving the sample efficiency of GANs through transfer learning.
The most straightforward approaches are finetuning a pre-trained generator and discriminator on the samples in the target domain~\citep{wang2018eccv,mo2020freeze}.
When the number of training examples is severely limited, however, finetuning the network weights often leads to poor results, particularly when the source and target domains are distant. 
Instead of finetuning the entire network weights, the method in \citep{noguchi2019iccv} focuses on adapting batch norm statistics, constraining the optimization problem to a smaller set of parameters.
% However, this limits the flexibility for adapting to a new domain. 
The authors report that this method achieves better results using MLE-based optimization but fails for GAN-based optimization. 
Although their quality outperforms GAN-based methods in the low-shot setting, the images are blurry and lack details due to maximum likelihood optimization.
Invertible flow-based models have shown promising results in data-efficient adaptation~\citep{gambardella2019arxiv}, but require compute- and memory-intensive architectures with high-dimensional latent spaces.

\begin{figure}[th]
    \centering
    \includegraphics[width=\textwidth]{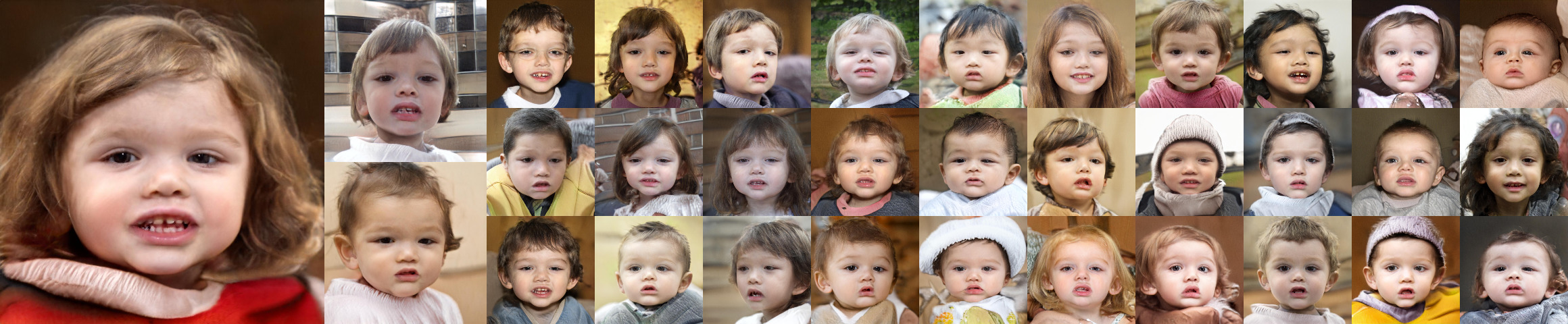}
    \includegraphics[width=\textwidth]{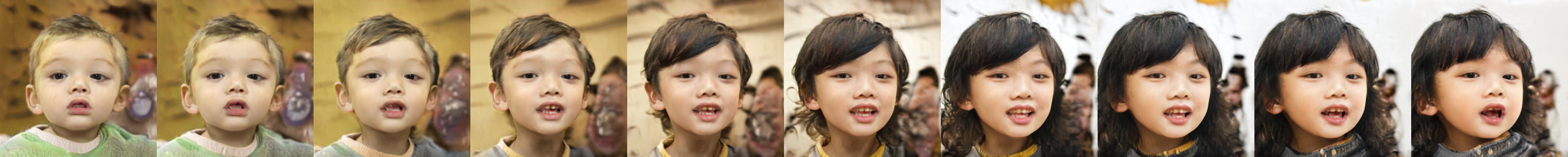}
    \caption{
    \tb{Few-shot image generation.} 
    Our method generates novel and high-quality samples in a new domain with a small amount of training data.
    (\emph{Top}) 
    Diverse random samples from adapting a FFHQ-pretrained StyleGAN2 to toddler images from the CelebA dataset (with {\bf only 30 images}) using our method. 
    % Random samples are generated by adapting on FFHQ-pretrained StyleGAN2 using our method {\bf only 30 images} of toddlers from CelebA.
    (\emph{Bottom}) Smooth latent space interpolation between two random seeds shows that our method produces novel samples instead of simply memorizing the 30 images. Please see the supplementary video for more results.
    }
    \label{fig:teaser}
\end{figure}

In this paper, we propose a method for adapting a pre-trained GAN to generate novel, high-quality sample images with a small number of training images from a new target domain (Figure \ref{fig:teaser}). 
To accomplish this, we restrict the space of trainable parameters to a small number of highly-expressive parameters that modulate orthogonal features of the pre-trained weight space. 
Our method first applies singular value decomposition (SVD) to the network weights of a pretrained GAN (generator + discriminator). 
We then adapts the singular values using GAN optimization on the target few-shot domain, with \emph{fixed} left/right singular vectors.
We show that varying singular values in the weight space corresponds to semantically meaningful changes of the synthesized image while preserving natural structure.
Compared with methods that finetune all weights of the GAN \citep{wang2018eccv}, individual layers \citep{mo2020freeze}, or only adapt batch norm statistics~\citep{noguchi2019iccv}, our method demonstrates higher image quality after adaptation.
We additionally highlight problems with the standard evaluation practice in the low-shot GAN setting. % and discuss alternatives.

%%%%%%%%%% Prior %%%%%%%%%%%%%%
\section{Background}
\label{sec:related}
\topic{Generative Adversarial Networks (GANs)}
GANs~\citep{goodfellow2014neurips} use adversarial training to learn a mapping of random noise to the distribution of an image dataset, allowing for sampling of novel images.
GANs optimize a competitive objective where a generator $G(Z)$ maximizes the classification error of a discriminator $D(X)$ trained to distinguish real data $p(X)$ from fake data $G(Z)$.
The GAN~\citep{goodfellow2014neurips} objective is expressed formally as:
\begin{align}
    \max_G \min_D \mathbb{E}_{x\sim p(X)}[\log\; D(x)] - \mathbb{E}_{x\sim G(X)}[1 - \log\; D(x)]
    \label{eqn:gan}
\end{align}
Recent research reformulated this objective to address instability problems \citep{arjovsky2017wasserstein,heusel2017gans,gulrajani2017improved}. 
Improved architecture and training has led to remarkable performance in synthesis \citep{karras2019analyzing,brock2018iclr}.
%and other downstream applications such as  representation learning \citep{chen2016infogan,donahue2019large} and conditional GANs~\citep{isola2017image,zhu2017cyclegan,wang2018high,lee2018diverse,huang2018multimodal,albahar2019guided,park2019semantic,hoffman2018cycada,chen2019crdoco}.
% and conditional GAN applications such as 
% image-to-image translation~\citep{isola2017image,zhu2017cyclegan,wang2018high,lee2018diverse,huang2018multimodal,albahar2019guided,park2019semantic},
% domain adaptation~\citep{hoffman2018cycada,chen2019crdoco},
% and image restoration~\citep{ledig2017photo,bau2020semantic,zhu2020domain}.
Compared to pixel-reconstruction losses \citep{kingma2013vae,higgins2017bvae,bojanowski2017optimizing} GANs typically produce sharper images, although strong priors over the latent space can offer competitive quality~\citep{razavi2019vqvae2}.
A high-quality generation has relied on large datasets of high-quality images (10K+) that may be expensive or infeasible to collect in many scenarios. 
Additionally, GANs can suffer from a lack of diversity, even when large training sets are used because the objective does not penalize the absence of outlier modes \citep{poole2016improved}.
Data-efficient GAN methods are, therefore, of great utility.
% These problems affect practical applications, so there is a need to improve diverse image synthesis in the low-data regime.

\begin{figure}
    \begin{minipage}[t]{0.57\textwidth}
        \small
        \begin{subfigure}{\textwidth}
            \begin{tabu} to \textwidth {X[l 2.0] X[c 2.4] X[c] X[c]} 
                \toprule
                \textbf{Method} & \textbf{Conv layer} & \tb{\#params} & \tb{Count} \\
                \midrule
                Pretrain & $conv(x,W_0)$ & -- & -- \\
                TGAN & $conv(x, {\color{red}{W}})$  & $k^2  c_\mathrm{in} c_\mathrm{out} $ &  59M\\
                FreezeD & $conv(x, {\color{red}{W}})$  & $k^2  c_\mathrm{in} c_\mathrm{out} $ &  47M \\
                SSGAN & $conv(x, W_0)\cdot {\color{red}{\gamma}} + {\color{red}{\beta}}$  & $2c_\mathrm{out}$ & 23K\\
                FSGAN (Ours) & $conv(x, W_{{\color{red}{\Sigma}}})$  & $c_\mathrm{out}$ & 16K\\
                \bottomrule
            \end{tabu}
            \caption{Adaptation method formulations.}
        \end{subfigure}
        \begin{subfigure}{\textwidth}
            \centering
            \includegraphics[width=0.89\textwidth]{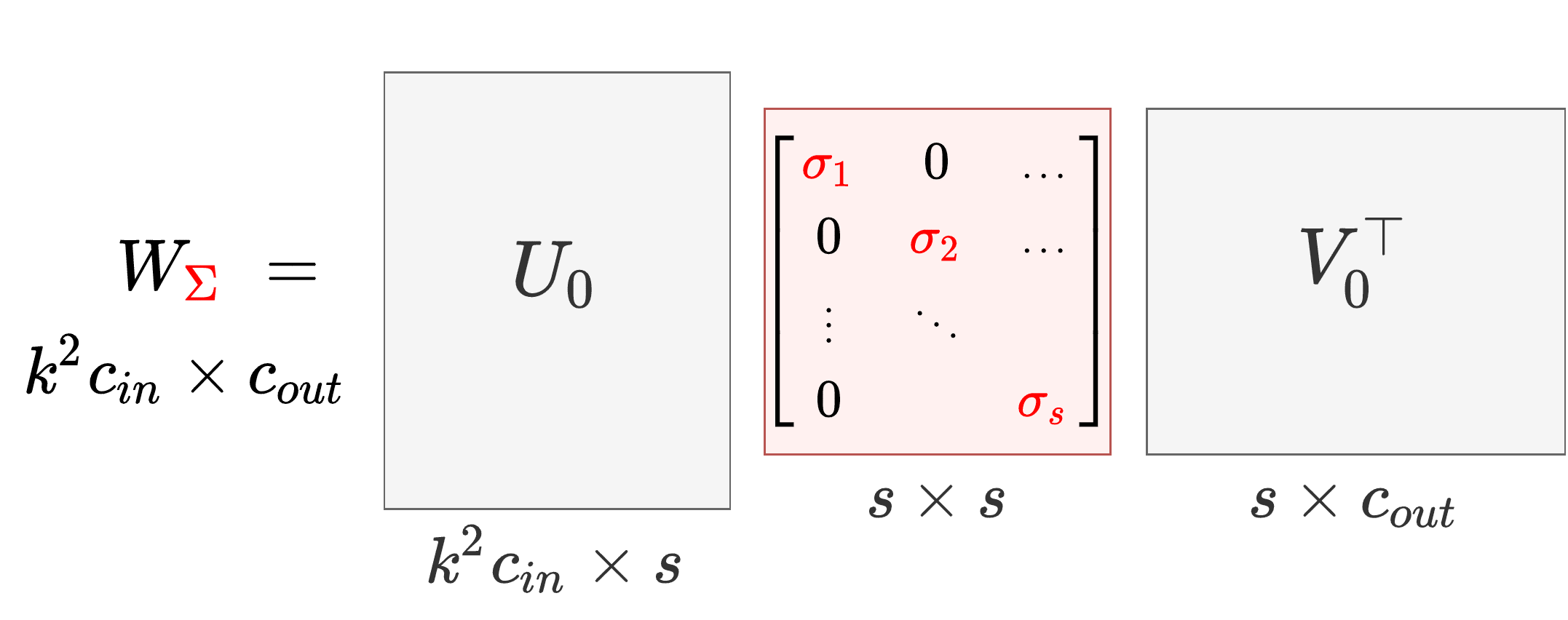}
            \vspace{-1.2ex}
            \caption{FSGAN singular value adaptation.}
        \end{subfigure}
    \end{minipage}
    \hspace{0.5em}
    \begin{minipage}[t]{0.42\textwidth}
        \vspace{-4.5em}
        \caption{
        {\bf Comparing methods for GAN adaption.}
        Learnable parameters are denoted in red. 
        \tb{(a)} TransferGAN (TGAN for simplicity) \citep{wang2018eccv} and FreezeD \citep{mo2020freeze} retrain all weights $W$ in a layer. 
        SSGAN \citep{noguchi2019iccv} and FSGAN train significantly fewer parameters per layer. 
        Note FSGAN adapts both conv and FC layers, while SSGAN adapts only conv layers. 
        \emph{\#params} is the number of learnable paramaters per conv layer; \emph{Count} gives parameter counts over the full StyleGAN2 generator and discriminator. 
        \tb{(b)} FSGAN (ours) adapts singular values $\Sigma=\{\sigma_1,\dots,\sigma_s\}$ of pretrained weights $W_0$ to obtain adapted weights $W_\Sigma$.
        }
        \label{fig:methods}
    \end{minipage}
\end{figure}

\topic{Sample-efficient Image Synthesis}
Sample-efficient image synthesis methods encourage diverse and high-quality generation in the low-data regime, most commonly through pretraining \citep{wang2018eccv,noguchi2019iccv} or simultaneous training \citep{yamaguchi2020aaai} on large image datasets.
The main differences among these methods lie in the choice of learnable parameters used for adaptation. 
Examples include adapting all weights of the generator and discriminator \citet{wang2018eccv}, freezing only lower layers of the discriminator \cite{mo2020freeze}, or changing only channel-wise batch statistics \citet{noguchi2019iccv}.
% TransferGAN \citet{wang2018eccv} adapts all weights of the generator and discriminator, and ~\citep{mo2020freeze} demonstrates improvement by freezing early layers of the discriminator.
% \citet{noguchi2019iccv} freezes all pretrained weights $W_0$ and introduces scaling and shift parameters so that a given pretrained generator or discriminator layer $conv(x,W_0)$ is transformed to $conv(x, W_0)\gamma + \beta$. 
% They test a GAN-based formualation, SSGAN, but due to poor performance in this setting they choose to optimize a maximum-likelihood Generative Latent Optimization loss~\citep{bojanowski2017optimizing} over the generator, which stabilizes training but results in blurry image output.
Flow-based methods \citep{gambardella2019arxiv} show promising results in few-shot adaptation, but their architecture is compute- and memory-intensive and requires latent space of the same dimensionality as the data. 
Our method uses a \emph{smaller but more expressive set of parameters} (\figref{methods}), resulting in more natural adapted samples.

\topic{One-shot Image Re-synthesis.}
Recent work in one-shot image synthesis has demonstrated high-quality and diverse results by modeling the \emph{internal} distribution of features from a single image without pretraining \citep{shaham2019iccv,shocher2019iccv}. 
Our work differs as we transfer \emph{external} knowledge from a pretrained GAN to a new domain and, therefore, can generate drastically more diverse samples.
% Our work shares more in common with domain-adaptation than one-shot image synthesis, however, as we seek to transfer outside knowledge from a pretrained GAN to a novel but related low-shot domain.

\topic{Singular Value Decomposition (SVD).} 
SVD factorizes any matrix $M \in \mathbb{R}^{m \times n}$ into unitary matrices $U \in \mathbb{R}^{m \times m}, V \in \mathbb{R}^{n \times n}$ and diagonal matrix $\Sigma$ such that
$M = U \Sigma V^\top$,
where $U,V$ contain the left and right singular vectors respectively and $\Sigma$ contains the singular values along the diagonal entries.
SVD can be interpreted as a decomposition of a linear transformation $x \xrightarrow{} Mx$ into three separate transformations: a rotation/reflection $U$, followed by rescaling $\Sigma$, followed be another rotation/reflection $V^\top$.
The transformation defined by the maximum singular value $\sigma_0=\Sigma^{(1,1)}$ and its corresponding normalized singular vectors represent the maximal axis of variation in the matrix $M$. 
This interpretation is commonly used in data science for dimensionality reduction via PCA~\citep{kwak2008principal}.
PCA can be obtained via SVD on a column-normalized matrix~\citep{golub1971singular}. 
SVD is also used for a wide number of other applications, including regularization \citep{sedghi2018singular}, and quantification of entanglement \citep{martyn2020svdentanglement}, and has also been used to build theoretical background for semantic development in neural networks \citep{saxe2019svdsemantics}.
The work most closely related to ours is GANSpace~\citep{harkonen2020ganspace} for image synthesis editing. 
GANSpace applies PCA within the \emph{latent feature} space of a pretrained GAN to discover semantically-interpretable directions for image editing in the latent space.
In contrast, our work performs SVD on the \emph{weight} space of a GAN to discover meaningful directions for domain adaptation. 
Performing SVD on the weight space enables two critical differences between our work and \citet{harkonen2020ganspace}: 
(i) we edit the entire output \emph{distribution} rather than one image, and 
(ii) rather than manual editing, we adapt to a new domain.

%%%%%%%%%%%%%%%%%%%%%%%%%%%%%%%%%%%%%%%%%%%%%%%%%%%%%%
% Method
%%%%%%%%%%%%%%%%%%%%%%%%%%%%%%%%%%%%%%%%%%%%%%%%%%%%%%
\section{Few-Shot GAN} 
% In this section we first describe background of our singular value adaptation method (Section~\ref{method:analysis}).
% We then present the details of our methodology (Section \ref{method:overview}), and finally discuss challenges and problems with the standard evaluation metrics in the low-shot GAN transfer setting (Section \ref{method:evaluation}).

\subsection{Overview}
\label{method:analysis}

Our goal is to improve GAN finetuning on small image domains by discovering a more effective and constrained parameter space for adapting the pretrained weights. 
We are inspired by prior work in GAN adaptation showing that constraining the space of trainable parameters can lead to improved performance on target domain~\citep{rebuffi2017residual,mo2020freeze,noguchi2019iccv}. 
In contrast to identifying the parameter space within the model architecture, we propose to discover a parameter space based on the pretrained weights. 
Specifically, we apply singular value decomposition to the pretrained weights and uncover a basis representing orthogonal directions of maximum variance in the weight space.
To explore the interpretation of the SVD representation, we visualize the top three singular values of synthesis and style layers of StyleGAN2~\citep{karras2019analyzing}.
We observe that varying the singular values corresponds to natural and semantically-meaningful changes in the output image as shown in Figure \ref{fig:svd-viz}.
% TODO: Show all layers in the supplementary material.
Changing the singular values can be interpreted as changing the entanglement between orthogonal factors of variation in the data (singular vectors), providing an expressive parameterization of the pretrained weights, which we leverage for adaptation as described in the following section.

\subsection{Adaptation Procedure}
\label{method:overview}
Our method first performs SVD on both the generator and discriminator of a pretrained GAN and adapts the singular values to a new domain using standard GAN training objectives.
A generator layer $G^{(\ell)}$ or a discriminator layer $D^{(\ell)}$ may consist of either 2D ($c_\mathrm{in} \times c_\mathrm{out}$) fully-connected weights or 4D ($k \times k \times c_\mathrm{in} \times c_\mathrm{out}$) convolutional filter weights.
We apply SVD separately at every layer of the generator $G^{(\ell)}$ and discriminator $D^{(\ell)}$.
Next, we describe the decomposition process for a single layer of pretrained weights $W_0^{(\ell)}$.
For fully-connected layer $W_0^{(\ell)}$, we can apply SVD directly on the weight matrix.
For 4D convolution weights $W_0^{(\ell)} \in \mathbb{R}^{k \times k \times c_{in} \times c_{out}}$ this is not feasible because SVD operates only on a 2D matrix. 
We therefore reshape the 4D tensor by flattening across the spatial and input feature channels before performing SVD to obtain a 2D matrix $W_0^{(\ell)} \in \mathbb{R}^{k^2 c_{in} \times c_{out}}$.
Our intuition is that the spatial-feature relationship in the pretrained model should be preserved during the adaptation. 
We apply SVD over each set of flattened convolutional weights or fully convolution weights to obtain the decomposition:
\begin{align}
    W_0^{(\ell)} = (U_0 \; \Sigma_0 \; V_0^\top)^{(\ell)}.
\end{align}

After decomposing the pretrained weights, we perform domain adaptation by freezing pretrained left/right singular vectors in $(U_0, V_0)^{(\ell)}$ and optimizing the singular values $\Sigma = \lambda \Sigma_0$ using a standard GAN objective to obtain transferred weights (Figure \ref{fig:methods}):
\begin{align}
    W^{(\ell)}_\Sigma = (U_0 \; \Sigma \; V_0^\top)^{(\ell)}
\end{align}
Effectively, our GAN domain adaptation aims to find a new set of singular values in each layer of a pretrained model so that the generated outputs match the distribution of the target domain. 

During forward propagation, we reconstruct weights $W_\Sigma$ using the finetuned singular values at each convolution or fully-connected layer of the generator and discriminator before applying the operation.

\newcommand{\svwidth}{0.0952}
\begin{figure}
    \centering
    \setlength{\tabcolsep}{2pt}
    \begin{tabular}{c c@{}c@{}c c@{}c@{}c c@{}c@{}c}
     &  & $\text{style}_4$ & & & $\text{conv}_{8\times8}$ & & \multicolumn{3}{c}{$\text{conv}_{1024\times1024}$} \\
    Original & $10\sigma_0$ & $10\sigma_1$ & $10\sigma_2$ & $5\sigma_0$ & $5\sigma_1$ & $5\sigma_2$ & $10\sigma_0$ & $10\sigma_1$ & $10\sigma_2$ \\
         \includegraphics[width=\svwidth\textwidth]{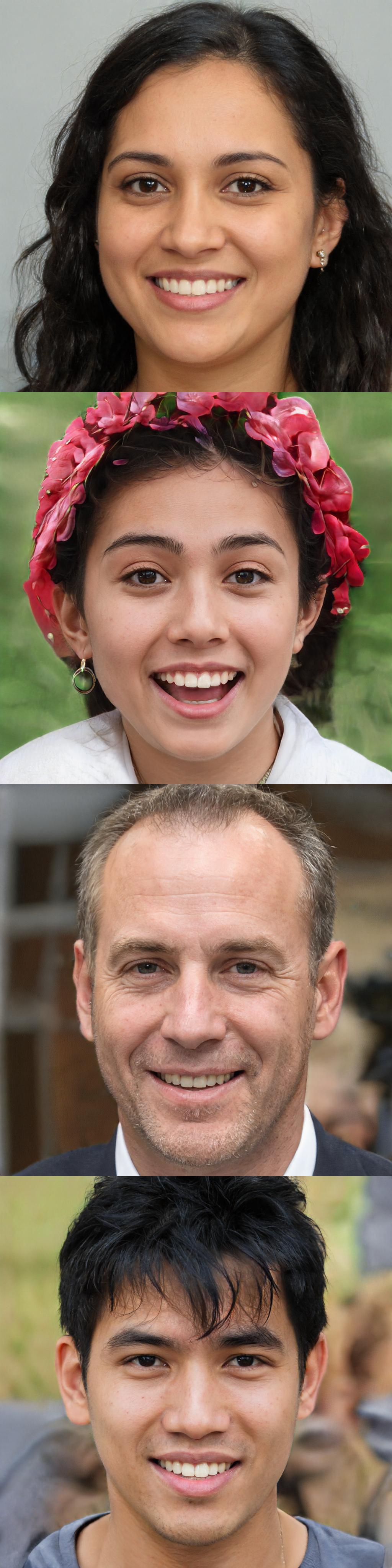} &
         %%%
        % Map3
        \includegraphics[width=\svwidth\textwidth]{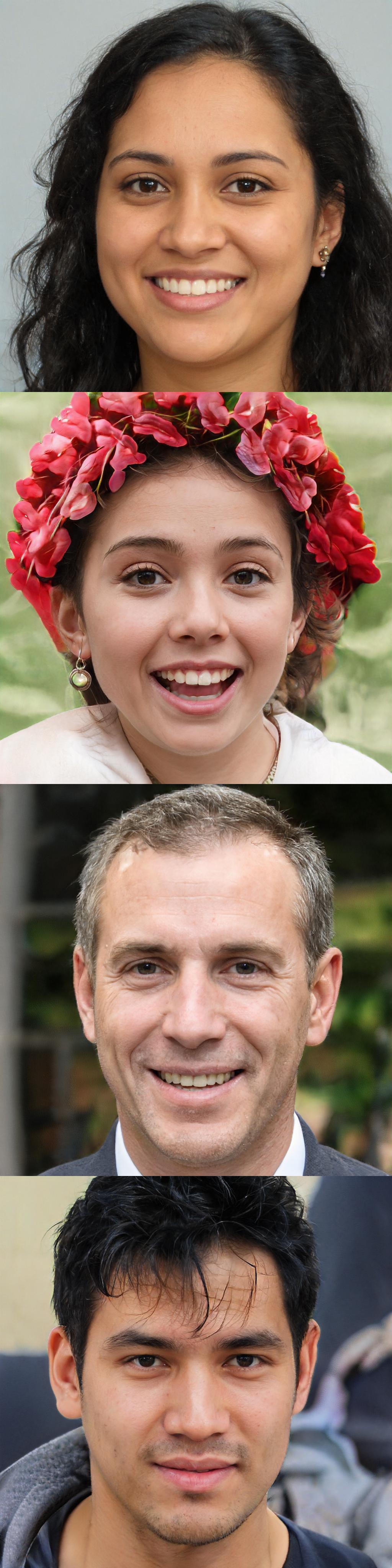} & %
        \includegraphics[width=\svwidth\textwidth]{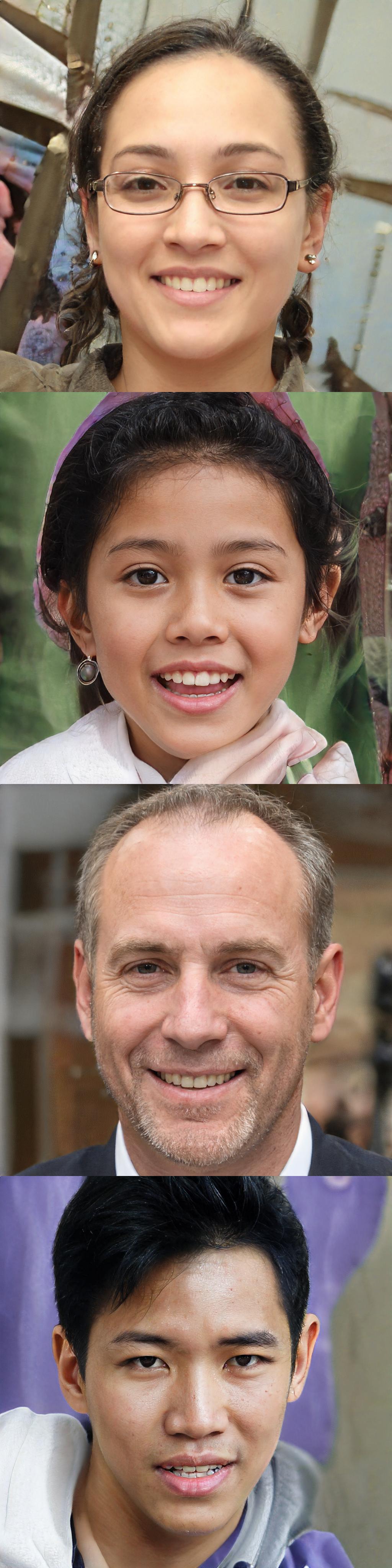} & %
        \includegraphics[width=\svwidth\textwidth]{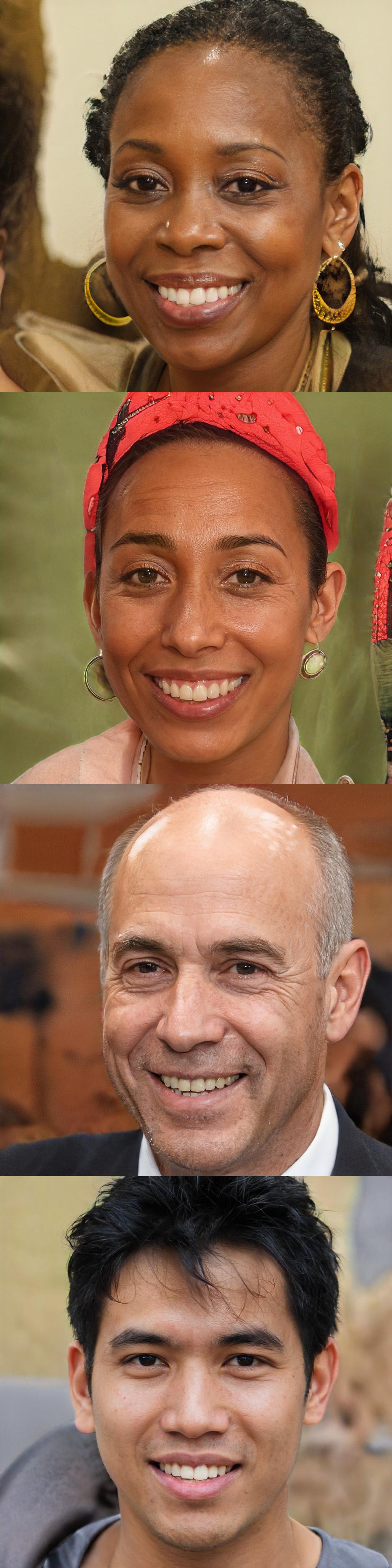} & %
        % 8x8_Conv0_up
        \includegraphics[width=\svwidth\textwidth]{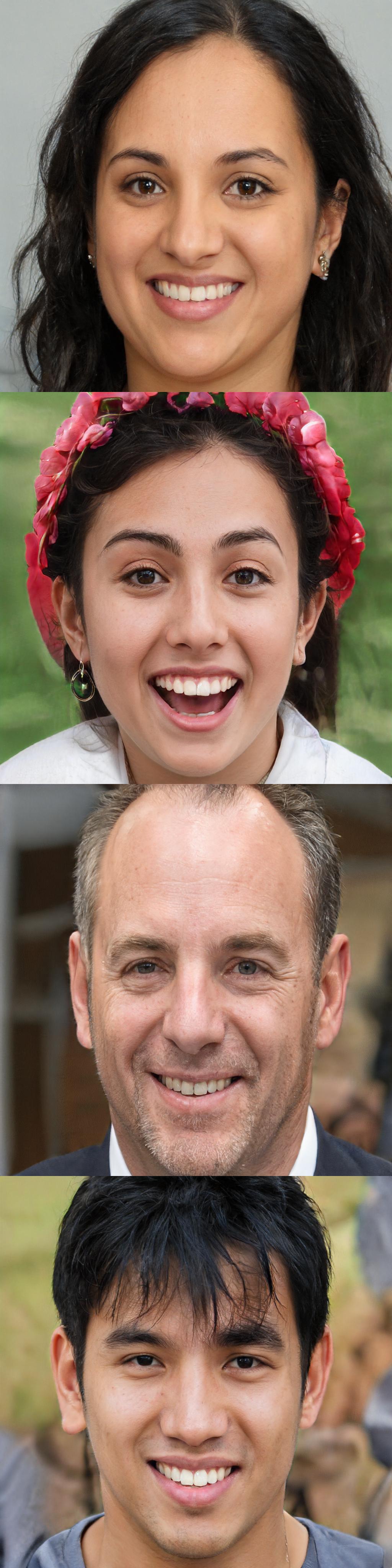} & %
        \includegraphics[width=\svwidth\textwidth]{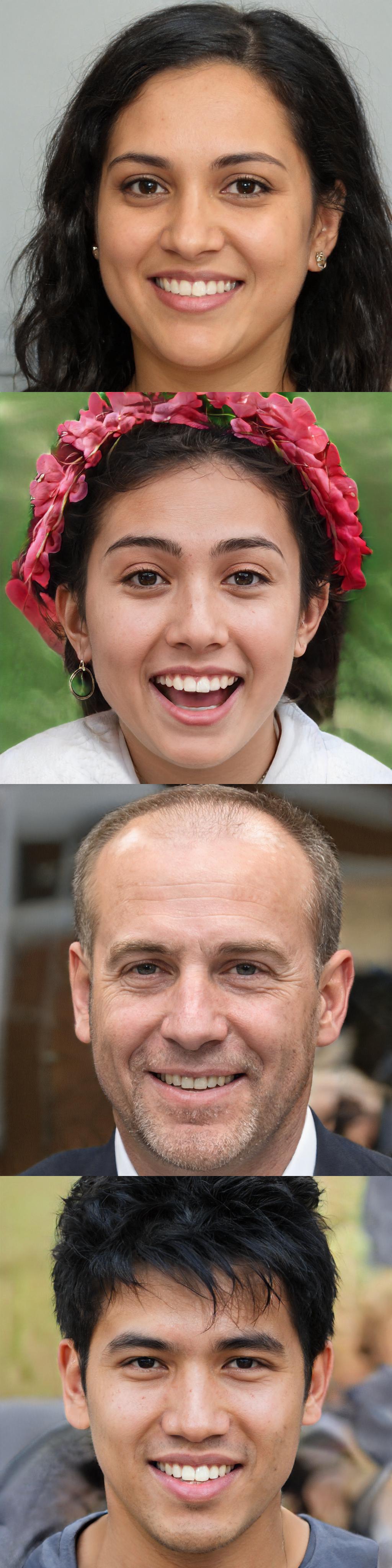} & %
        \includegraphics[width=\svwidth\textwidth]{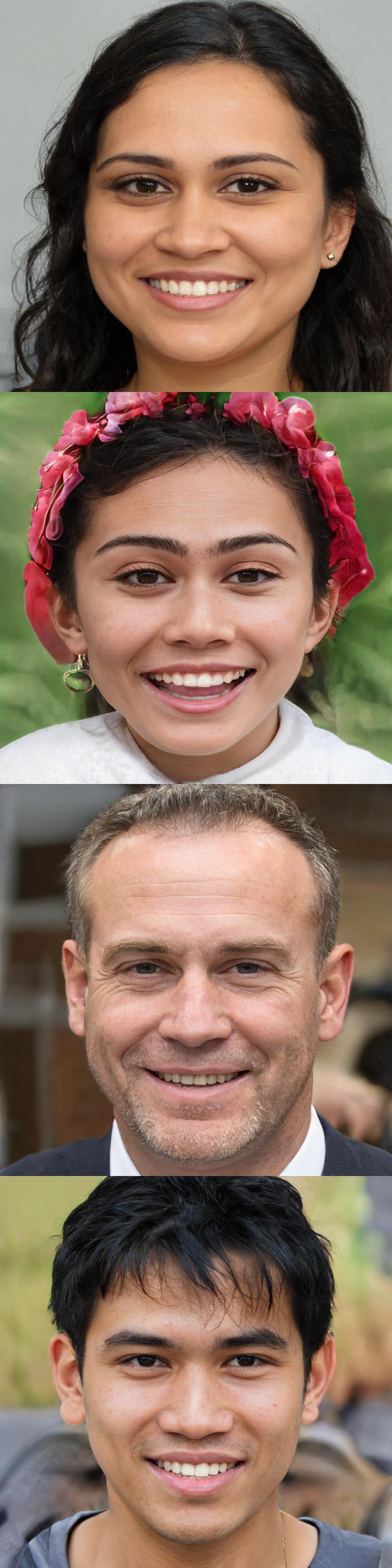} & %
        % 1024
        \includegraphics[width=\svwidth\textwidth]{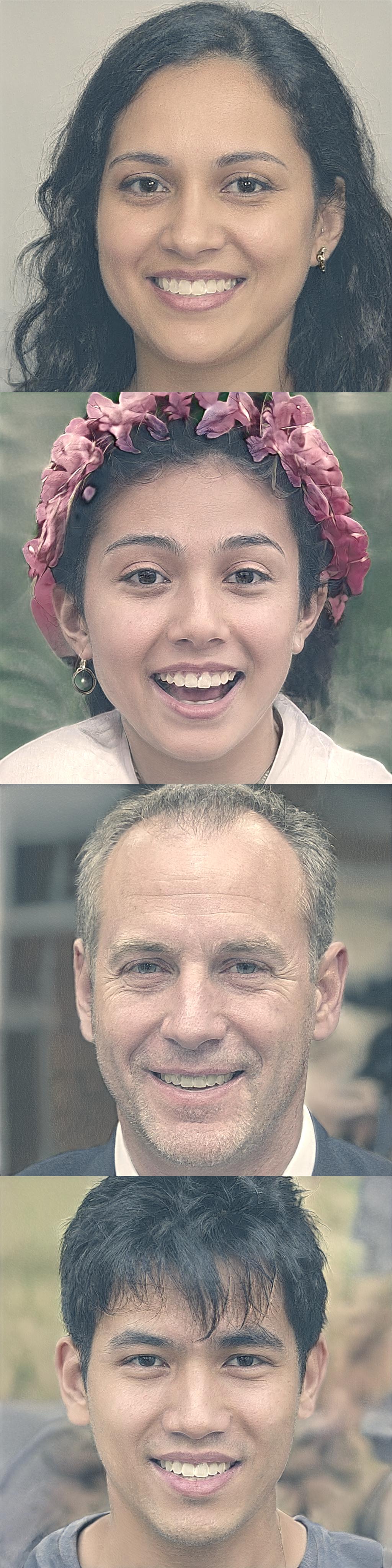} & %
        \includegraphics[width=\svwidth\textwidth]{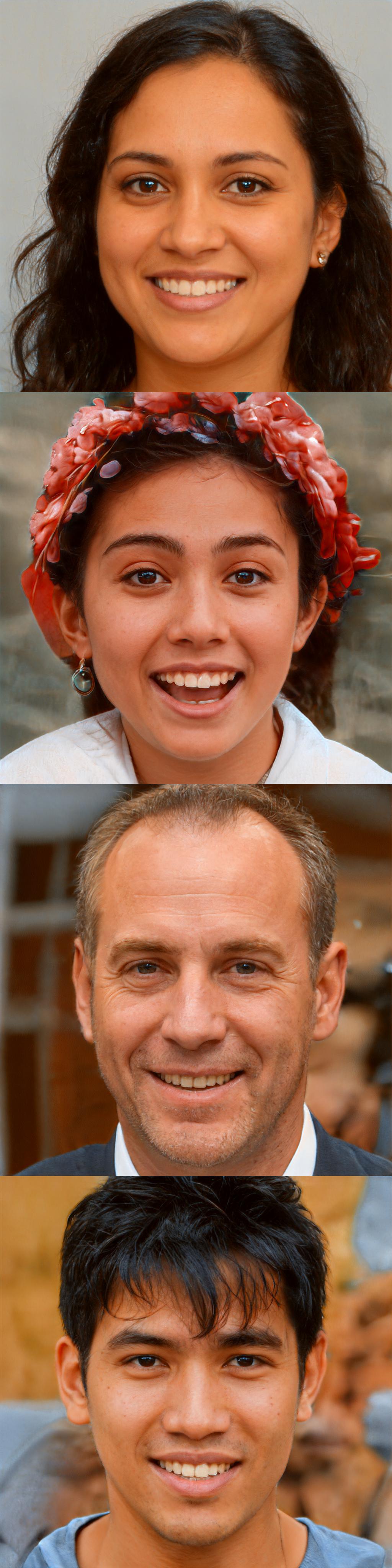} & %
        \includegraphics[width=\svwidth\textwidth]{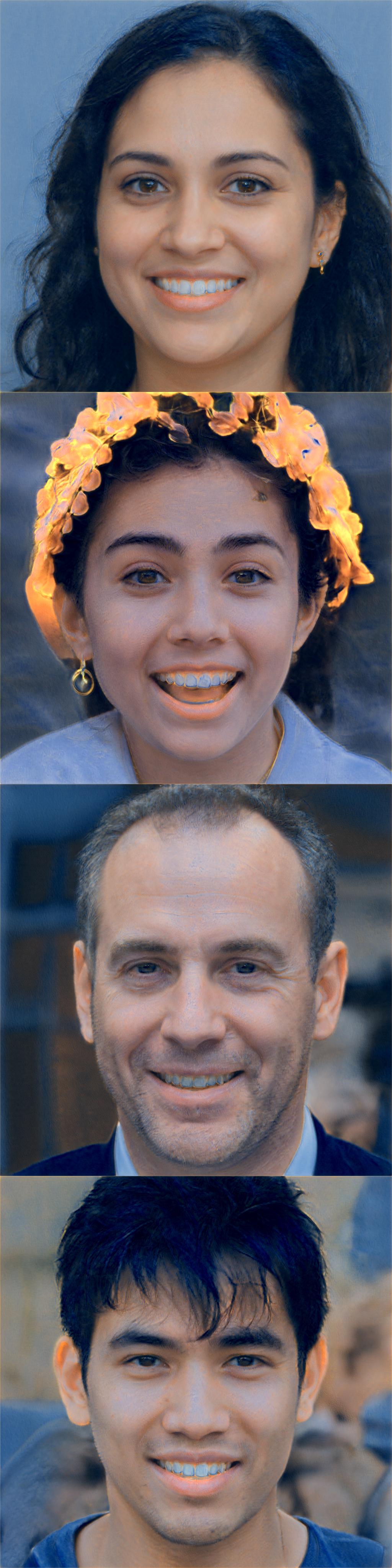} \\
    \end{tabular}
    \caption{\textbf{Effects of singular values.} 
    We visualize FSGAN's adaptation space by magnifying the top 3 singular values $\sigma_0, \sigma_1, \sigma_2$ from SVD performed on style and conv layers of a StyleGAN2 \citep{karras2019style,karras2019analyzing} pretrained on FFHQ.
    In mapping layer 4 ($\text{style}_4$), the leading $\sigma$s change the age, skin tone, and head pose. 
    In synthesis layer 2 ($\text{conv}_{8\times 8}$), face dimensions are modified in term of face height/size/width.
    In synthesis layer 9 ($\text{conv}_{1024 \times 1024}$), the face appearance changes in finer pixel stats such as saturation, contrast, and color balance.
    }
    \label{fig:svd-viz}
\end{figure}

\subsection{Training \& Inference}

\begin{figure}
    \centering
    \small
    \newlength\ftes \setlength\ftes{0.35cm} \setlength\tabcolsep{0.2pt}
    \vspace{-.65cm}
    %\begin{tabular}{lcc}
    \begin{minipage}{0.43 \linewidth}
        \begin{tabu} to \linewidth {X[l 0.1]X[c 0.15]X[c 0.4]}
          \toprule
          \textbf{t} & \textbf{FID} & \textbf{Interpolation} \\ 
          \midrule
          0 & 121.21 &
          \includegraphics[height=\ftes]{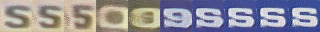} \\
          20 & 154.25 &
          \includegraphics[height=\ftes]{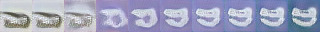} \\
          40 & 134.22 & 
          \includegraphics[height=\ftes]{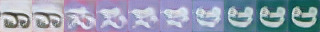} \\
          80 & 102.87 & 
          \includegraphics[height=\ftes]{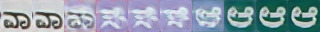} \\
          120 & 93.65 & 
          \includegraphics[height=\ftes]{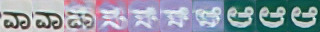} \\  
          180 &\textbf{92.94}& 
          \includegraphics[height=\ftes]{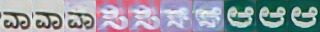} \\
          \midrule
          \multicolumn{2}{l}{Train set (10-shot):} & \includegraphics[height=\ftes]{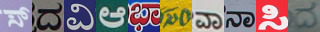} \\
        \bottomrule
        %\end{tabular}
        \end{tabu}
    \end{minipage}
    \hfill
    \begin{minipage}[t]{0.54 \linewidth}
        \vspace{-15ex}
        \caption{\textbf{Problem with FID as a few-shot metric.}
        TGAN \citep{wang2018eccv} adaptation from English characters to 10-shot Kannada characters ({\em Bottom}) \citep{de2009character}. 
        The adaptation process is illustrated by interpolating two random latent vectors at different timesteps (t=20 means 20K images seen during training).
        We measure FID against a 2K-image Kannada set, from which the 10 images was sampled.
        The interpolation shows larger timesteps (t) tend to memorize the 10-image training set while yielding lower FID, revealing that FID favors overfitting and is not suitable for the few-shot setting.
        %For this 10-shot adaptation example, FID appears to favor iterations (t) that memorize the train set. 
        %The optimal FID is at timestep 180, but this iteration appears to overfit, as earlier iterations have more novel outputs and smoother interpolation.
        %Transfer is from English$\rightarrow$10-shot Kannada characters \citep{de2009character} interpolated between random latent vectors at different timesteps (t=Kimg) of GAN fine-tuning.
        %FID is measured with respect to a 2K-image Kannada dataset, from which the 10-shot set was sampled uniformly.
        } 
        \label{fig:metric-problem}
    \end{minipage}
    \vspace{-2ex}
\end{figure}

Our experiments use the StyleGAN2~\citep{karras2019analyzing} training framework, which optimizes a logistic GAN loss (Equation \ref{eqn:gan}) with latent space gradient regularization and a discriminator gradient penalty.
We retrain the singular values $\Sigma$ for a fixed number of timesteps (20K images or 16K for 5-shot).
We find limiting the training time is essential for quality and diversity in the low-shot setting, as longer training often leads to overfitting or quality degradation (examples in Figure \ref{fig:metric-problem} \& \ref{fig:nshot}).
Like \citet{noguchi2019iccv}, we use the truncation trick~\citep{brock2018iclr} during inference, but our method works with a less-restrictive truncation parameter of $\psi=0.8$, which enables more diversity in the generated images.

\subsection{Evaluation in Few-Shot Synthesis}
\label{method:evaluation}
A common adverse outcome in few-shot image generation is overfitting to the target set, such that all generated images look similar to the training data. 
Evaluation metrics should reflect the diversity of generated images, so that memorization is penalized.
The standard evaluation practice used in prior low-shot GAN adaptation work \citep{wang2018eccv,noguchi2019iccv,mo2020freeze} is to estimate FID \citep{heusel2017nerips} using a large held-out \textit{test set} with 1K+ images, from which the low-shot \textit{training set} was sampled. 
Standard GAN evaluation typically measures FID with respect to the \emph{training set}, but in the low-shot setting, this is not desirable because the generator may simply memorize the training set.
However, we find that even when measuring FID against a held-out test set, this evaluation still favors overfitted or poor-quality models, as shown in Figure \ref{fig:metric-problem}.
FID between real and fake images is calculated as the Frechet distance between perceptual features $p_r(X)$ and $p_f(Z)$: 
\begin{align}
    ||\mu_r - \mu_f||^2  + \text{Tr}(C_r + C_f - 2\sqrt{C_rC_f}).
\end{align}
where it is assumed features are Gaussian \ie $p_f(Z) = N(\mu_f, C_f)$ and $p_r(X) = N(\mu_r, C_r)$. 
In the few-shot setting, our n-shot training set $T=(x_{1},x_{2},...,x_n)$ is sampled from our test set $p_r(X)$. 
Assuming $T$ is chosen at random, its sample mean and variance $\hat{\mu}, \hat{\sigma^2}$ are unbiased estimators of $\mu_r, C_r$.
Therefore if the generator \emph{memorizes} $T$, its statistics approximate $\mu_r, C_r$.
This artificially decreases the FID of an overfit model (Figure \ref{fig:metric-problem}). 
Consequently, we suggest that FID should be supplemented with additional metrics and extensive qualitative results in the low-shot setting. 
In high-data settings, a very large number of parameters would be required to memorize the images, so this problem is less likely to occur. 
Based on these observations, throughout our evaluation, we limit training timesteps rather than select the step with the best FID as we find the latter approach gives more inferior qualitative results. 
To address the limitations of standard metrics for GAN evaluation, we also report sharpness~\citep{kumar2012sharpness} and face quality index~\citep{hernandez2019faceqnet} for human face transfer.

%%%%%%%%%%%%%%%%%%%%%%%%%%%%%%%%%%%%%%%%%%%%%%%%%%%%%%%%%%%%%%%%%%%
% Experiments
%%%%%%%%%%%%%%%%%%%%%%%%%%%%%%%%%%%%%%%%%%%%%%%%%%%%%%%%%%%%%%%%%%%
\section{Experiments}

\subsection{Settings}

We adapt a pretrained model to a new target domain using only 5-100 target images, as we focus on scenarios with 1-2 orders fewer number of training samples than standard data-efficient GAN adaptation methods \citep{wang2018eccv,mo2020freeze,noguchi2019iccv}.
%As this setting is highly prone to overfitting, we limit training timesteps for all methods to preserve the diversity and quality of the pretrained model. 
As discussed in Section \ref{method:evaluation}, we find that the FID score is unsuitable in the low-shot regime due to overfitting bias.
However, we still report the FID scores of our experiments for completeness.
In addition, we report additional quality metrics and extensive qualitative results.

%Sections \ref{sec:near} \& \ref{sec:far} demonstrate near and far domain adaptation in the $\sim25$-shot setting, and Section \ref{sec:nshot} shows results across different n-shot settings.

\begin{figure}[th!]
    \newlength\ftqn
    \setlength\ftqn{1.1cm} % Image size
    \newcommand\nspc{4pt} % Manual spacing
    % Good seeds: 405, ~406, ~407, 411, 412, 413 / 421, 433, 441, ~442
    \newcommand{\none}{421} % Seed 1
    \newcommand{\ntwo}{433} % Seed 2
    \newcommand{\nthree}{411} % Seed 3 (411
    \newcommand{\nfour}{412} % Seed 4    
    \newcommand{\nfive}{407} % Seed 5
    \newcommand{\nsix}{408} % Seed 6
    \setlength\tabcolsep{0pt} % For tight image grids
    \renewcommand*{\arraystretch}{0} % For tight image grids
    \centering
    \begin{tabular}{ccccc}
        5-shot & 15-shot & 50-shot & 100-shot \vspace{\nspc} \\
        %%%%%%%%%%%%%%%%%%%%%%%%%%%%%%
        %%%%%%% FD %%%%%%%%%%%
        \rotatebox[origin=c]{90}{\small \tb{(a)} FD}%
        %%%%%%%%%%%% 5-shot %%%%%%%%%%%%%
        \begin{tabular}{ccc}
            \includegraphics[width=\ftqn]{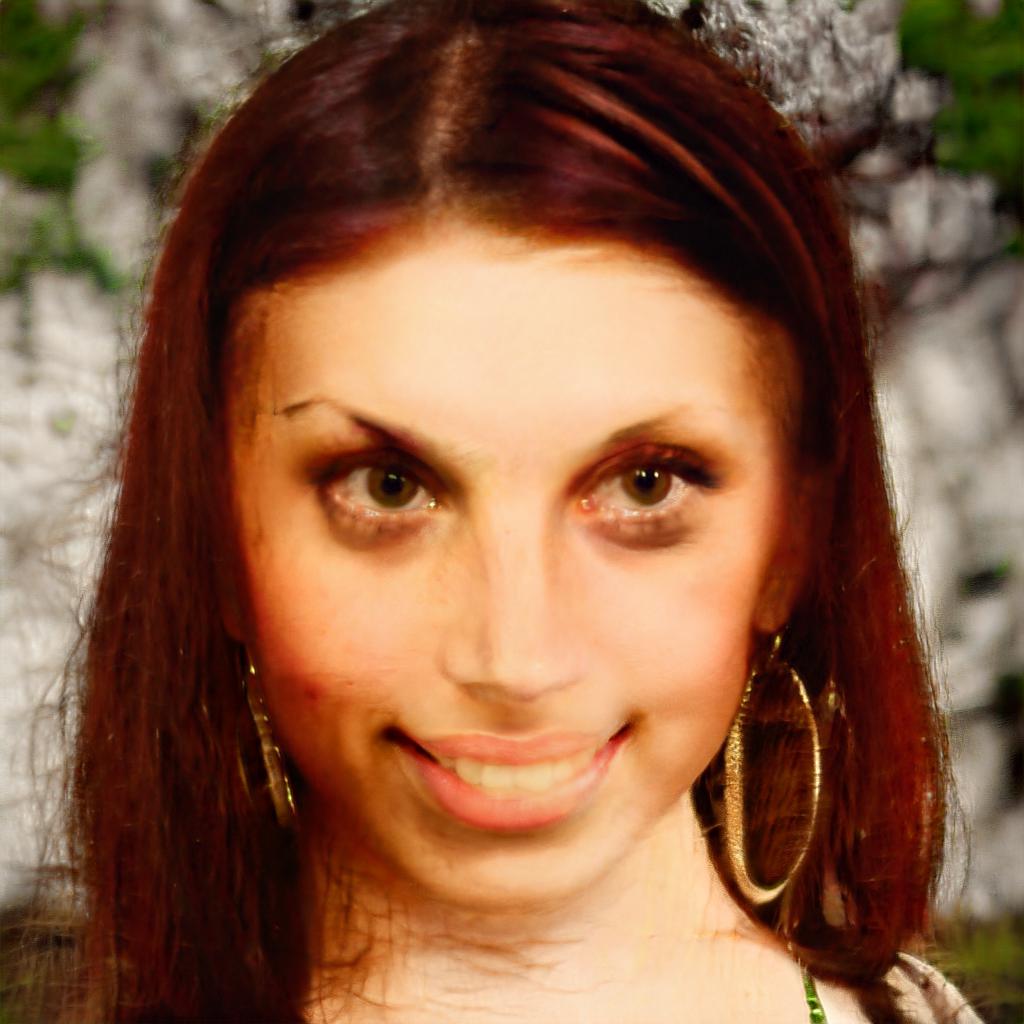} & \includegraphics[width=\ftqn]{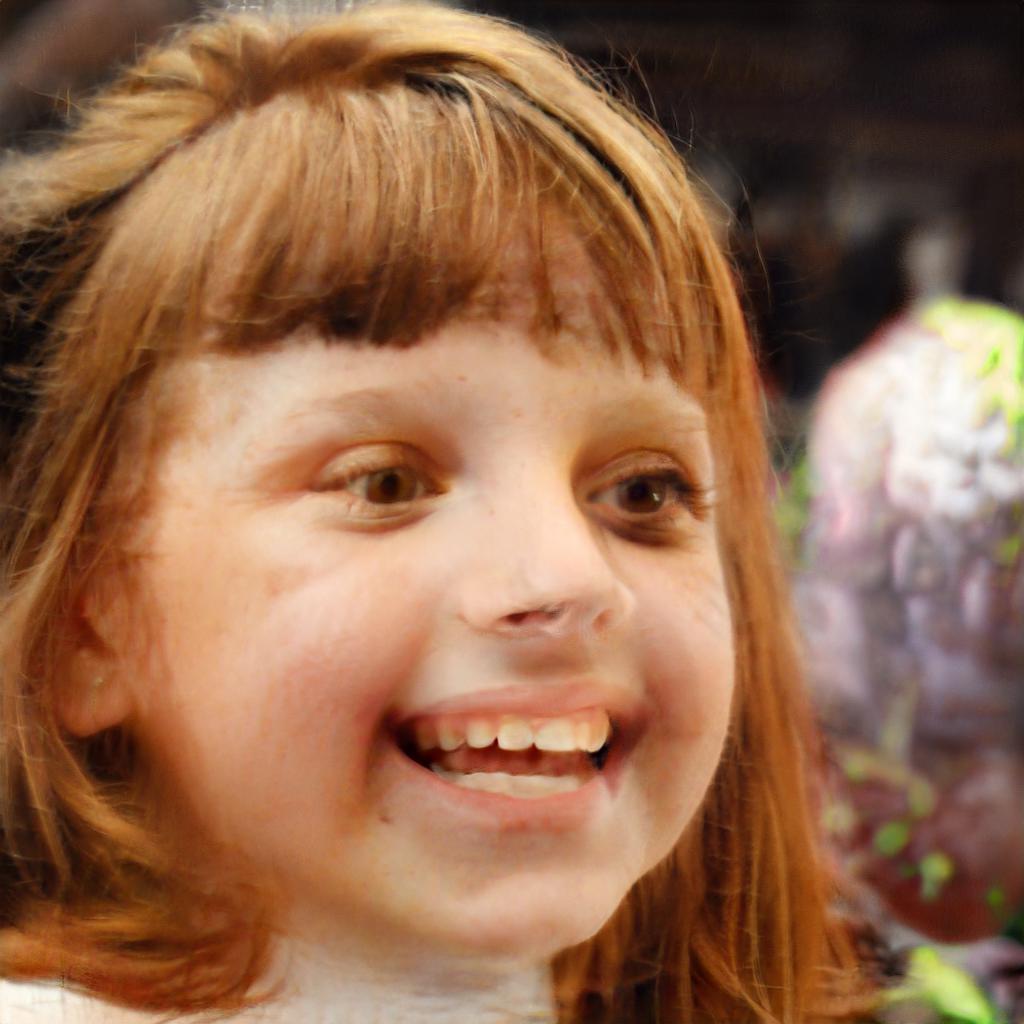} &
            \includegraphics[width=\ftqn]{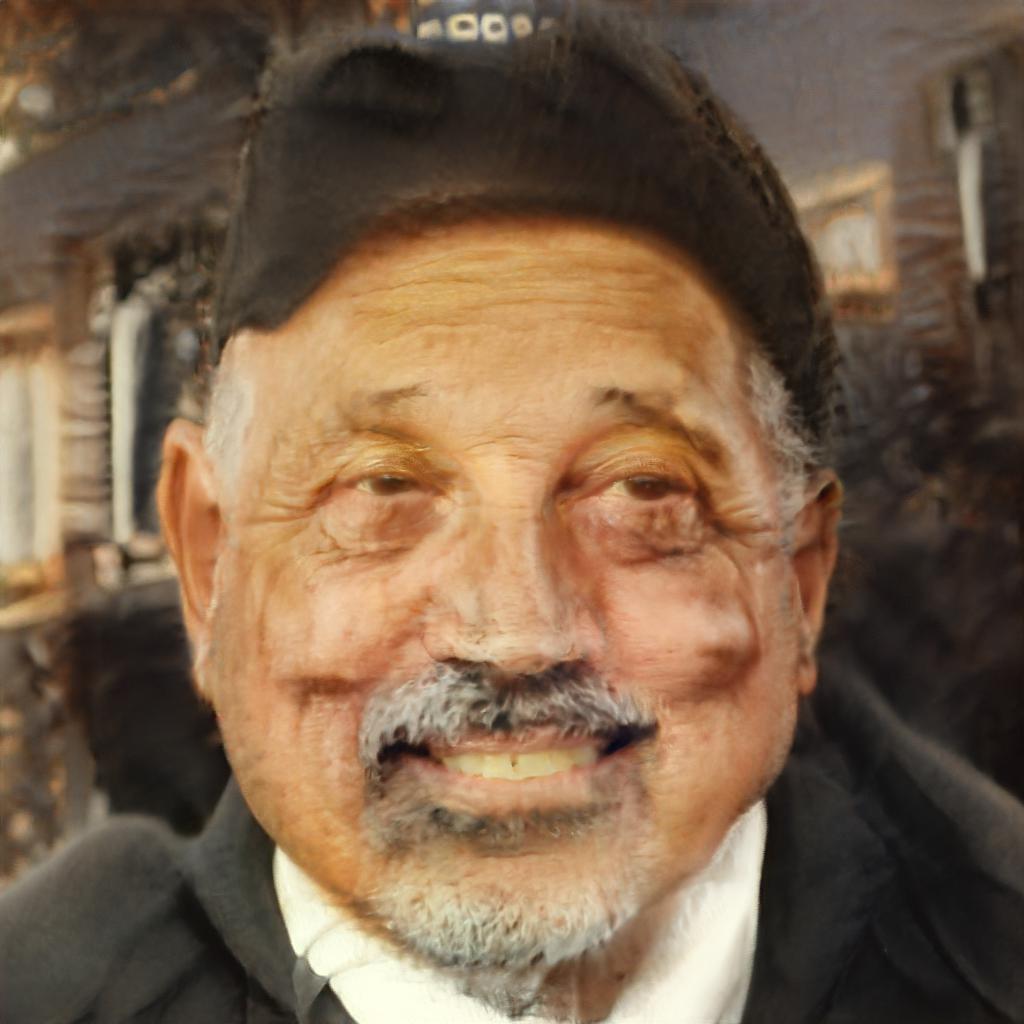} \\ \includegraphics[width=\ftqn]{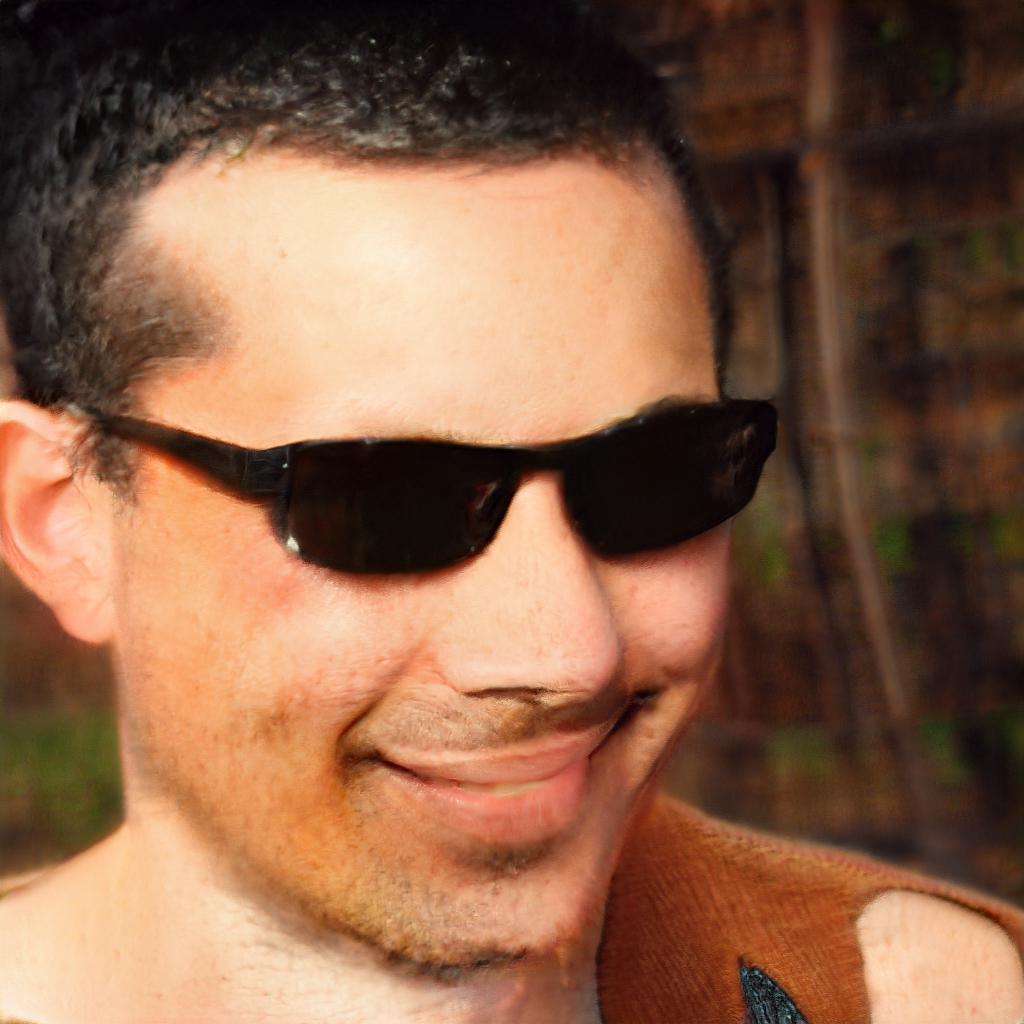} &
            \includegraphics[width=\ftqn]{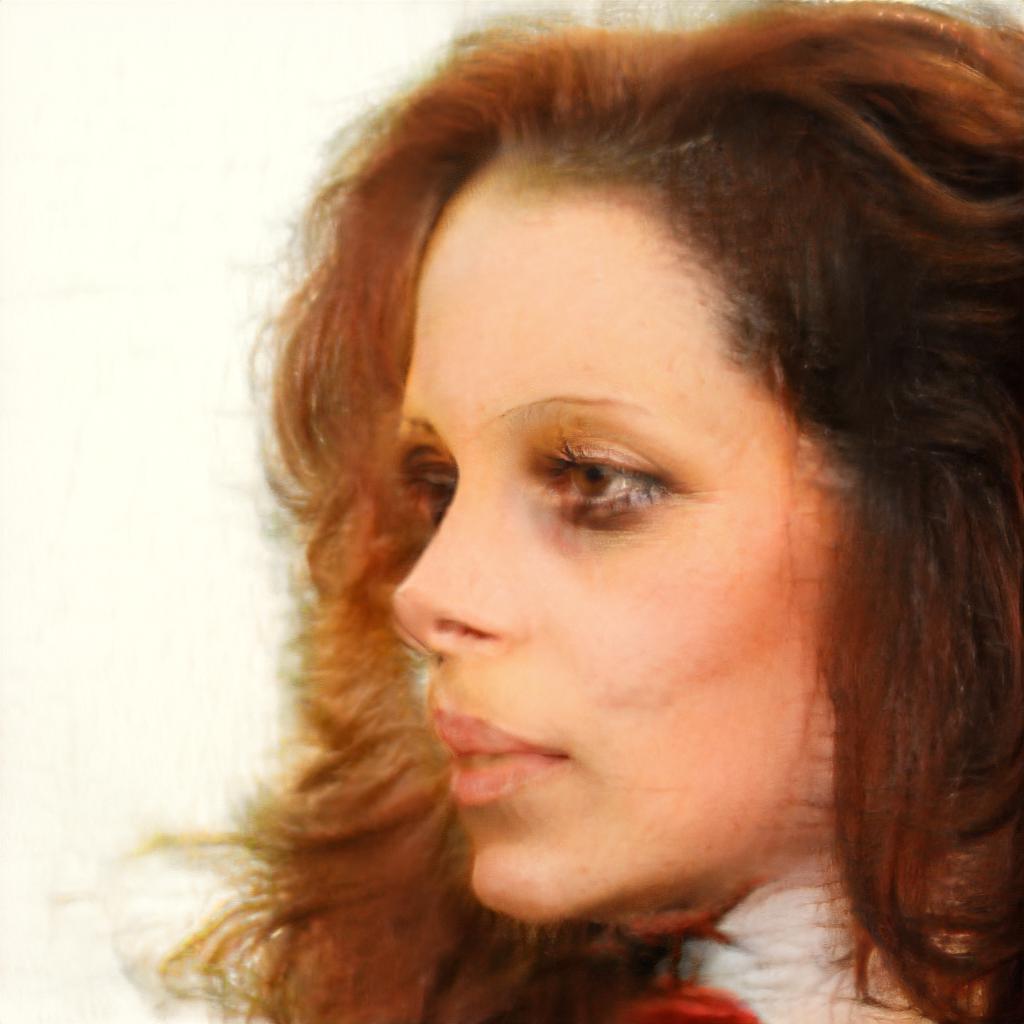} & \includegraphics[width=\ftqn]{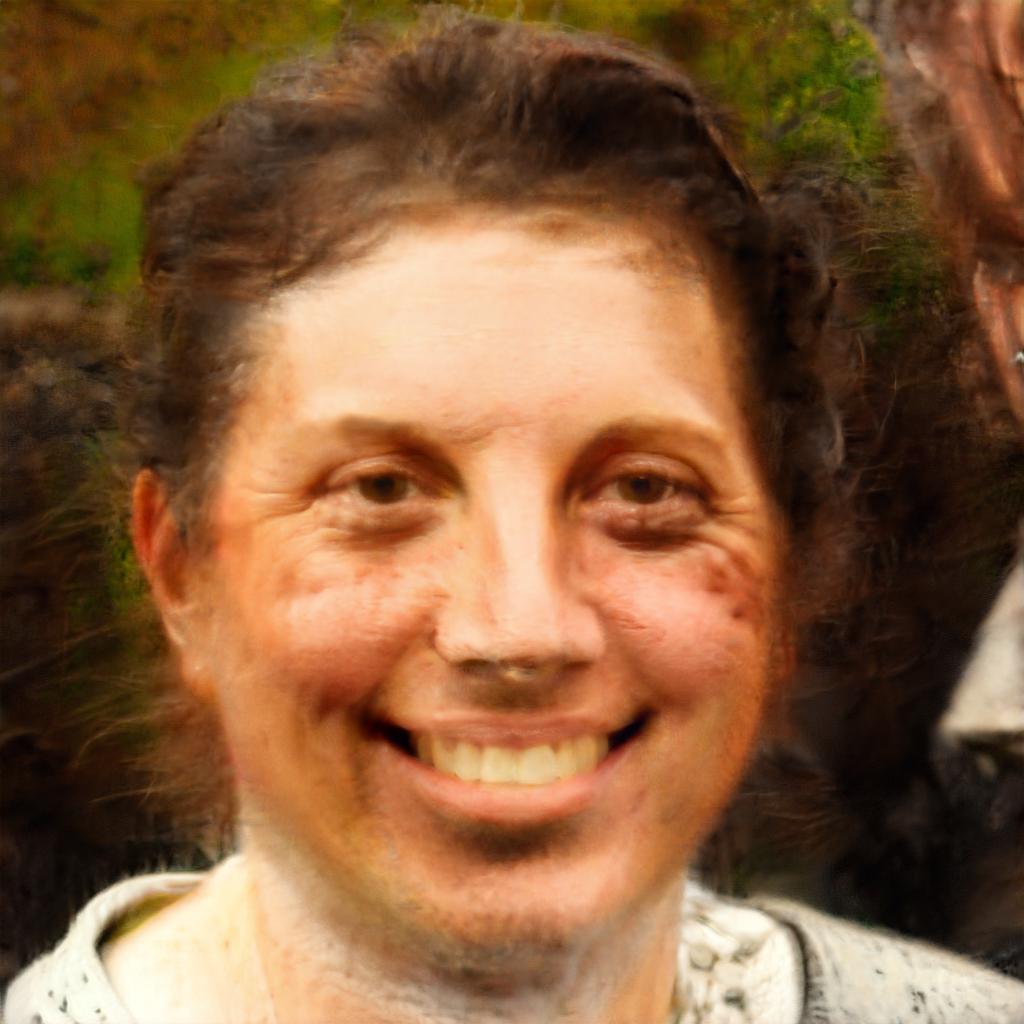} 
        \end{tabular}\hspace{\nspc}  & 
        %%%%%%%%%% 15-shot %%%%%%%%%%%%%%%
        \begin{tabular}{ccc}
            \includegraphics[width=\ftqn]{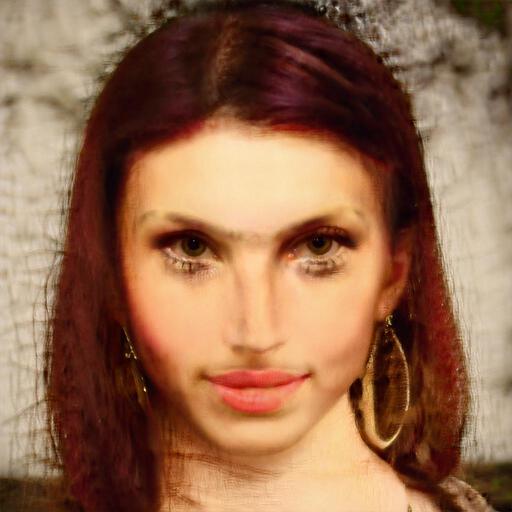} & \includegraphics[width=\ftqn]{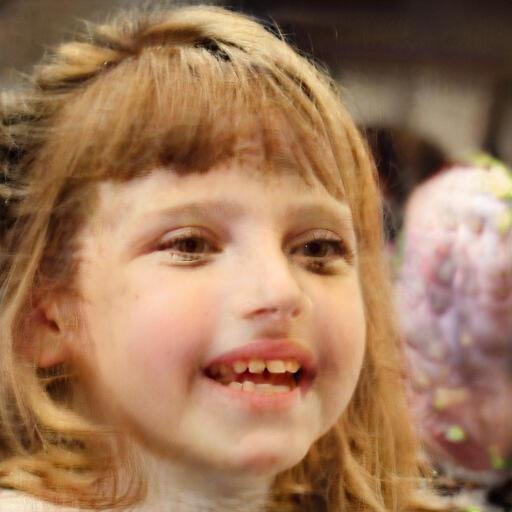} &
            \includegraphics[width=\ftqn]{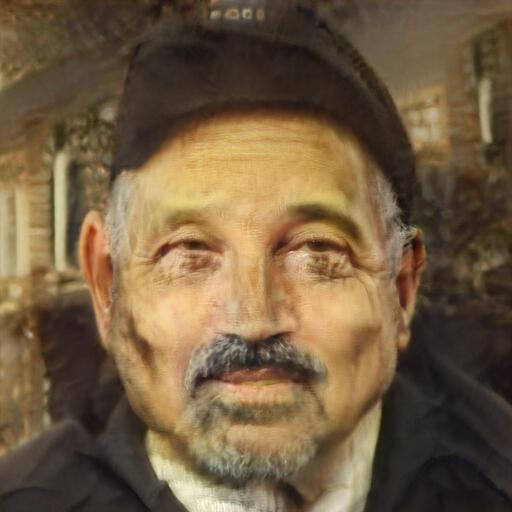} \\ \includegraphics[width=\ftqn]{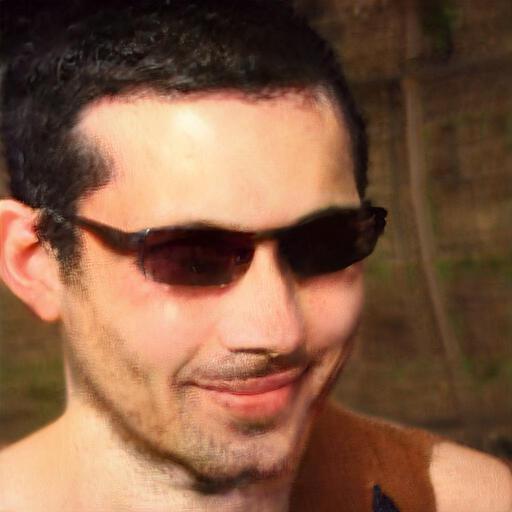} &
            \includegraphics[width=\ftqn]{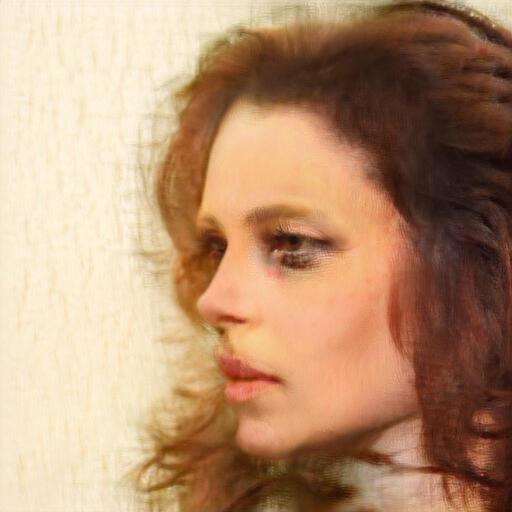} & \includegraphics[width=\ftqn]{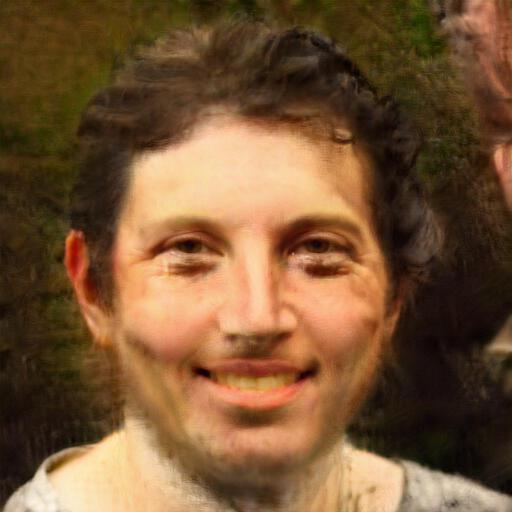}             
        \end{tabular}\hspace{\nspc}  & 
       %%%%%%%%%%%%% 50-shot %%%%%%%%%%%%
        \begin{tabular}{ccc}
            \includegraphics[width=\ftqn]{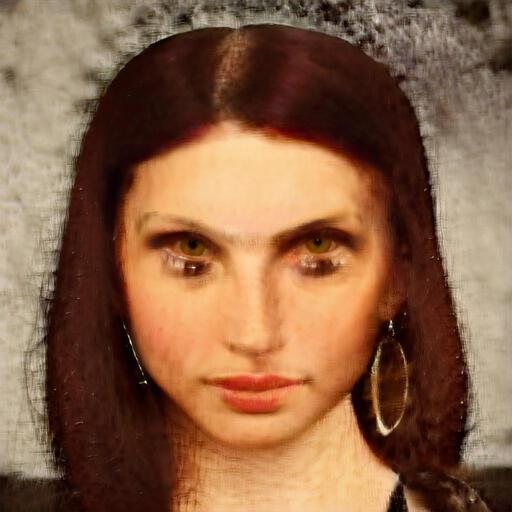} & \includegraphics[width=\ftqn]{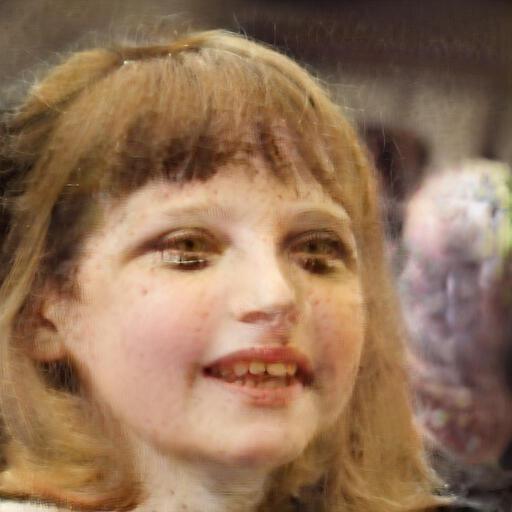} &
            \includegraphics[width=\ftqn]{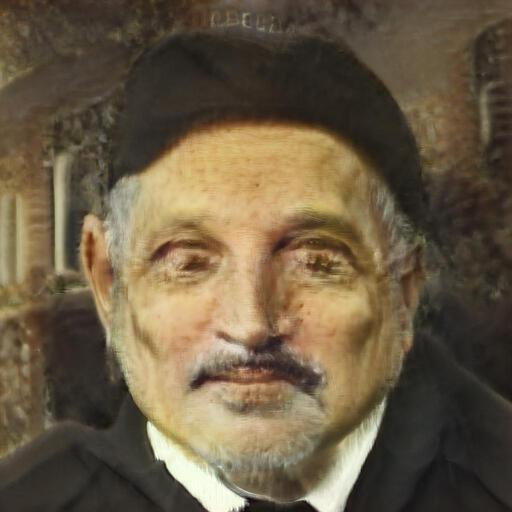} \\ \includegraphics[width=\ftqn]{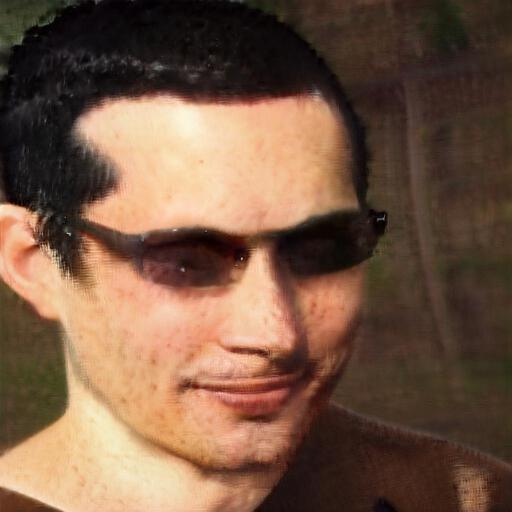} &
            \includegraphics[width=\ftqn]{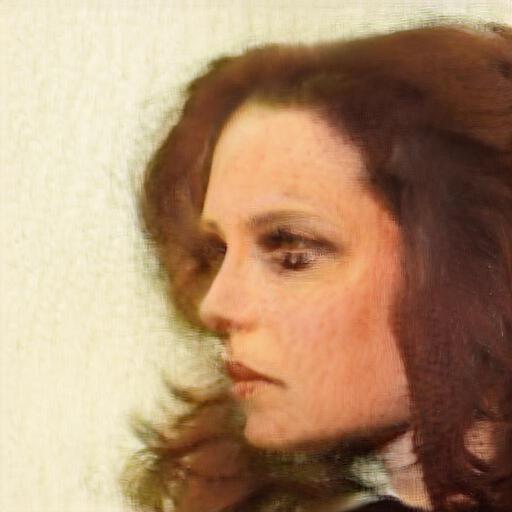} & \includegraphics[width=\ftqn]{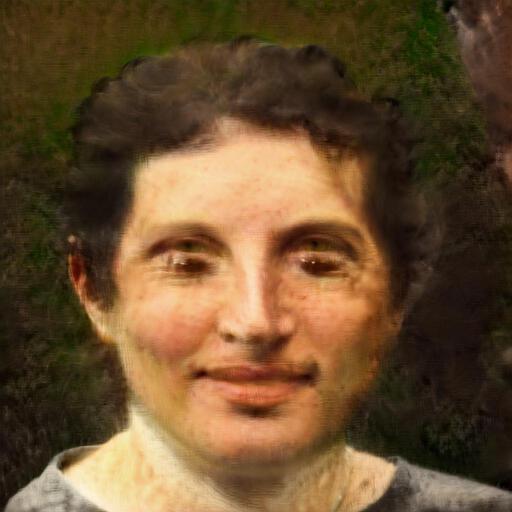} 
        \end{tabular}\hspace{\nspc}  &
       %%%%%%%%%%%%% 100-shot %%%%%%%%%%%%
        \begin{tabular}{ccc}
            \includegraphics[width=\ftqn]{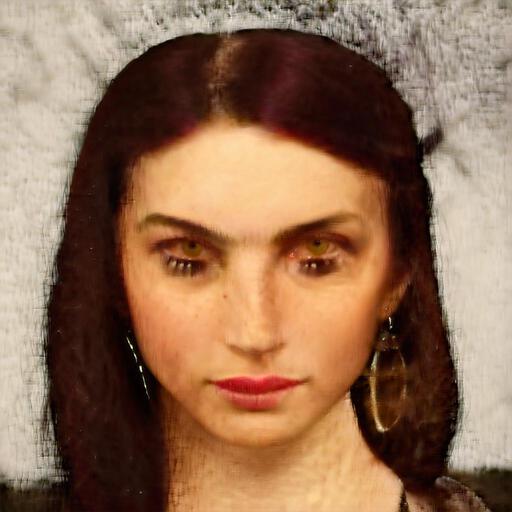} & \includegraphics[width=\ftqn]{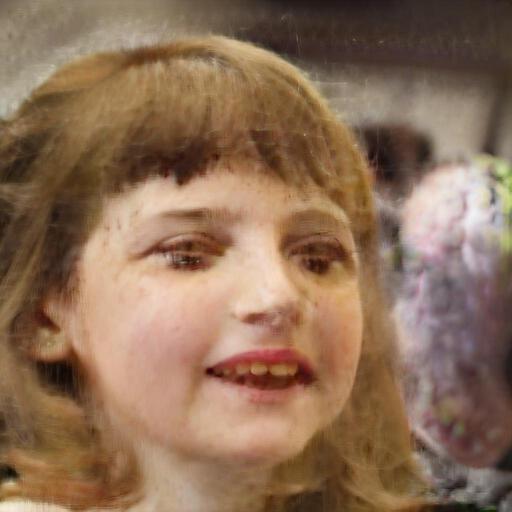} &
            \includegraphics[width=\ftqn]{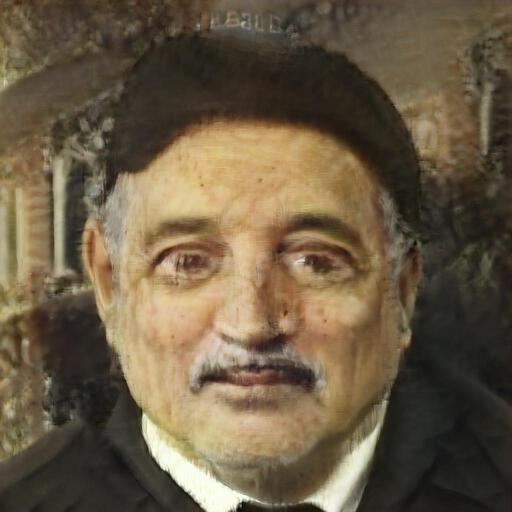} \\ \includegraphics[width=\ftqn]{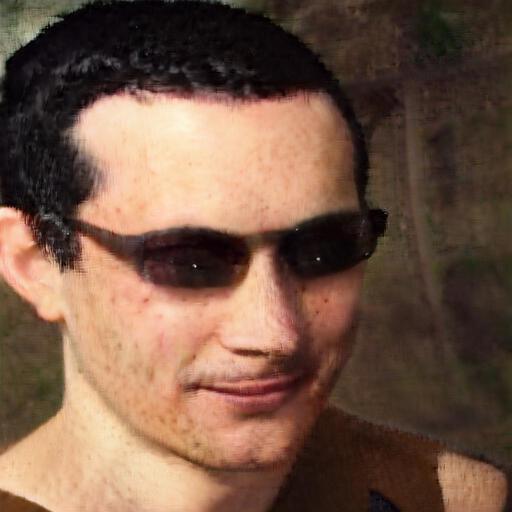} & 
            \includegraphics[width=\ftqn]{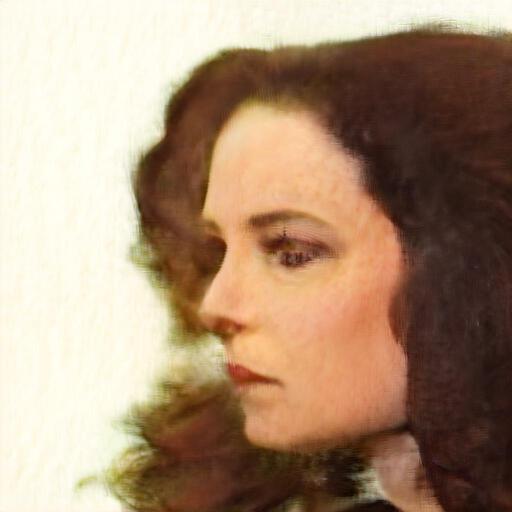} & \includegraphics[width=\ftqn]{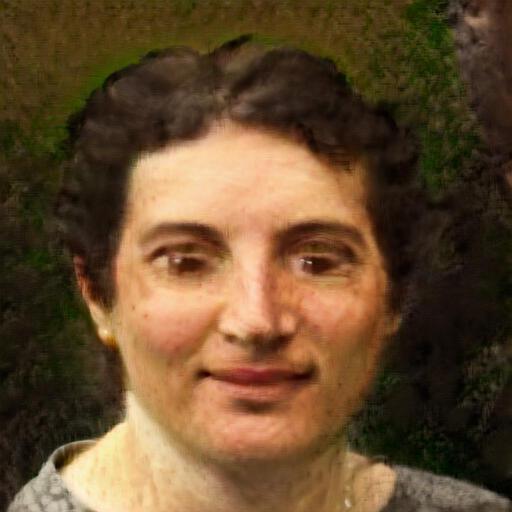}  
        \end{tabular} \vspace{\nspc} \\
        %%%%%%%%%%%%%%%%%%%%%%%%%%%%%%
        %%%%%%% FD-FT %%%%%%%%%%%%%%
        %%%%%%%%%%%%%%%%%%%%%%%%%%%%%%
        \rotatebox[origin=c]{90}{\small \tb{(b)} FD-FT}%
        %%%%%%%%%%%% 5-shot %%%%%%%%%%%%%
        \begin{tabular}{ccc}
            \includegraphics[width=\ftqn]{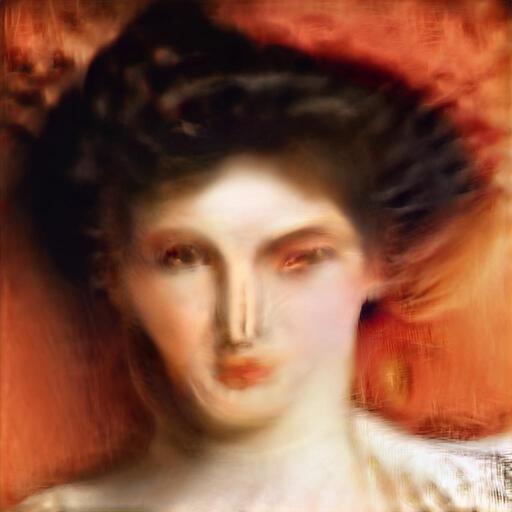} & \includegraphics[width=\ftqn]{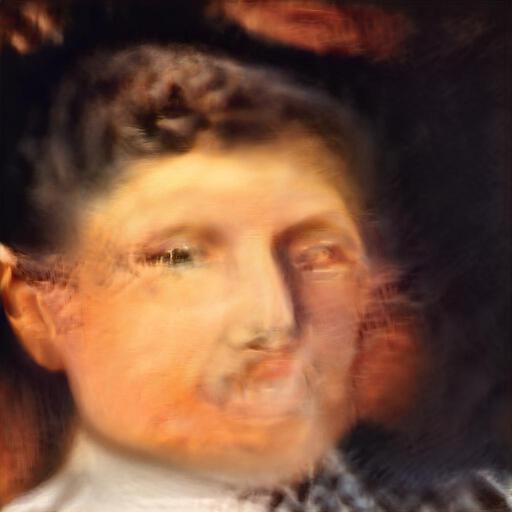} &
            \includegraphics[width=\ftqn]{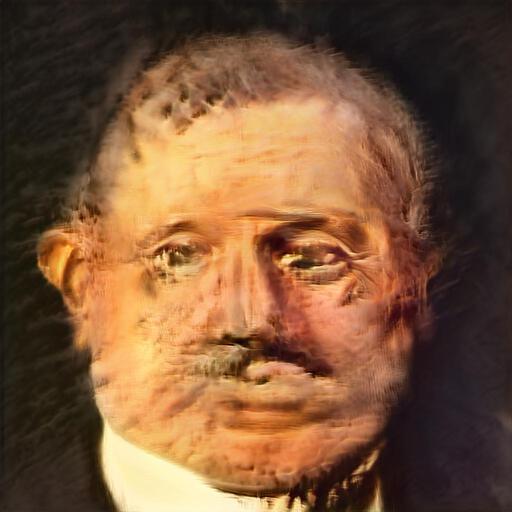} \\ \includegraphics[width=\ftqn]{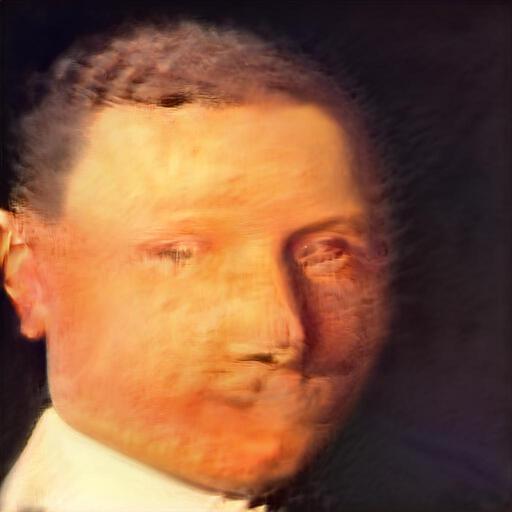} &
            \includegraphics[width=\ftqn]{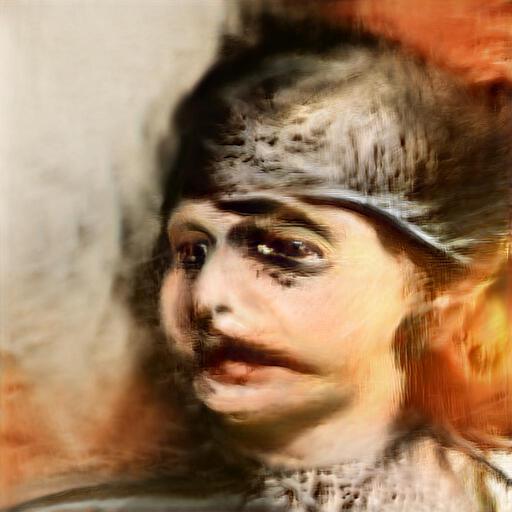} & \includegraphics[width=\ftqn]{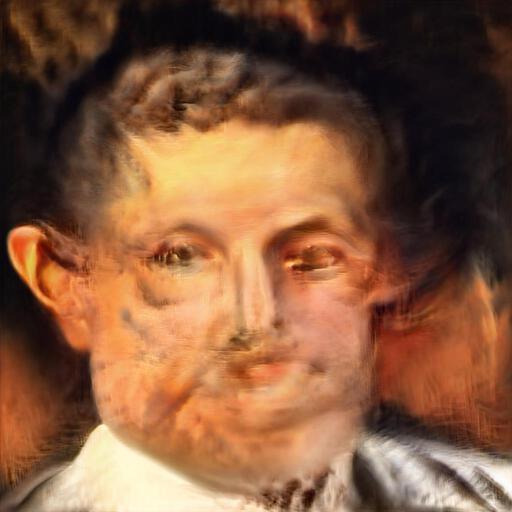}         \end{tabular}\hspace{\nspc}  & 
        %%%%%%%%%% 15-shot %%%%%%%%%%%%%%%
        \begin{tabular}{ccc}
            \includegraphics[width=\ftqn]{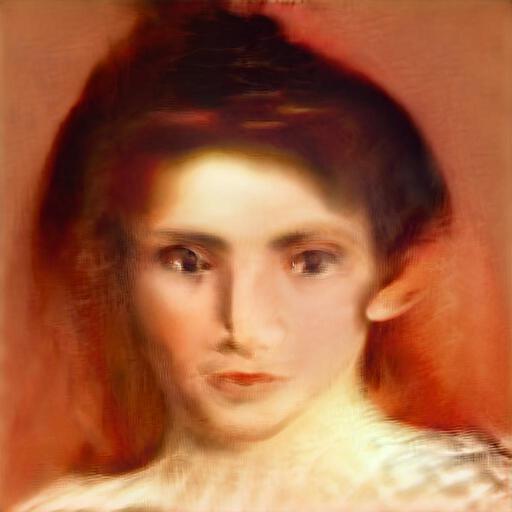} & \includegraphics[width=\ftqn]{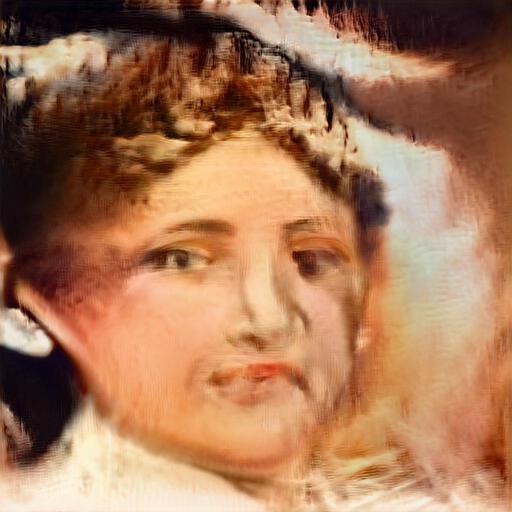} &
            \includegraphics[width=\ftqn]{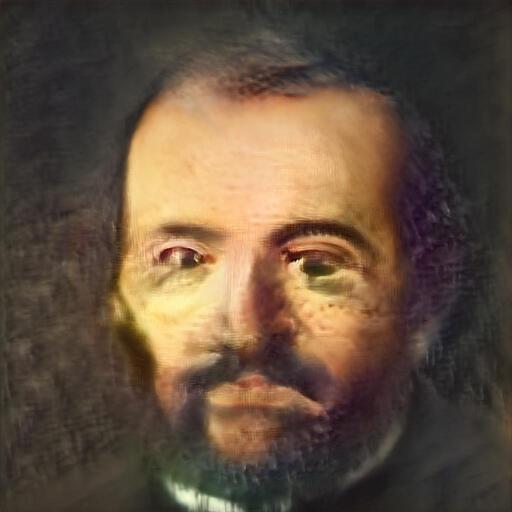} \\ \includegraphics[width=\ftqn]{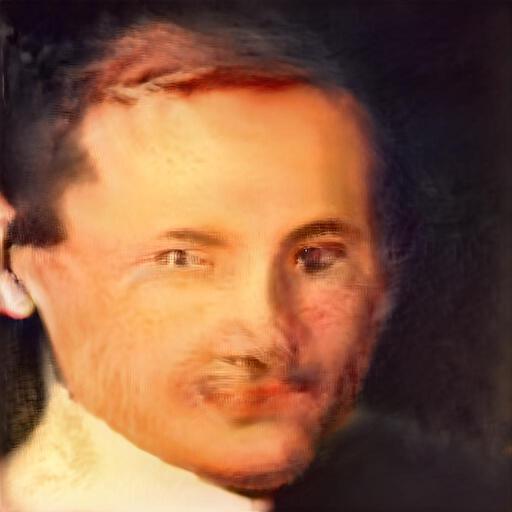} &
            \includegraphics[width=\ftqn]{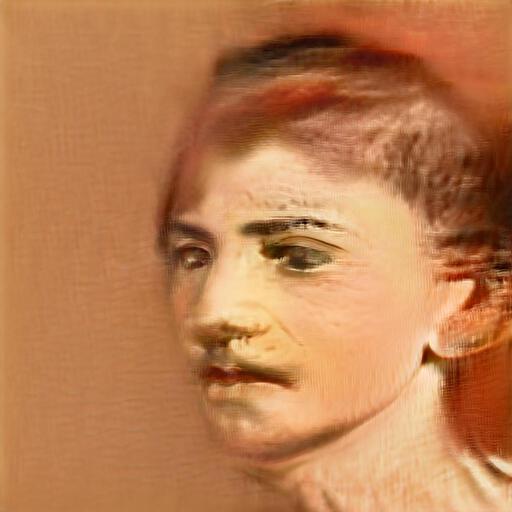} & \includegraphics[width=\ftqn]{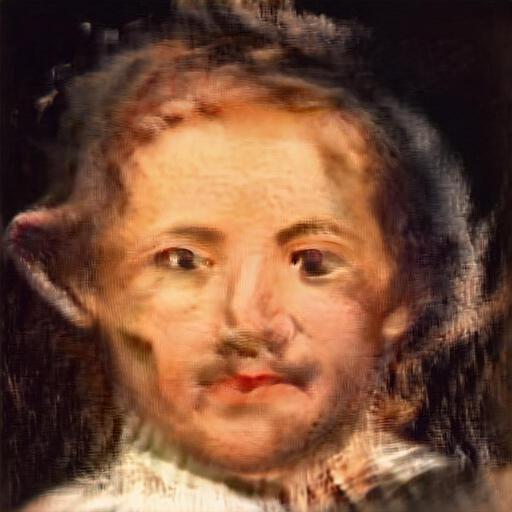}         \end{tabular}\hspace{\nspc}  & 
       %%%%%%%%%%%%% 50-shot %%%%%%%%%%%%
        \begin{tabular}{ccc}
            \includegraphics[width=\ftqn]{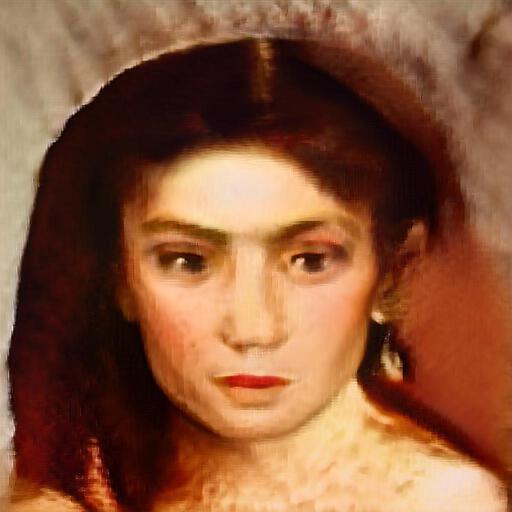} & \includegraphics[width=\ftqn]{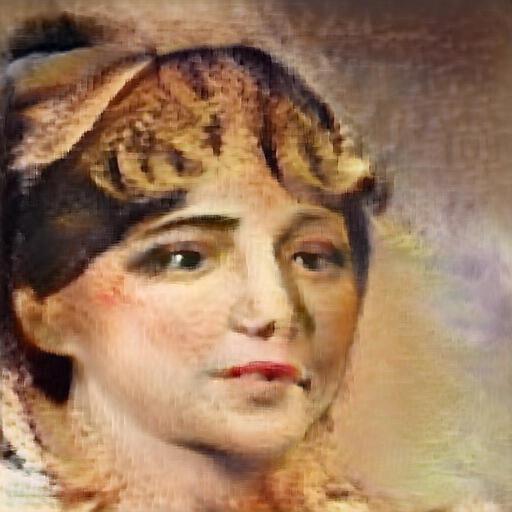} &
            \includegraphics[width=\ftqn]{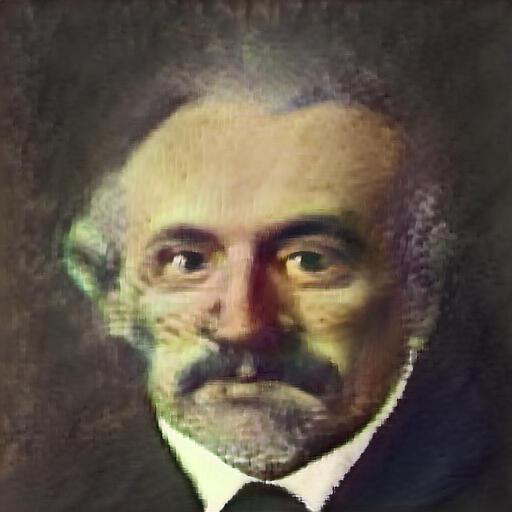} \\ \includegraphics[width=\ftqn]{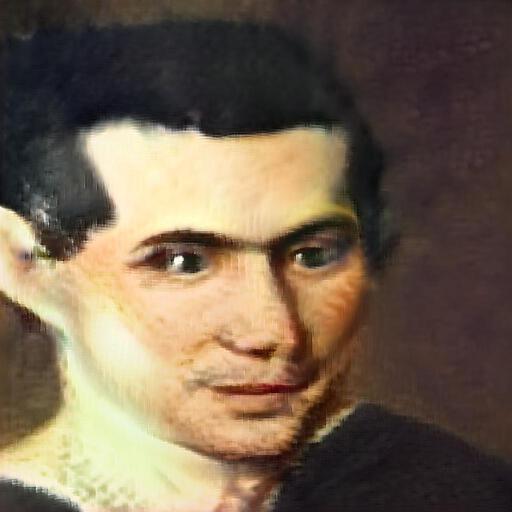} &
            \includegraphics[width=\ftqn]{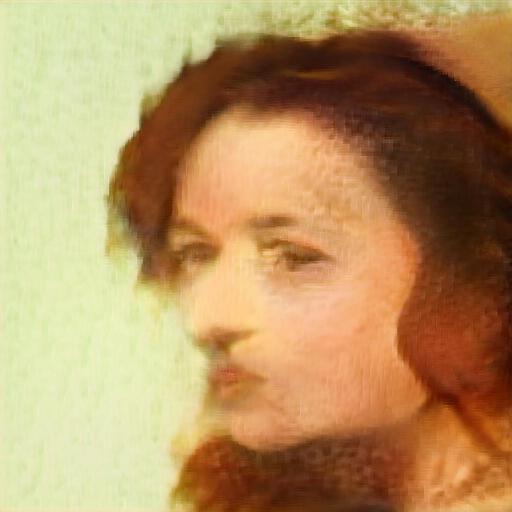} & \includegraphics[width=\ftqn]{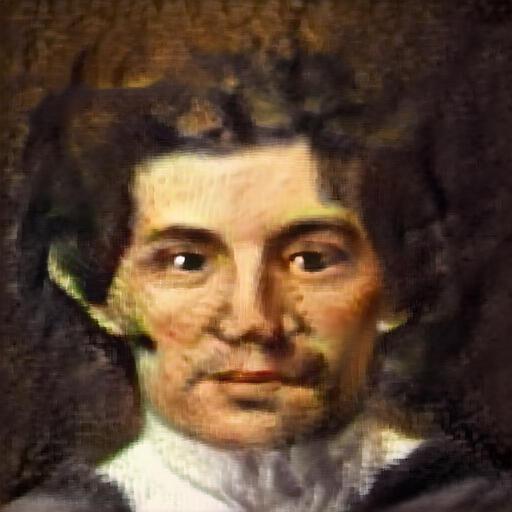} 
        \end{tabular}\hspace{\nspc}  &
       %%%%%%%%%%%%% 100-shot %%%%%%%%%%%%
        \begin{tabular}{ccc}
            \includegraphics[width=\ftqn]{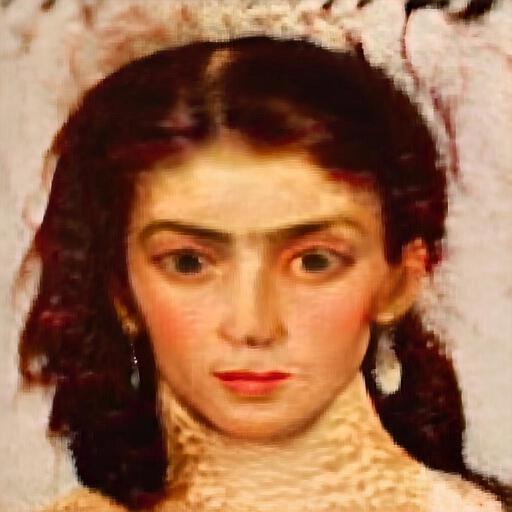} & \includegraphics[width=\ftqn]{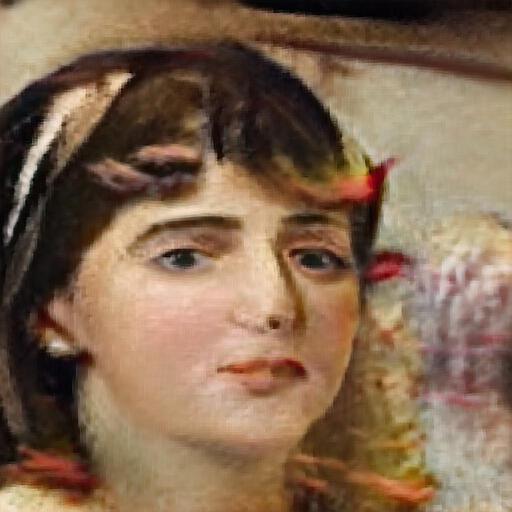} &
            \includegraphics[width=\ftqn]{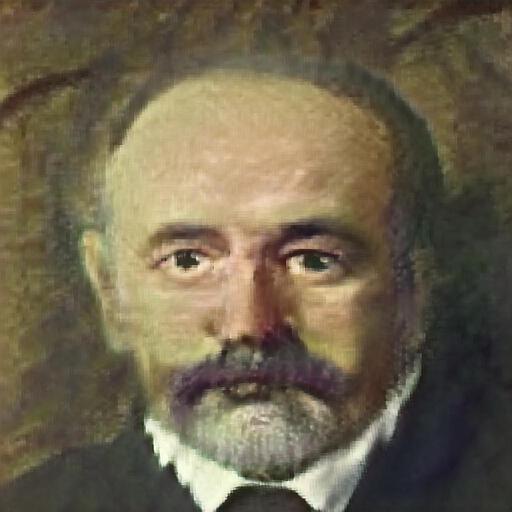} \\ \includegraphics[width=\ftqn]{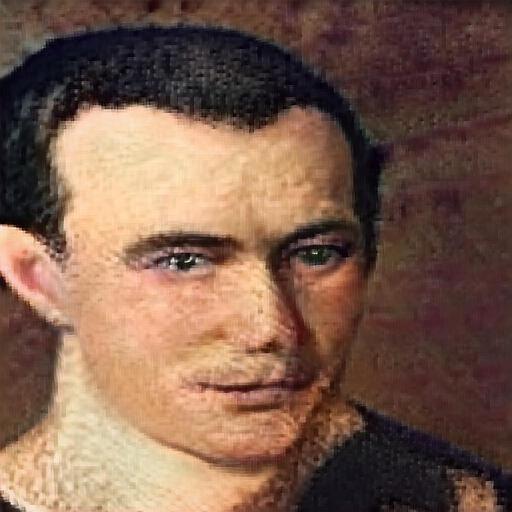}  &
            \includegraphics[width=\ftqn]{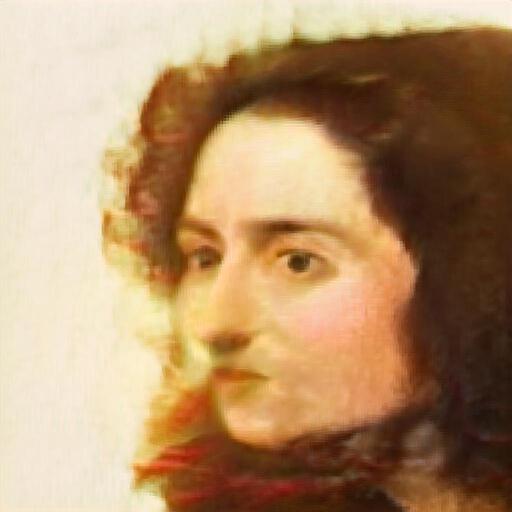} & \includegraphics[width=\ftqn]{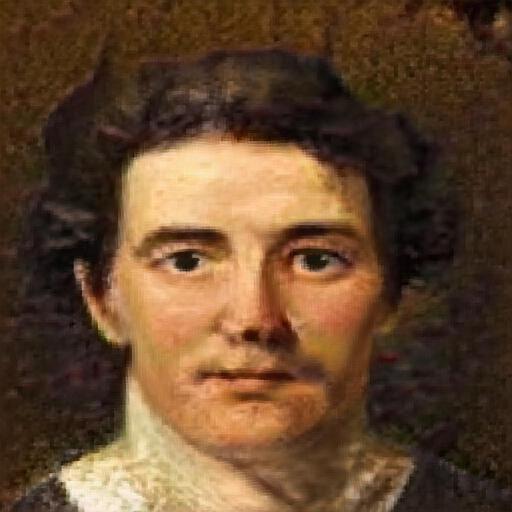}  
        \end{tabular}\vspace{\nspc} \\
        %%%%%%%%%%%%%%%%%%%%%%%%%%%%%%
        %%%%%%% FSGAN %%%%%%%%%%%%%%%%
        %%%%%%%%%%%%%%%%%%%%%%%%%%%%%%
        \rotatebox[origin=c]{90}{\small \tb{(c)} FSGAN}%
        %%%%%%%%%%%% 5-shot %%%%%%%%%%%%%
        \begin{tabular}{ccc}
            \includegraphics[width=\ftqn]{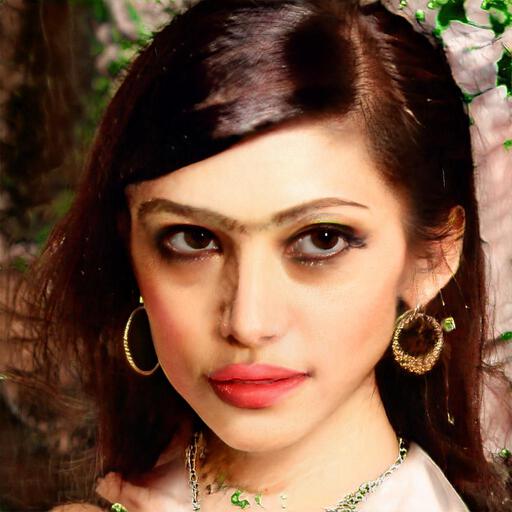} & \includegraphics[width=\ftqn]{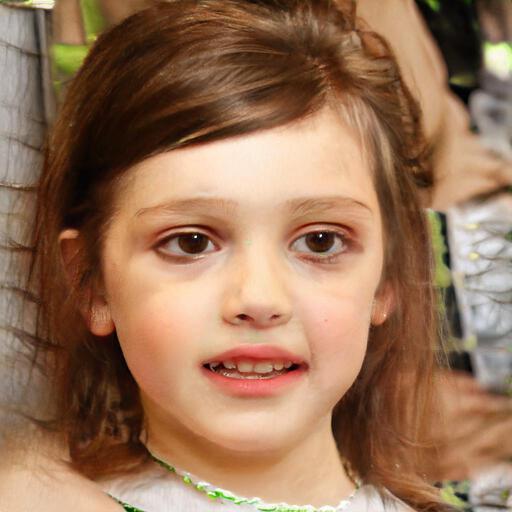} &
            \includegraphics[width=\ftqn]{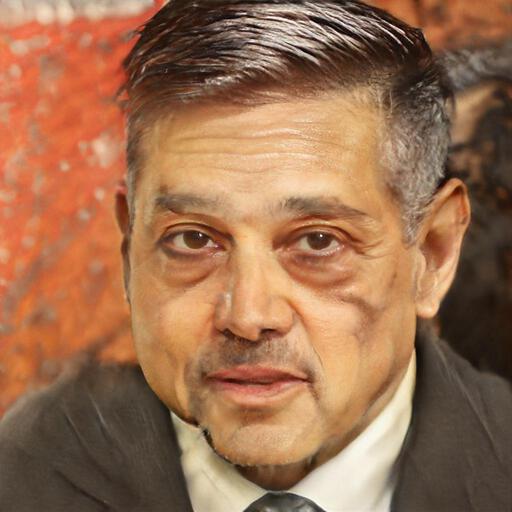} \\ \includegraphics[width=\ftqn]{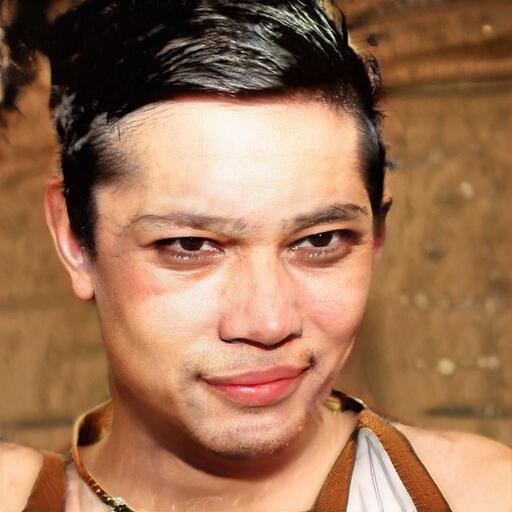} &
            \includegraphics[width=\ftqn]{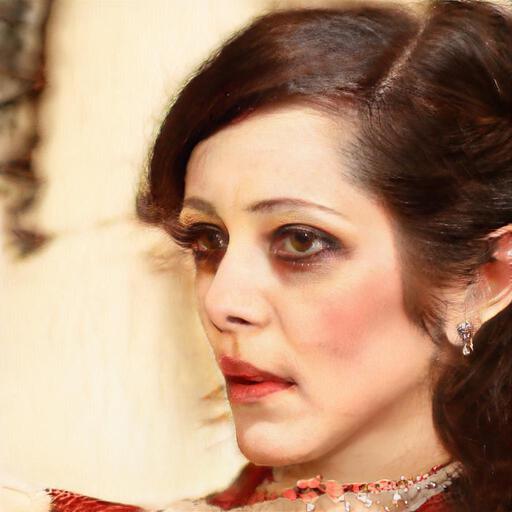} & \includegraphics[width=\ftqn]{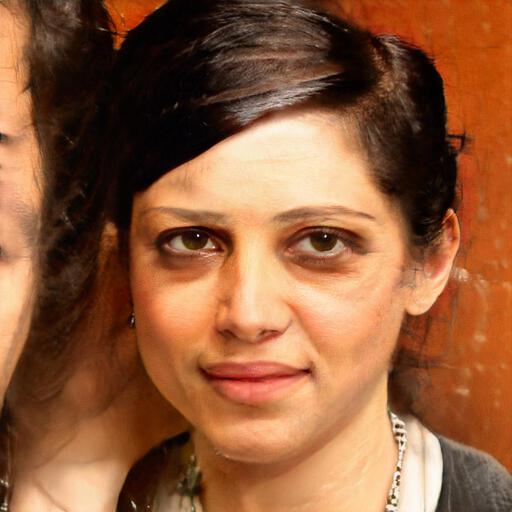}
        \end{tabular}\hspace{\nspc}  & 
        %%%%%%%%%% 15-shot %%%%%%%%%%%%%%%
        \begin{tabular}{ccc}
            \includegraphics[width=\ftqn]{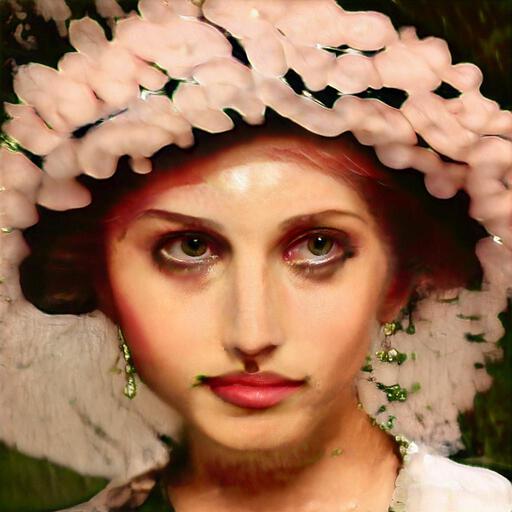} & \includegraphics[width=\ftqn]{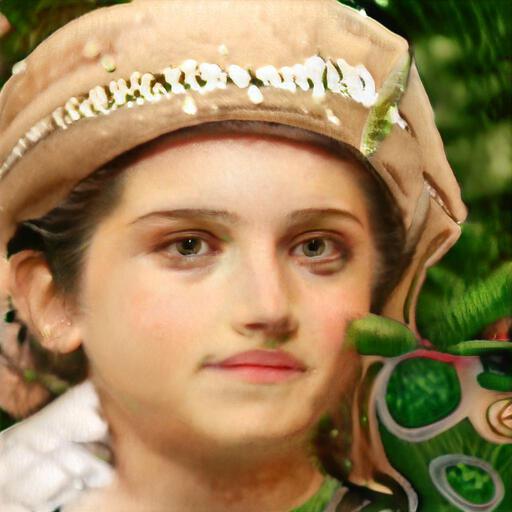} &
            \includegraphics[width=\ftqn]{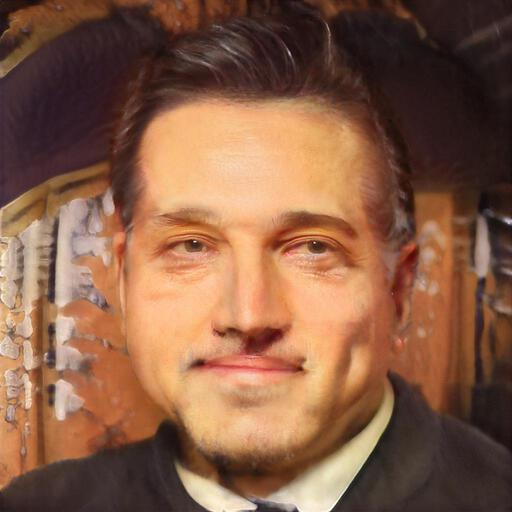} \\ \includegraphics[width=\ftqn]{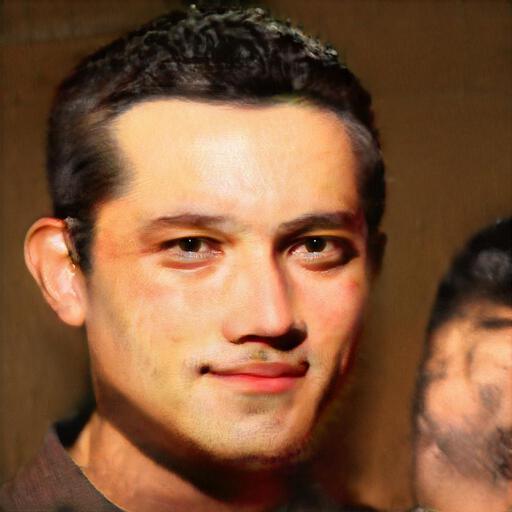} & 
            \includegraphics[width=\ftqn]{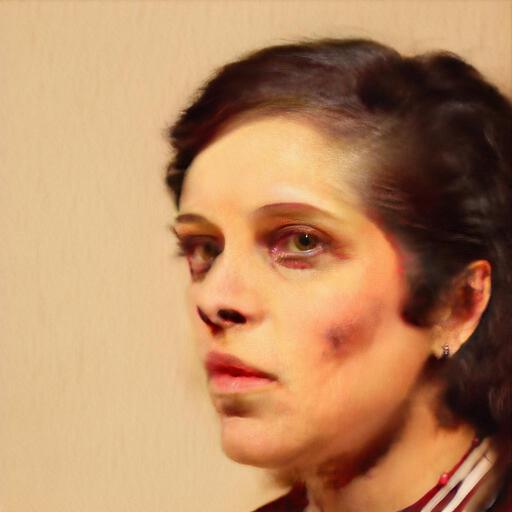} & \includegraphics[width=\ftqn]{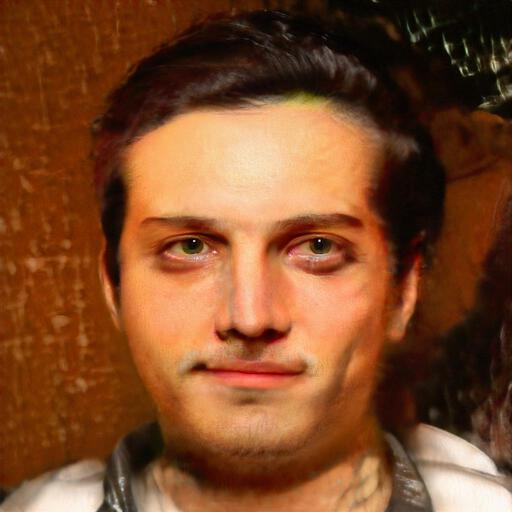}  
        \end{tabular}\hspace{\nspc}  & 
       %%%%%%%%%%%%% 50-shot %%%%%%%%%%%%
        \begin{tabular}{ccc}
            \includegraphics[width=\ftqn]{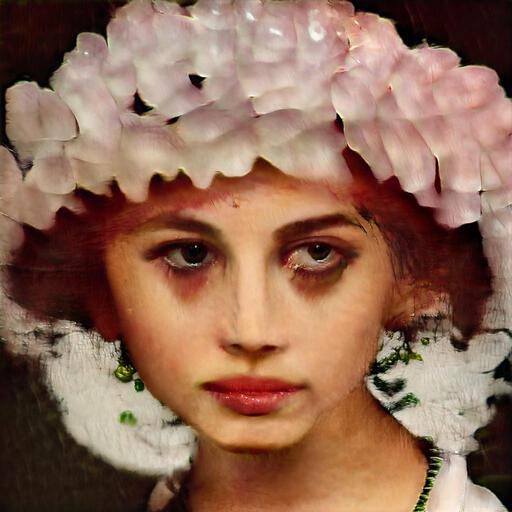} & \includegraphics[width=\ftqn]{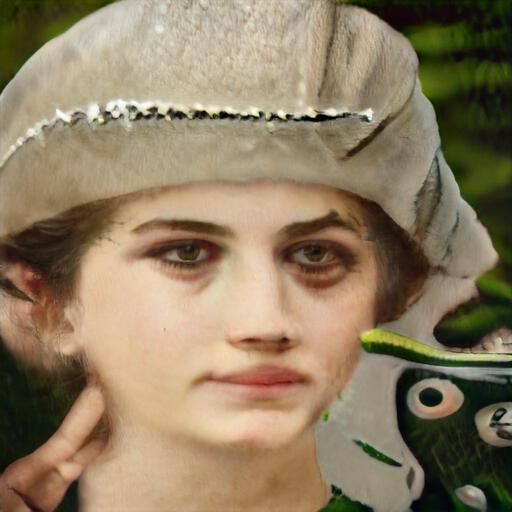} &
            \includegraphics[width=\ftqn]{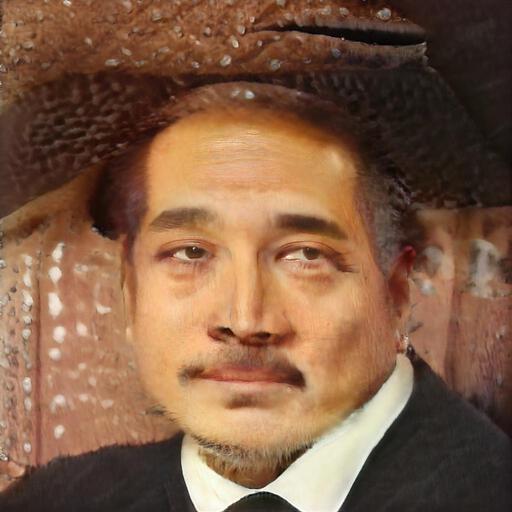} \\ \includegraphics[width=\ftqn]{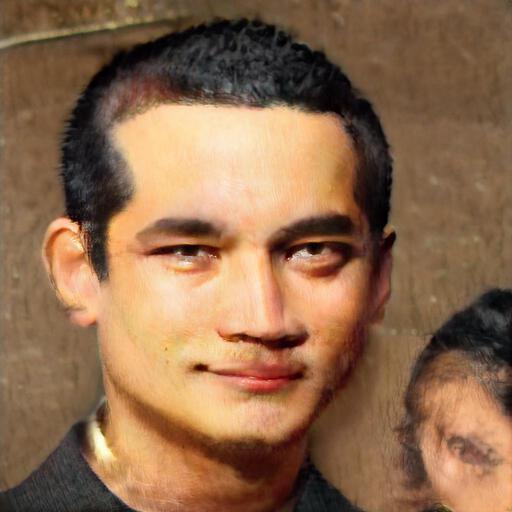} &
            \includegraphics[width=\ftqn]{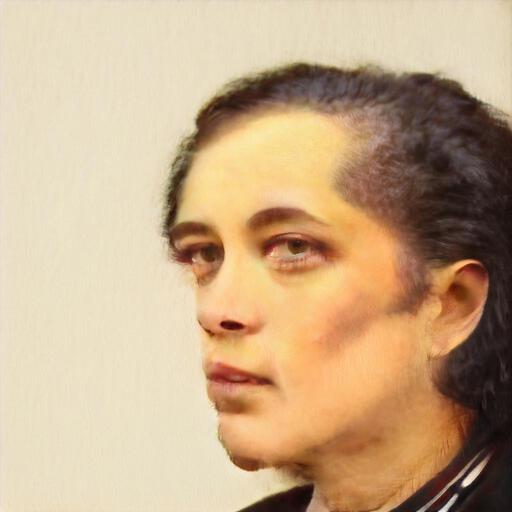} & \includegraphics[width=\ftqn]{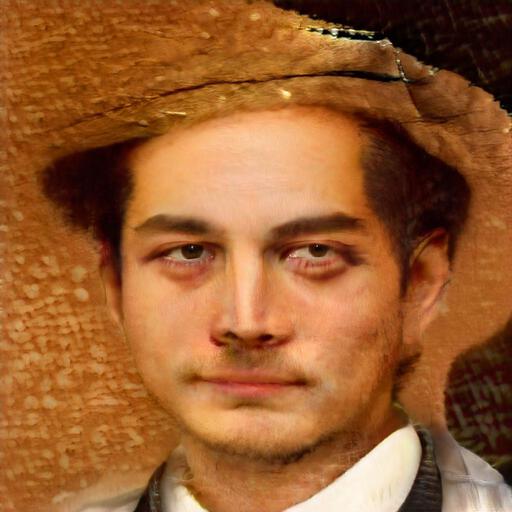} 
        \end{tabular}\hspace{\nspc}  &
       %%%%%%%%%%%%% 100-shot %%%%%%%%%%%%
        \begin{tabular}{ccc}
            \includegraphics[width=\ftqn]{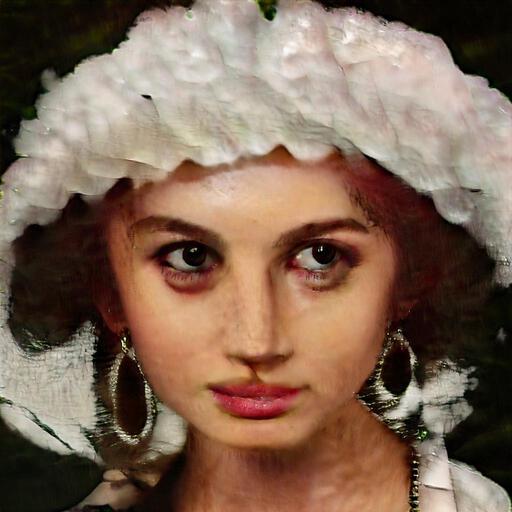} & \includegraphics[width=\ftqn]{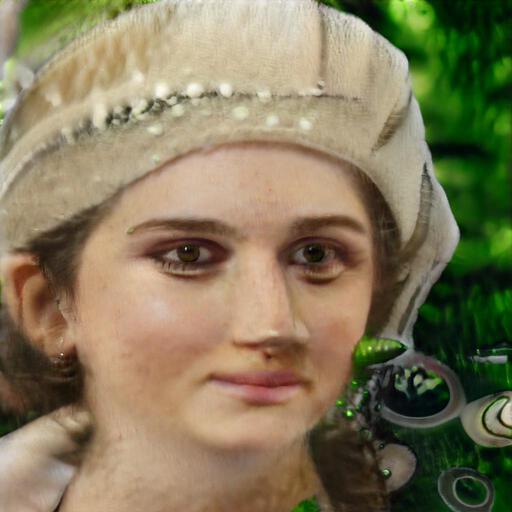} &
            \includegraphics[width=\ftqn]{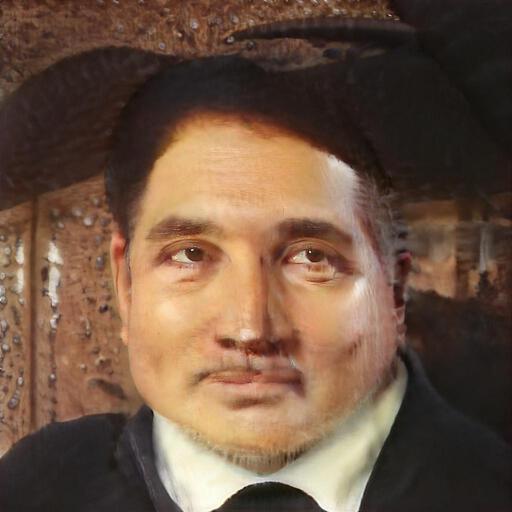} \\ \includegraphics[width=\ftqn]{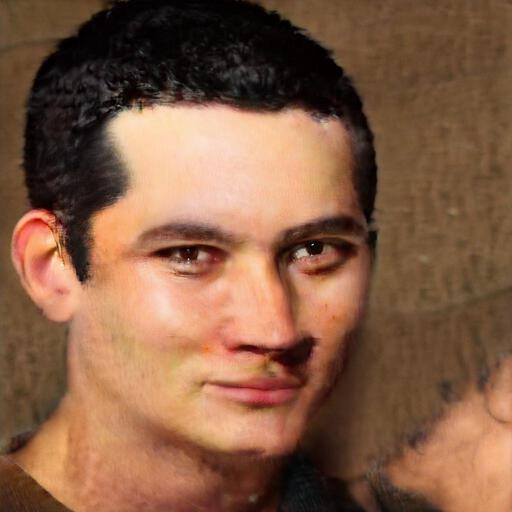} &
            \includegraphics[width=\ftqn]{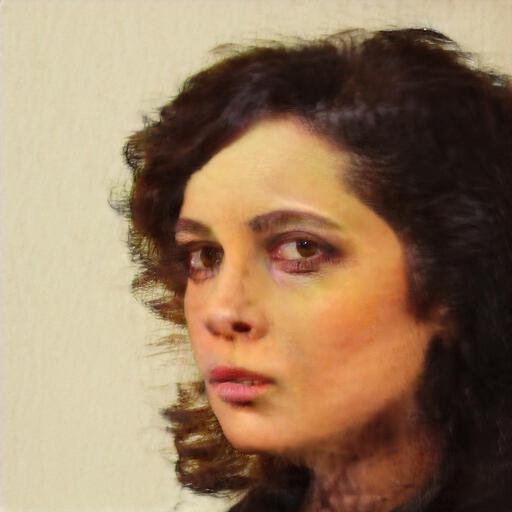} & \includegraphics[width=\ftqn]{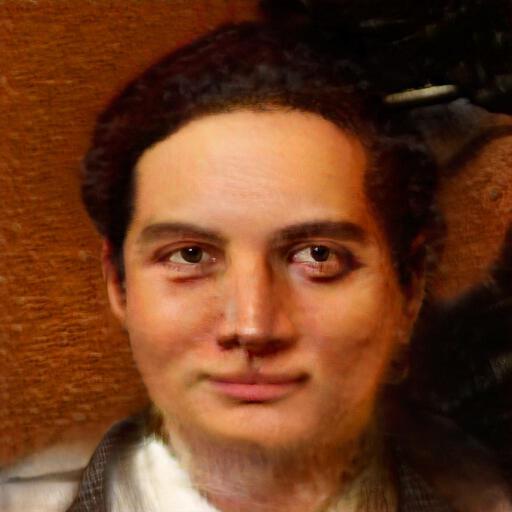} 
        \end{tabular} \\
    \end{tabular}
    
    \vspace{3pt}
    \begin{minipage}[t]{0.25\textwidth}
        \rotatebox[origin=c]{90}{\small \tb{(d)} Pretrain}%
        \begin{tabular}{ccc}
            \includegraphics[width=\ftqn]{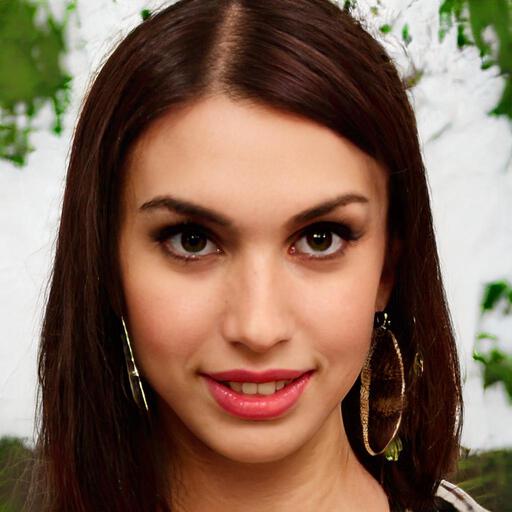} & \includegraphics[width=\ftqn]{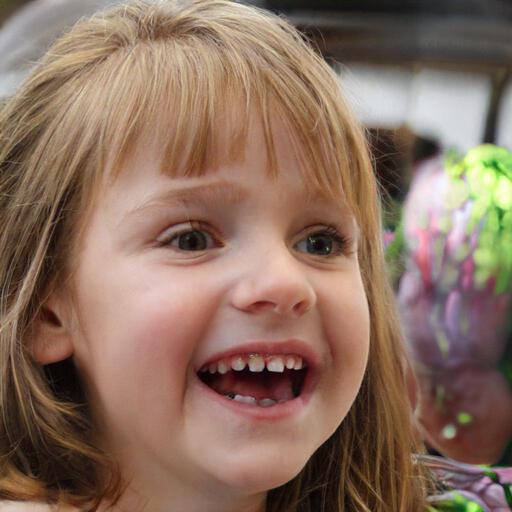} &
            \includegraphics[width=\ftqn]{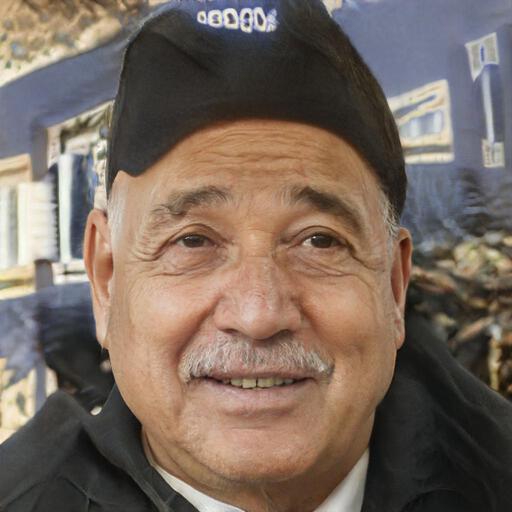} \\ \includegraphics[width=\ftqn]{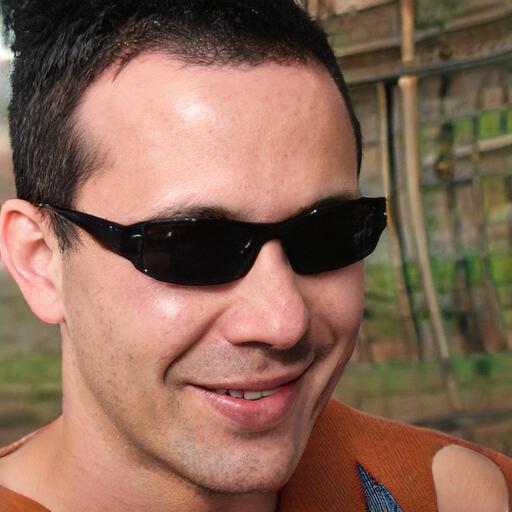} & 
            \includegraphics[width=\ftqn]{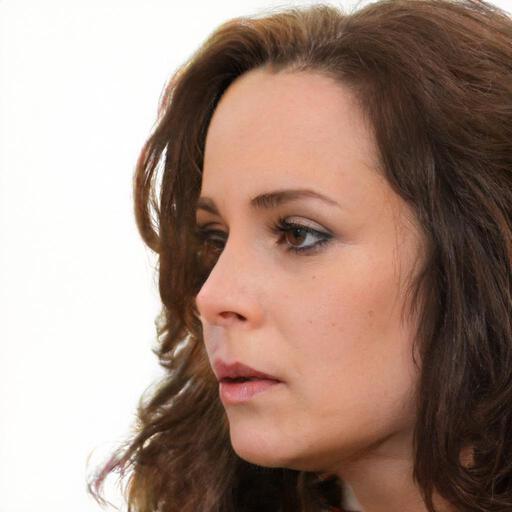} & \includegraphics[width=\ftqn]{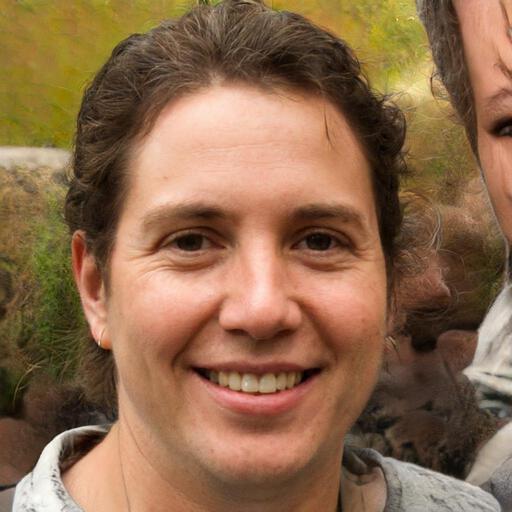} 
        \end{tabular}
    \end{minipage}
    \hfill
    \begin{minipage}[t]{0.73\textwidth}
        \vspace{-31pt}
        \caption{
            \textbf{N-shot settings} (FFHQ$\rightarrow$Portraits):
            %Comparing alternative methods with matched latent codes in 5-to-100-shot settings:
            \tb{(a)} \citet{mo2020freeze} with limited timesteps preserves diversity at all n-shots, but produces undesired artifacts and limited adaptation ({\em e.g.} sunglasses remain).
            \tb{(b)} \citet{mo2020freeze} with increased timesteps produces quality adaptation with 100 shots, but degenerates at $\le$50 shots.
            \tb{(c)} FSGAN (ours) is robust to n-shot settings, producing high-quality adaptation even at N=5. \\
            \tb{(d)} Pretrained FFHQ images.
        \label{fig:nshot}
        }
    \end{minipage}
\end{figure}

{\bf Adaptation Methods.}
We compare the proposed FSGAN with Transfer GAN (TGAN) \citep{wang2018eccv}, FreezeD (FD) \citep{mo2020freeze}, and the Scale \& Shift GAN (SSGAN) baseline of \citet{noguchi2019iccv}.
For a fair comparison in the GAN setting, we choose the GAN baseline of SSGAN \citep{noguchi2019iccv} instead of their GLO-based variant. 
We implement all methods using the StyleGAN2 \citep{karras2019analyzing} codebase.\footnote{\hyperlink{https://github.com/NVlabs/stylegan2}{https://github.com/NVlabs/stylegan2}}
We follow the training setting of StyleGAN, but change the learning rate to 0.003 to stabilize training and reduce the number of training steps to prevent overfitting in the low-shot setting. 
Figures \ref{fig:metric-problem}, \ref{fig:nshot} show comparisons of different training times.
%To prevent overfitting or distortion, we empirically limit training timesteps by trying our best picking the number of iterations that worked the best for all methods. 
%\vincent{what is distortion?? looks like another new concept we are adding to discussion}
%We detail this in the appendix.
%\vincent{Is this Fig 8?  Refer to it explicitly}

%We empirically pick the number of iterations that works the best for all methods.  However, in the low-shot setting, we observed alternative methods produce less satisfactory results at any iteration."

% CelebA single-ID transfer
\begin{figure*}[th]
    \centering\small
    \newlength\ftqc\setlength\ftqc{1.92cm}
    \newlength\ftqd\setlength\ftqd{3.8cm}
    \setlength\tabcolsep{0.1pt}
    \mpage{0.02}\hfill
    \mpage{0.25}{Target Images} \hfill
    \mpage{0.13}{Pretrain} \hfill
    \mpage{0.13}{TGAN} \hfill %\citet{wang2018eccv} 
    \mpage{0.12}{FD} \hfill %\citet{mo2020freeze}
    \mpage{0.12}{SSGAN} \hfill % \citet{noguchi2019iccv}
    \mpage{0.13}{FSGAN}\hfill
    %\vspace{0.1mm}
    \mpage{0.01}{\rotatebox[origin=c]{90}{\small {\bf (a)} CelebA 4978 (31-shot)}}
    \mpage{0.26}{\includegraphics[height=\ftqd]{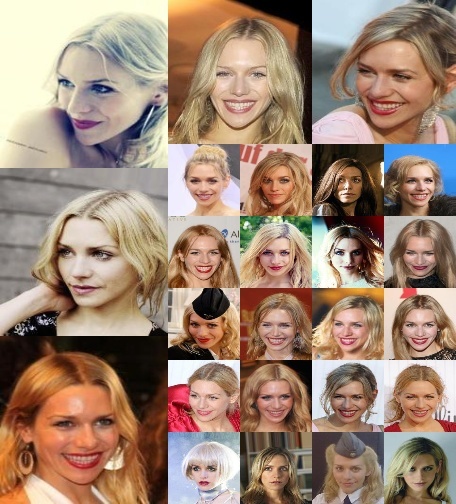}}
    \mpage{0.69}{
        \includegraphics[width=\ftqc]{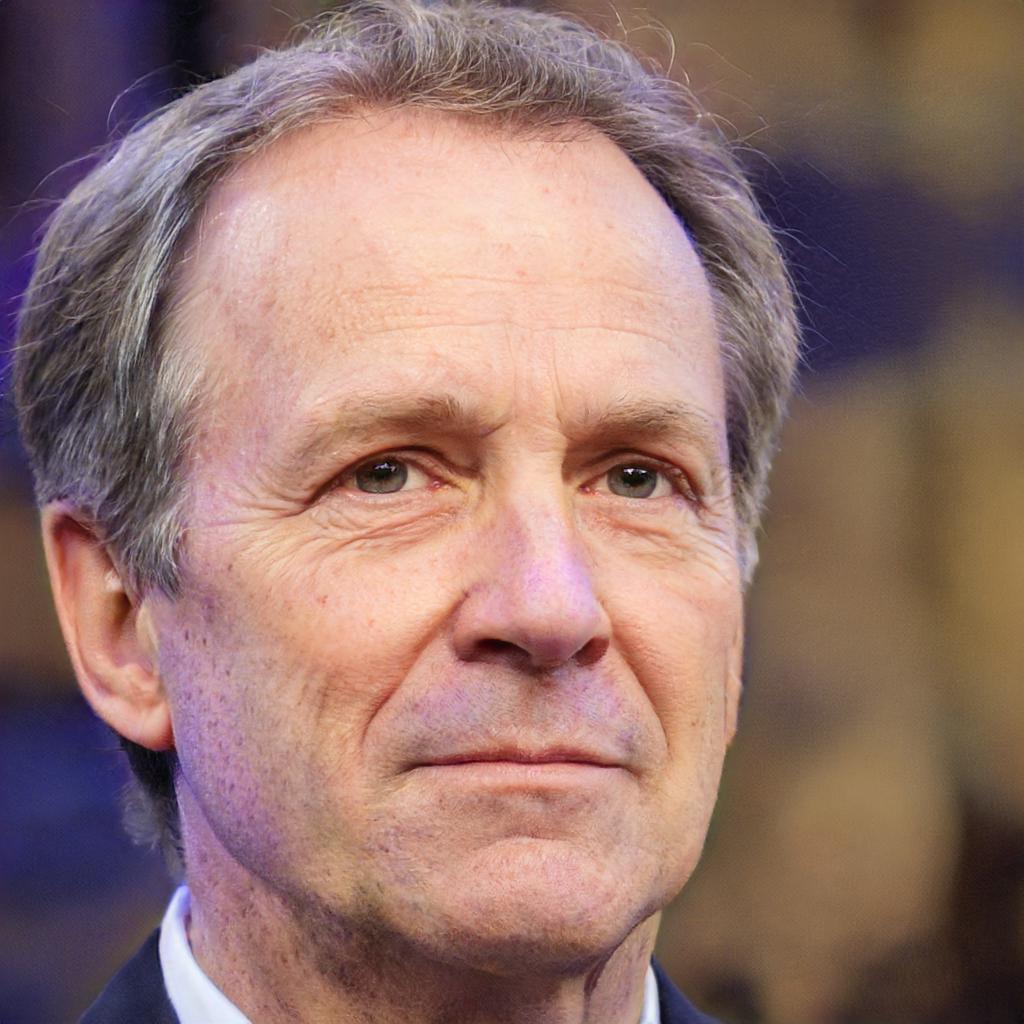}\hfill%
        \includegraphics[width=\ftqc]{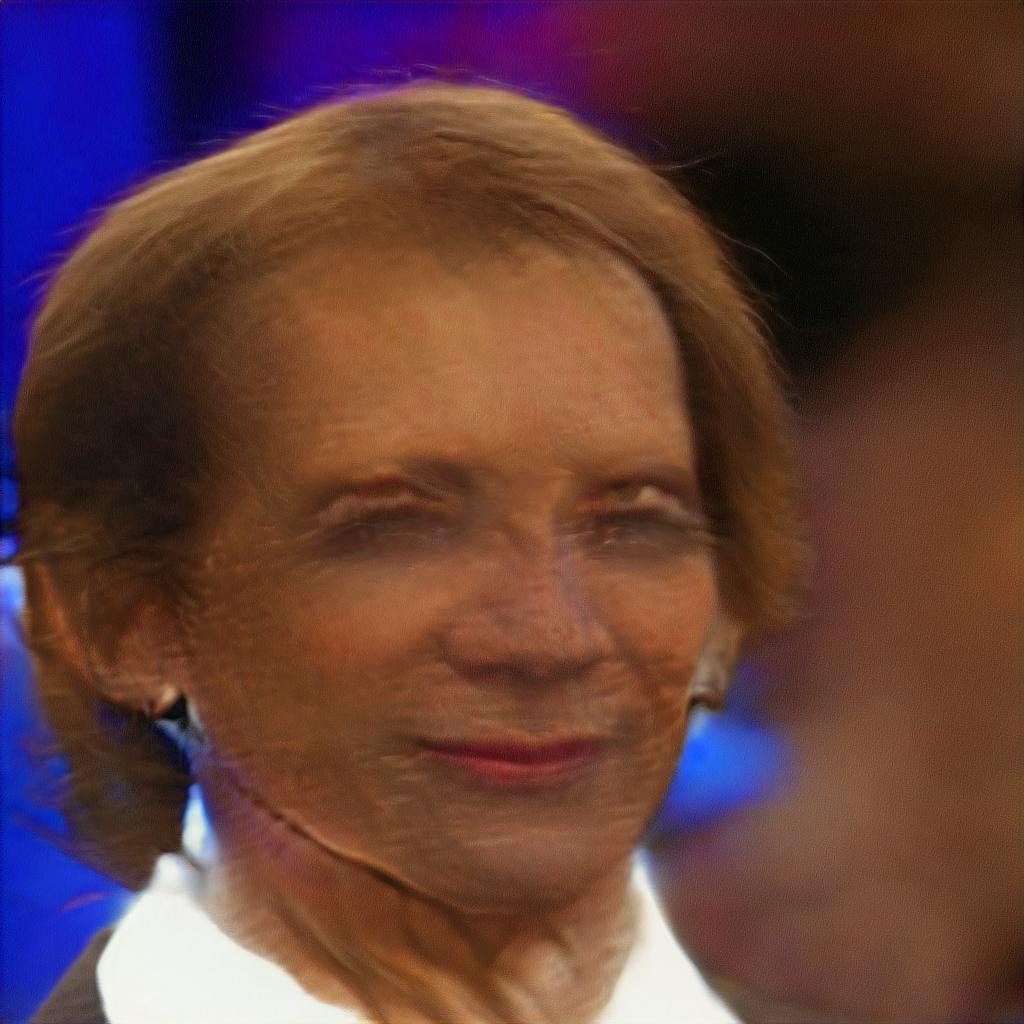}\hfill%
        \includegraphics[width=\ftqc]{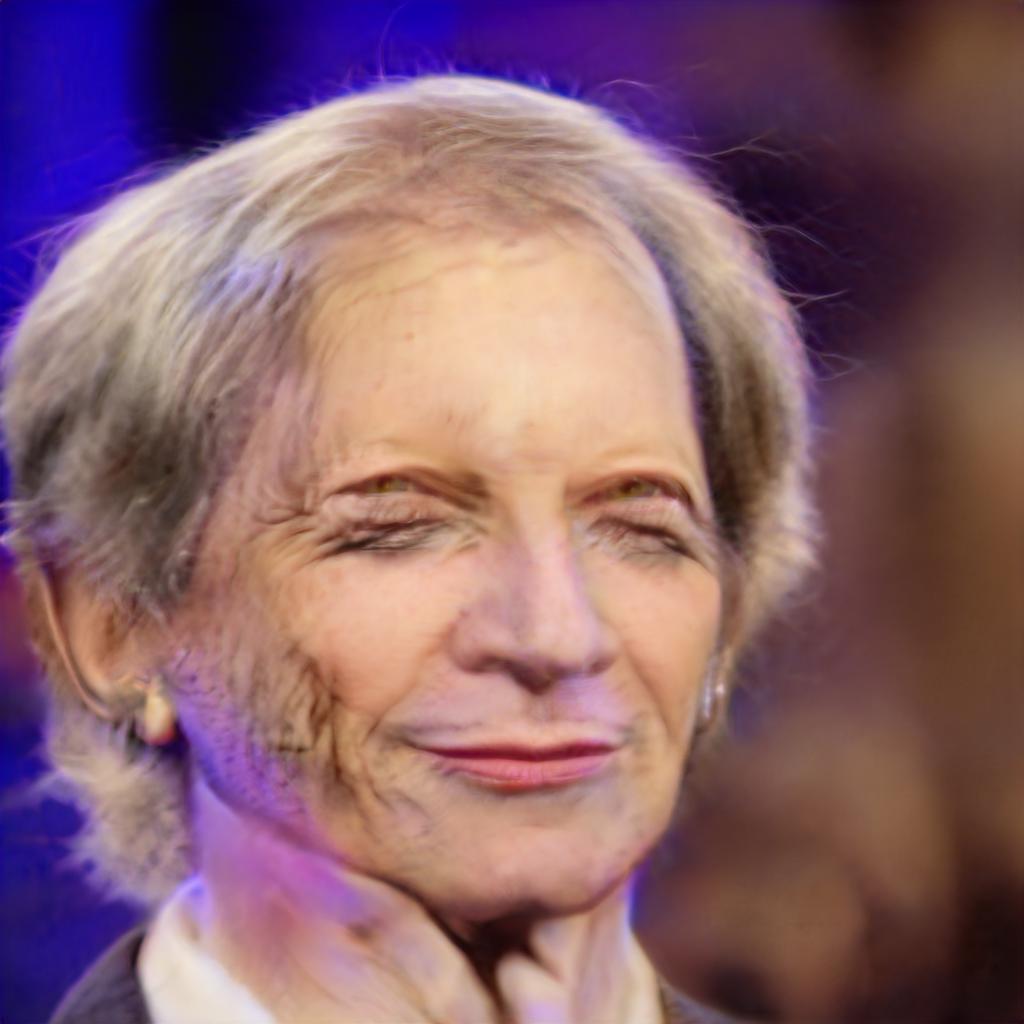}\hfill%
        \includegraphics[width=\ftqc]{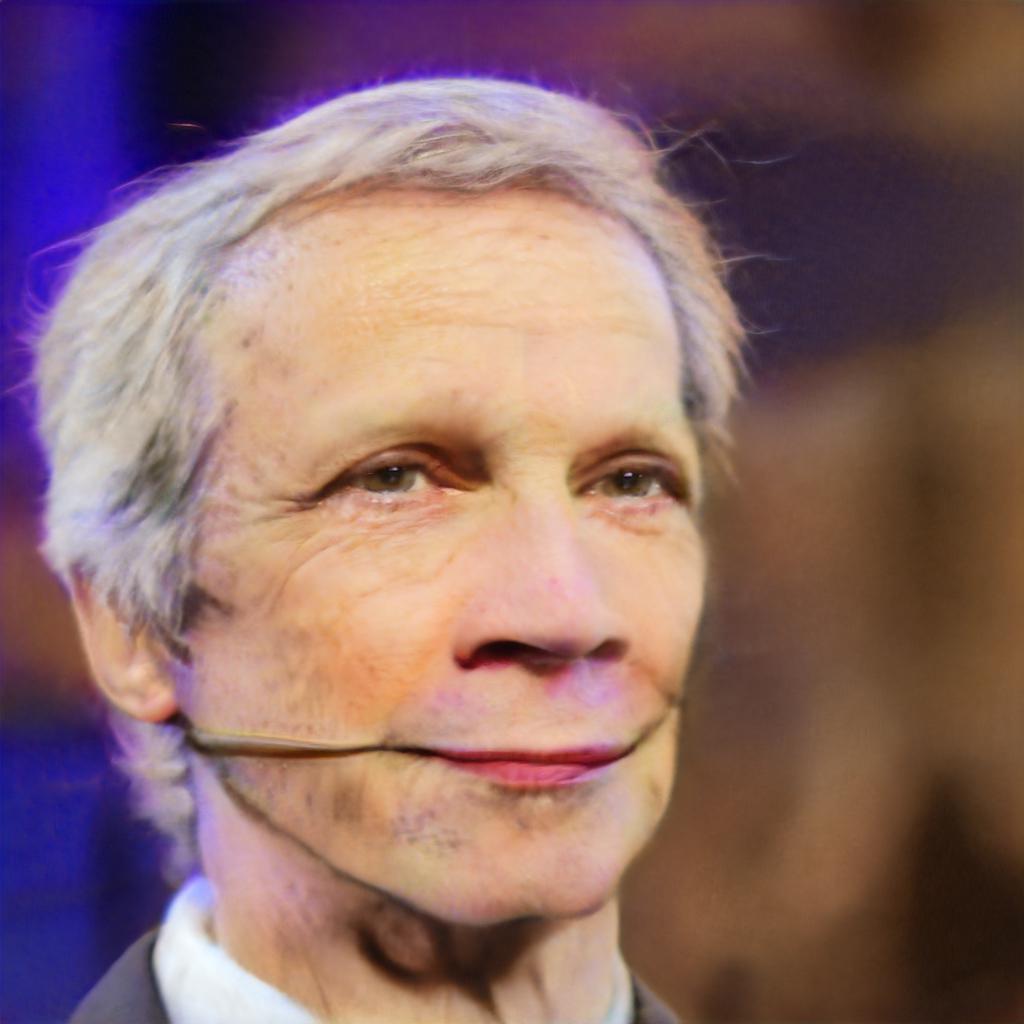}\hfill%
        \includegraphics[width=\ftqc]{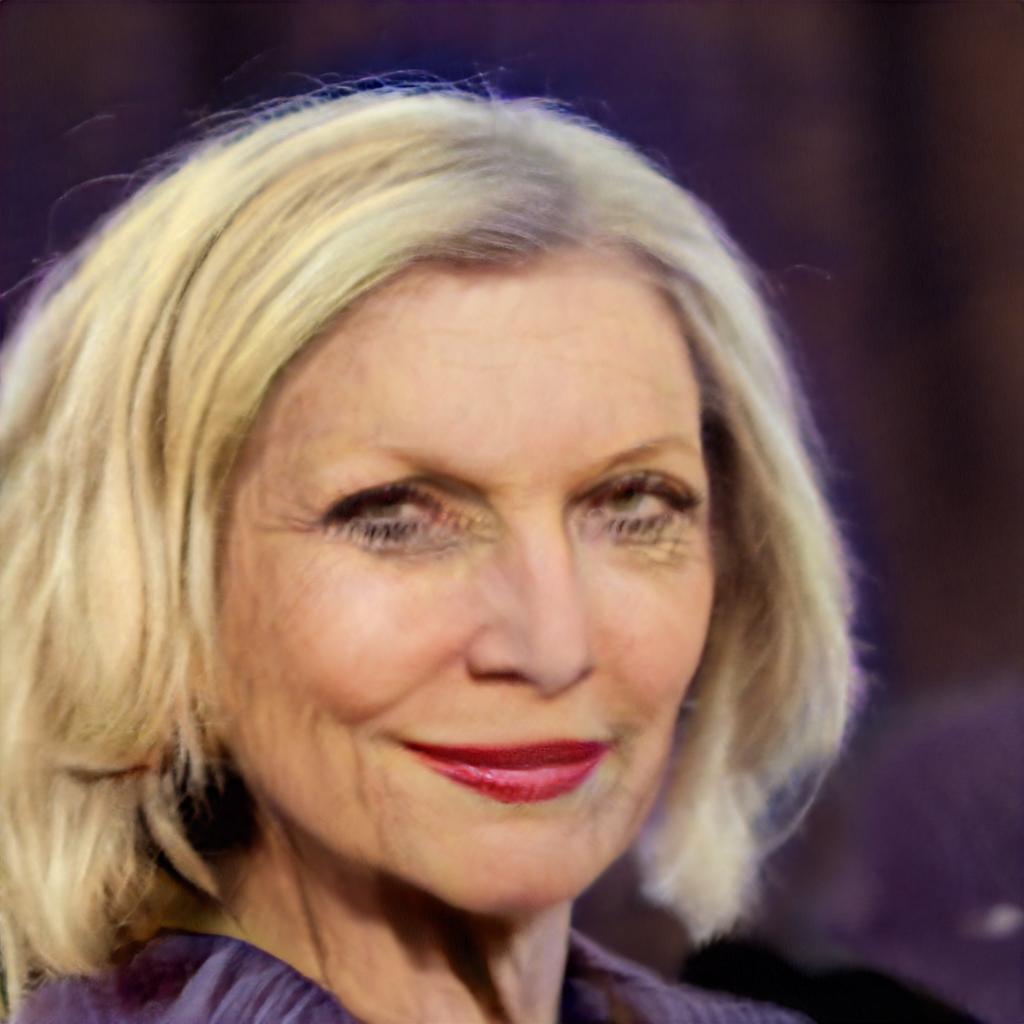}\hfill
        \\
        \includegraphics[width=\ftqc]{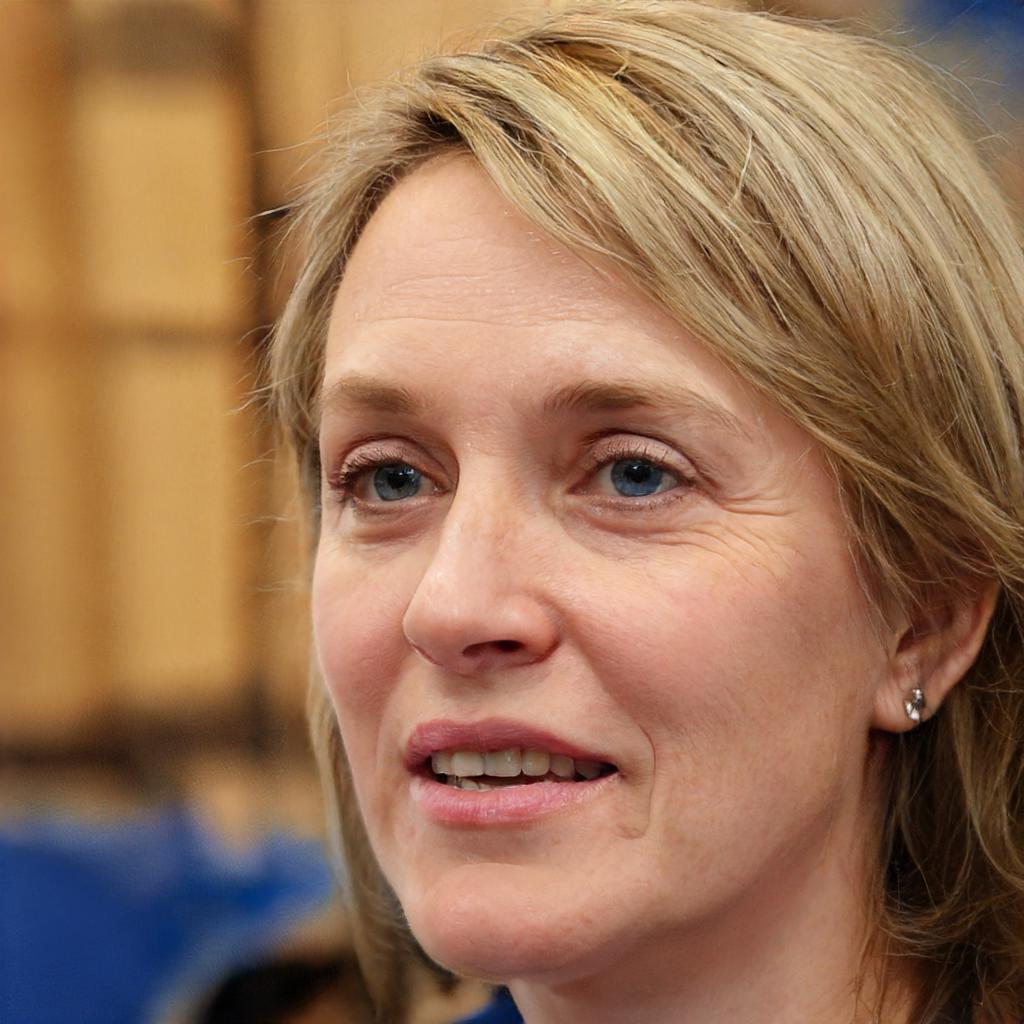}\hfill%
        \includegraphics[width=\ftqc]{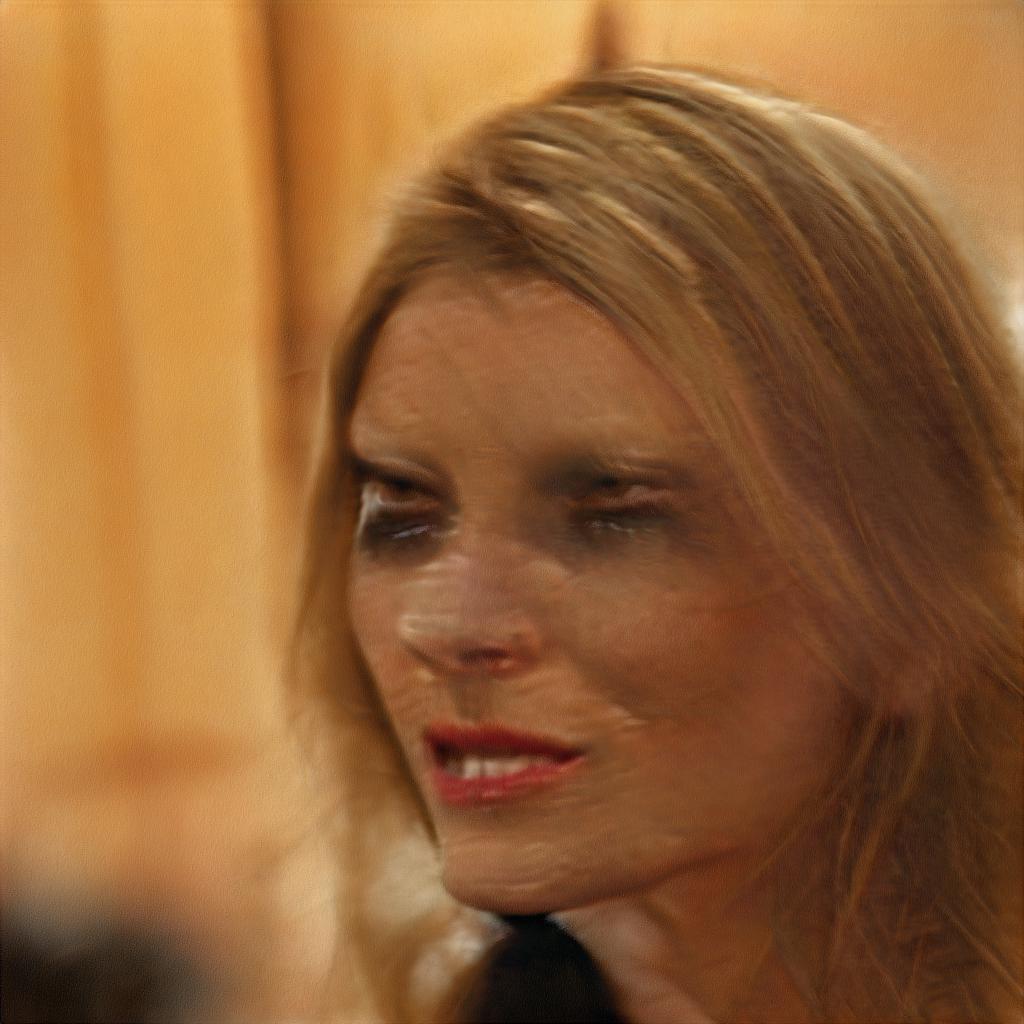}\hfill%
        \includegraphics[width=\ftqc]{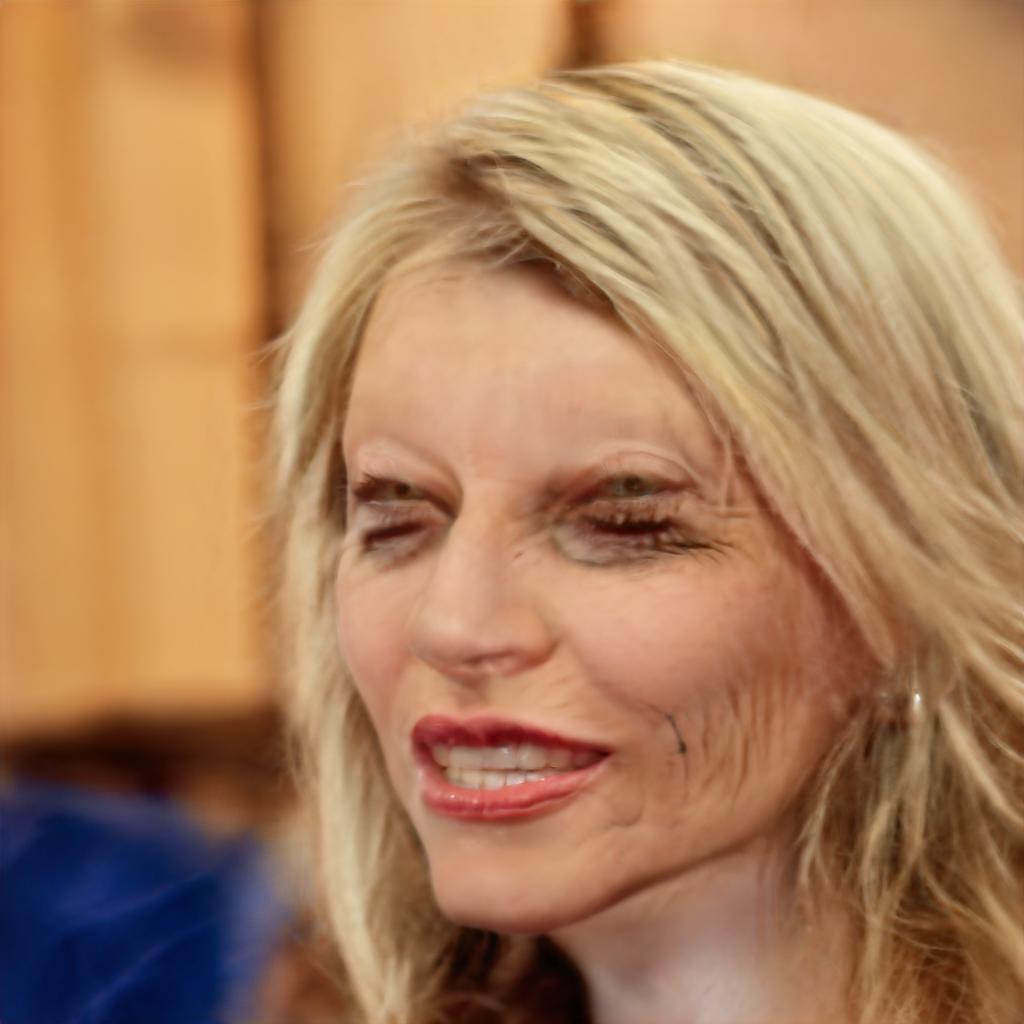}\hfill%
        \includegraphics[width=\ftqc]{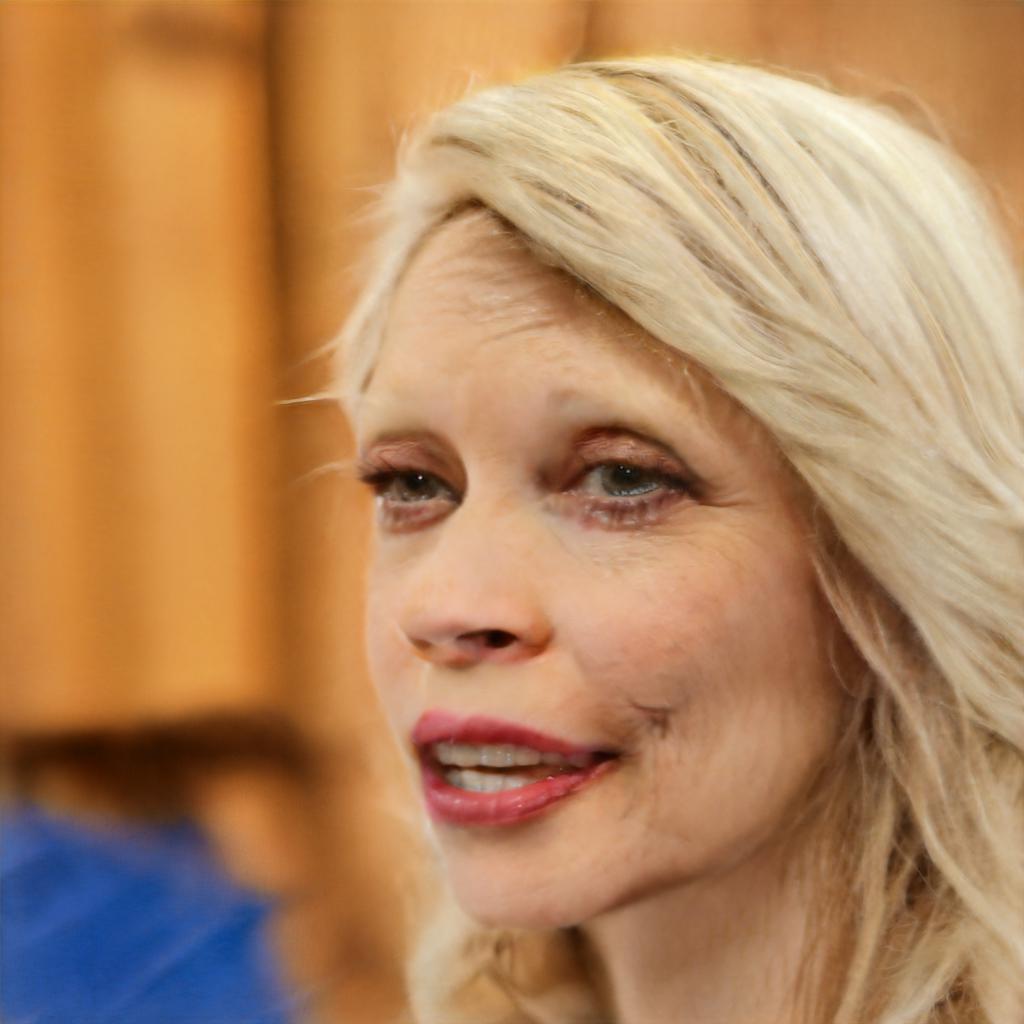}\hfill%
        \includegraphics[width=\ftqc]{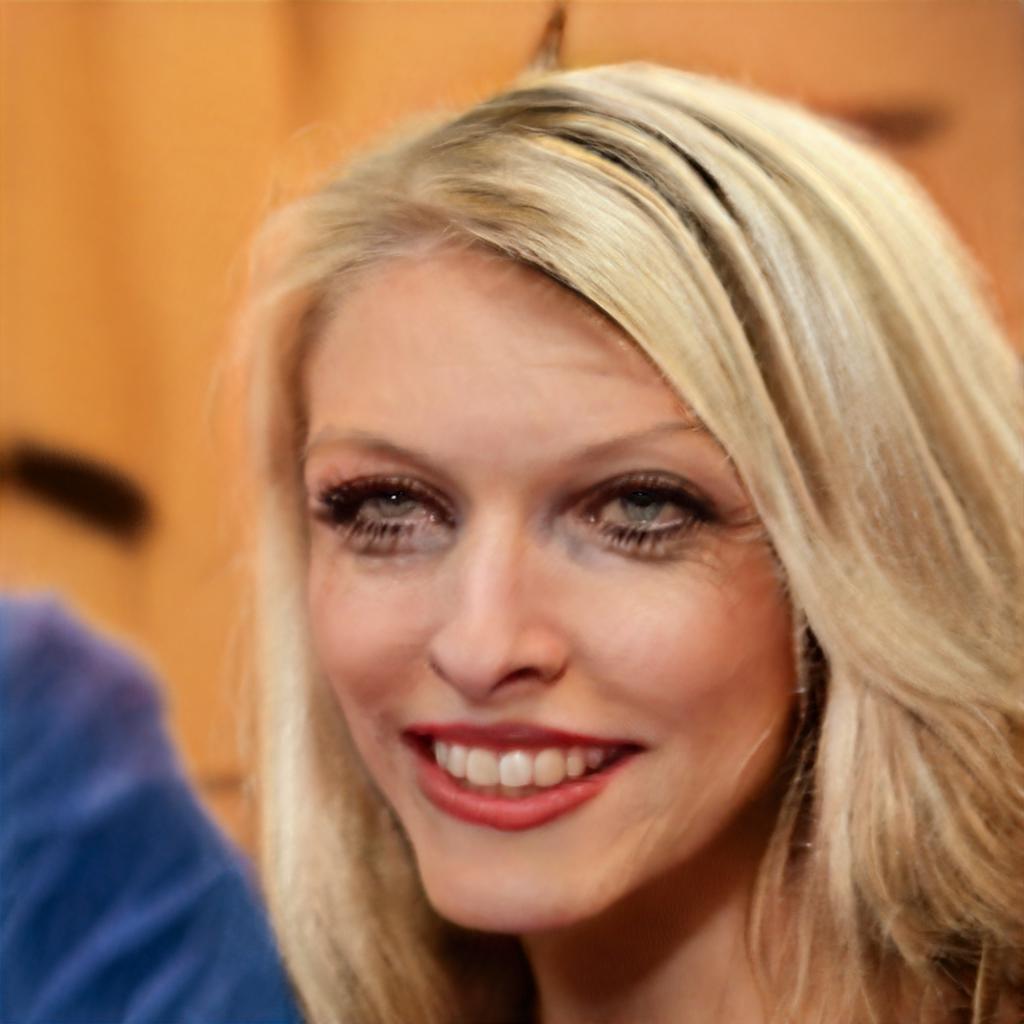}\hfill
    }
    \mpage{0.01}{\rotatebox[origin=c]{90}{\small {\bf \tb{(b)}} CelebA 3719 (30-shot)}}
    \mpage{0.26}{\includegraphics[height=\ftqd]{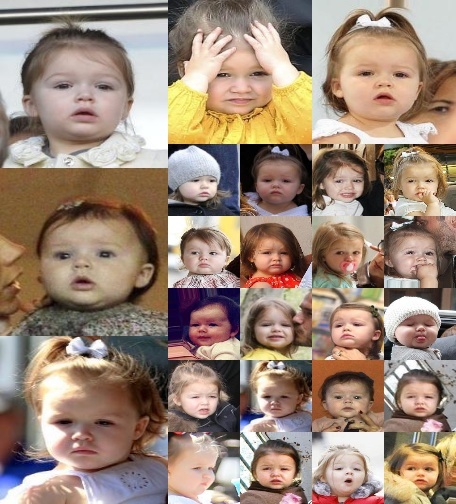}}
    \mpage{0.69}{
        \includegraphics[width=\ftqc]{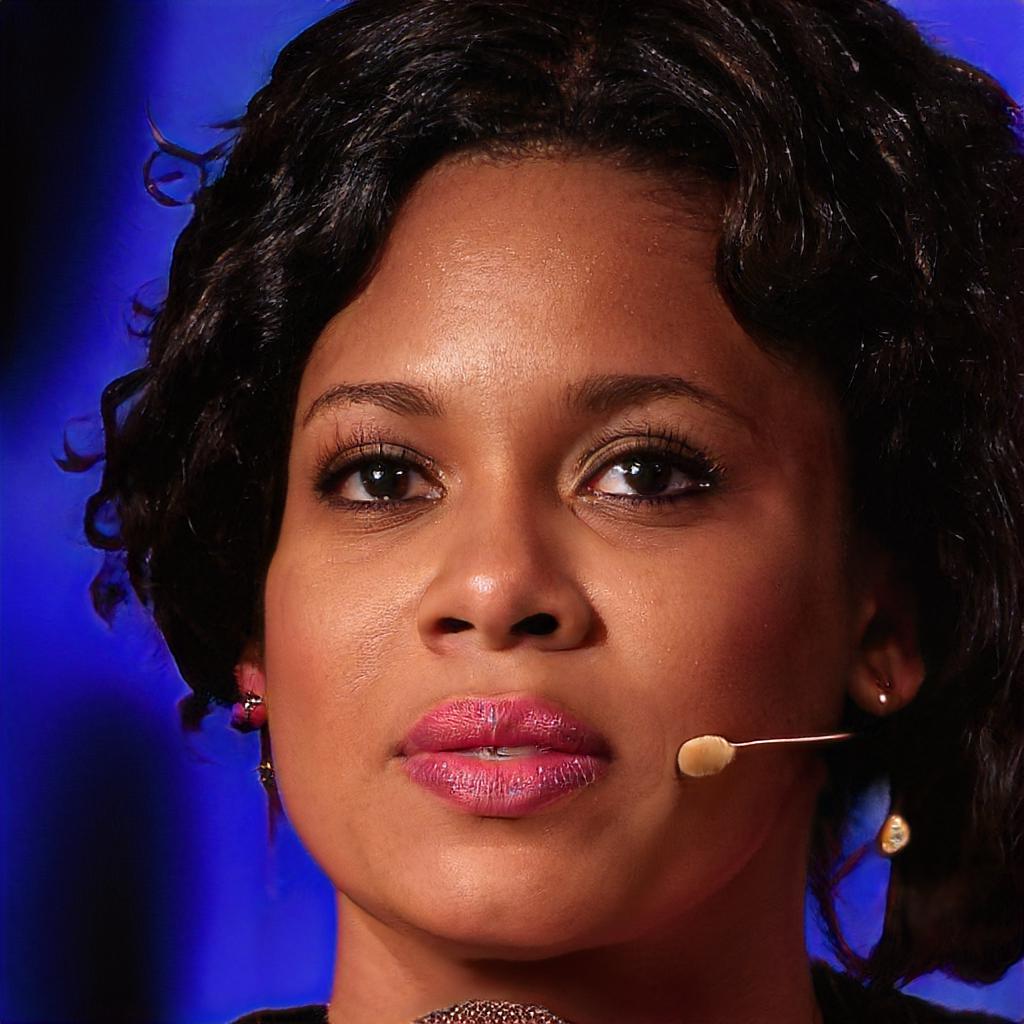}\hfill%
        \includegraphics[width=\ftqc]{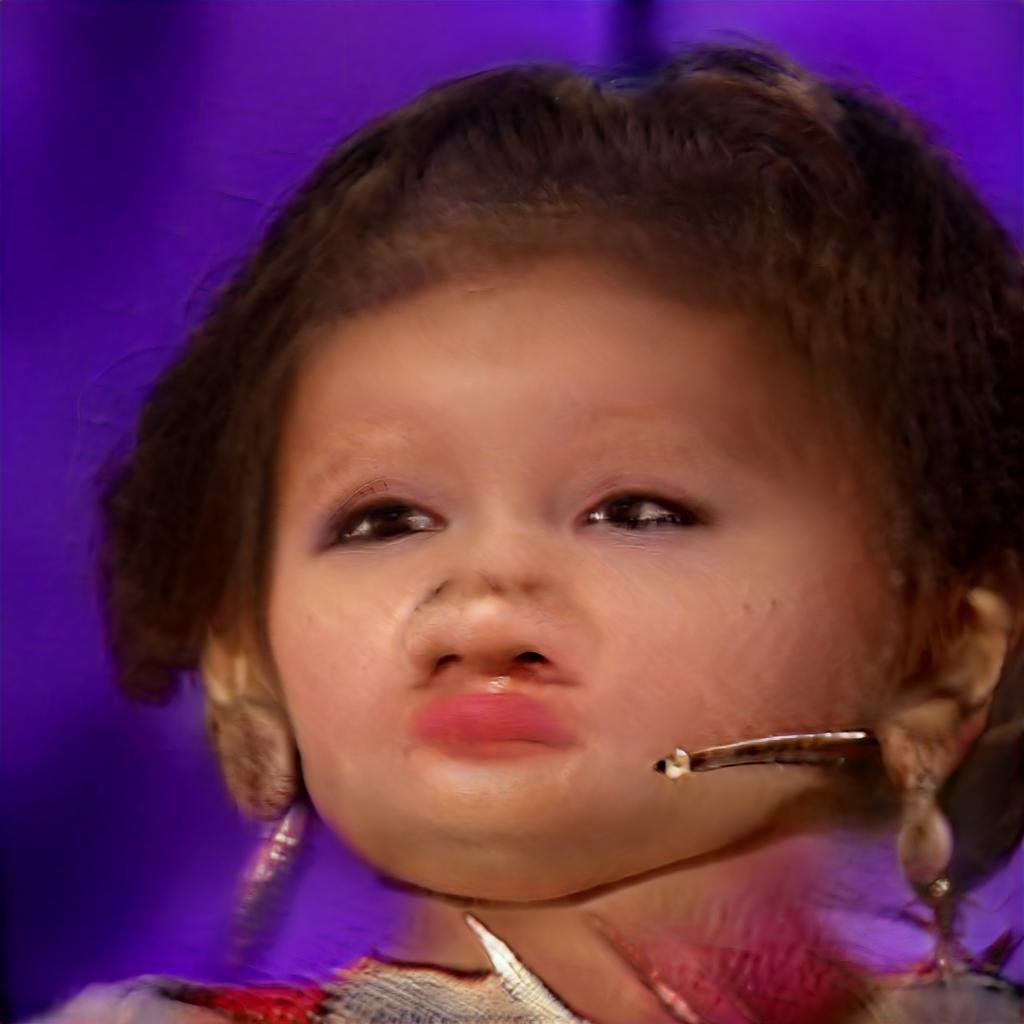}\hfill%
        \includegraphics[width=\ftqc]{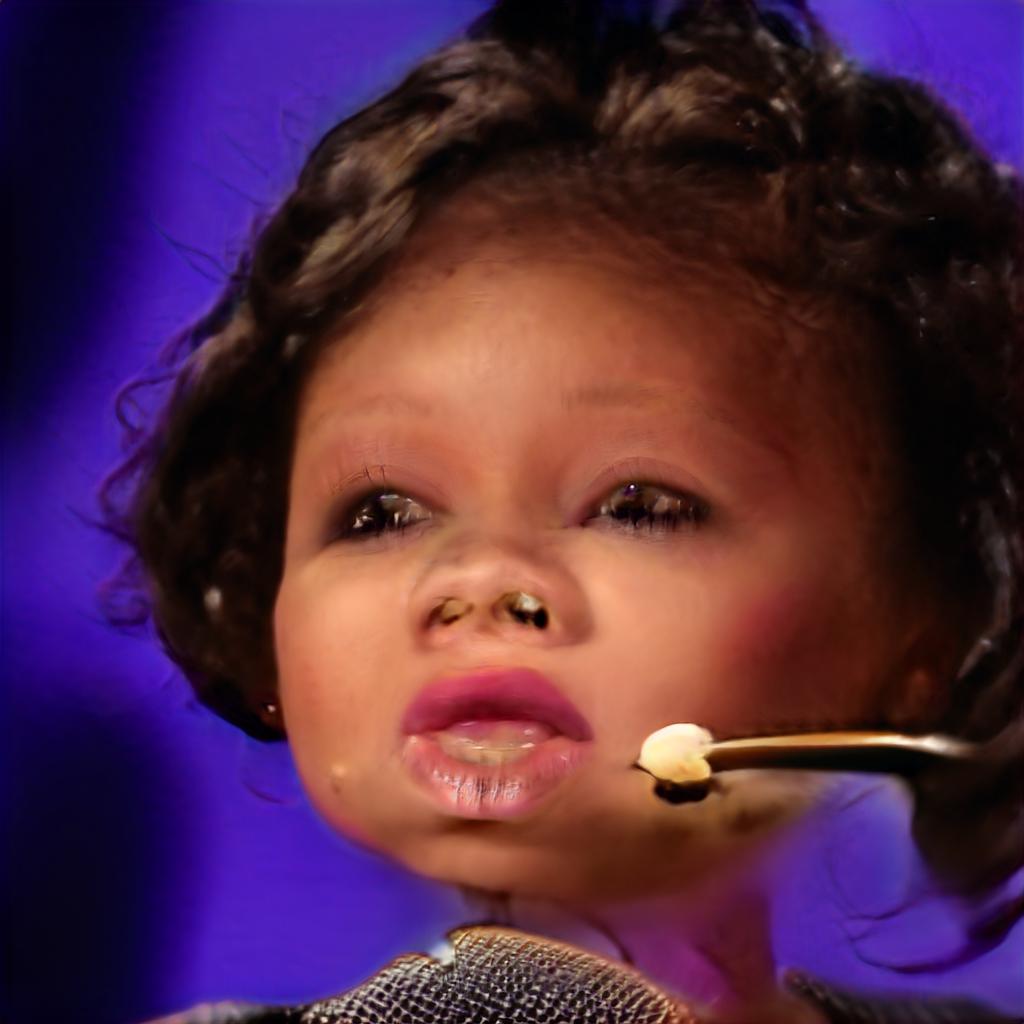}\hfill%
        \includegraphics[width=\ftqc]{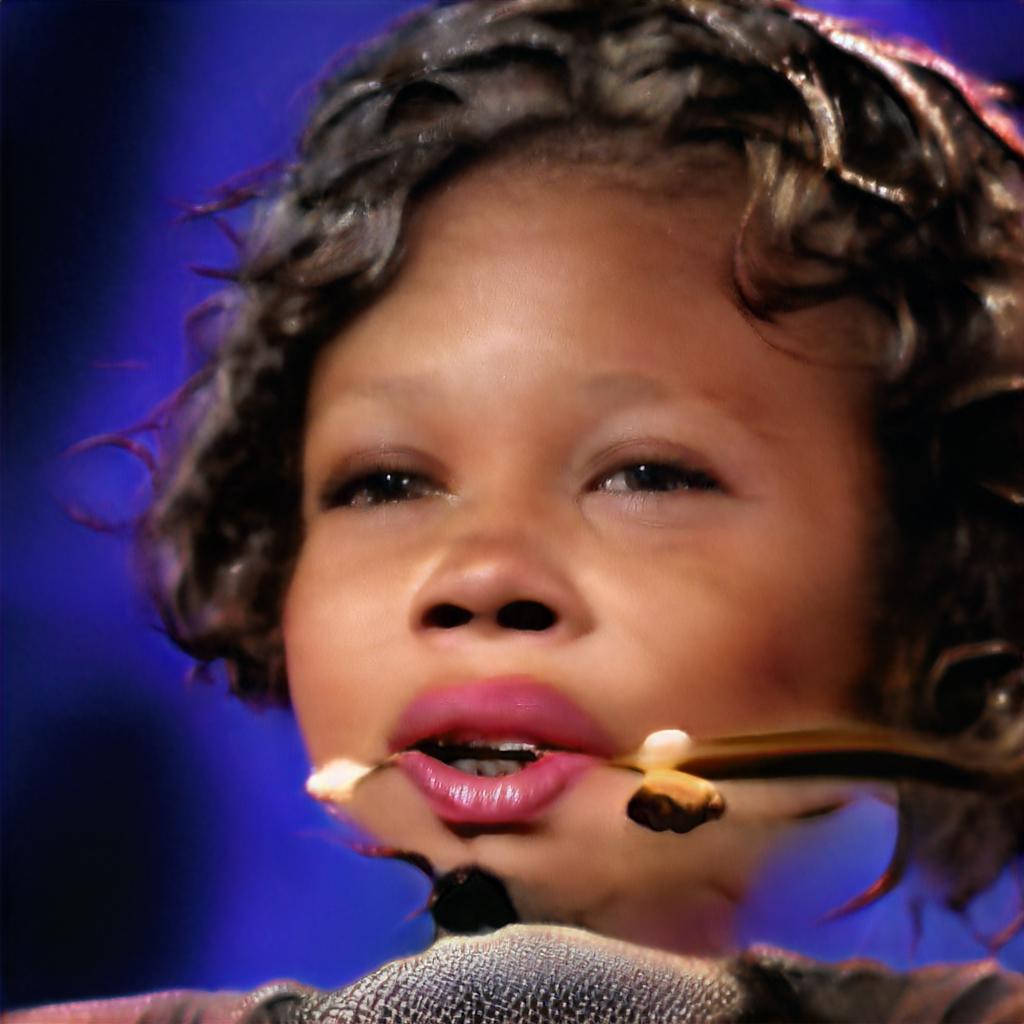}\hfill%
        \includegraphics[width=\ftqc]{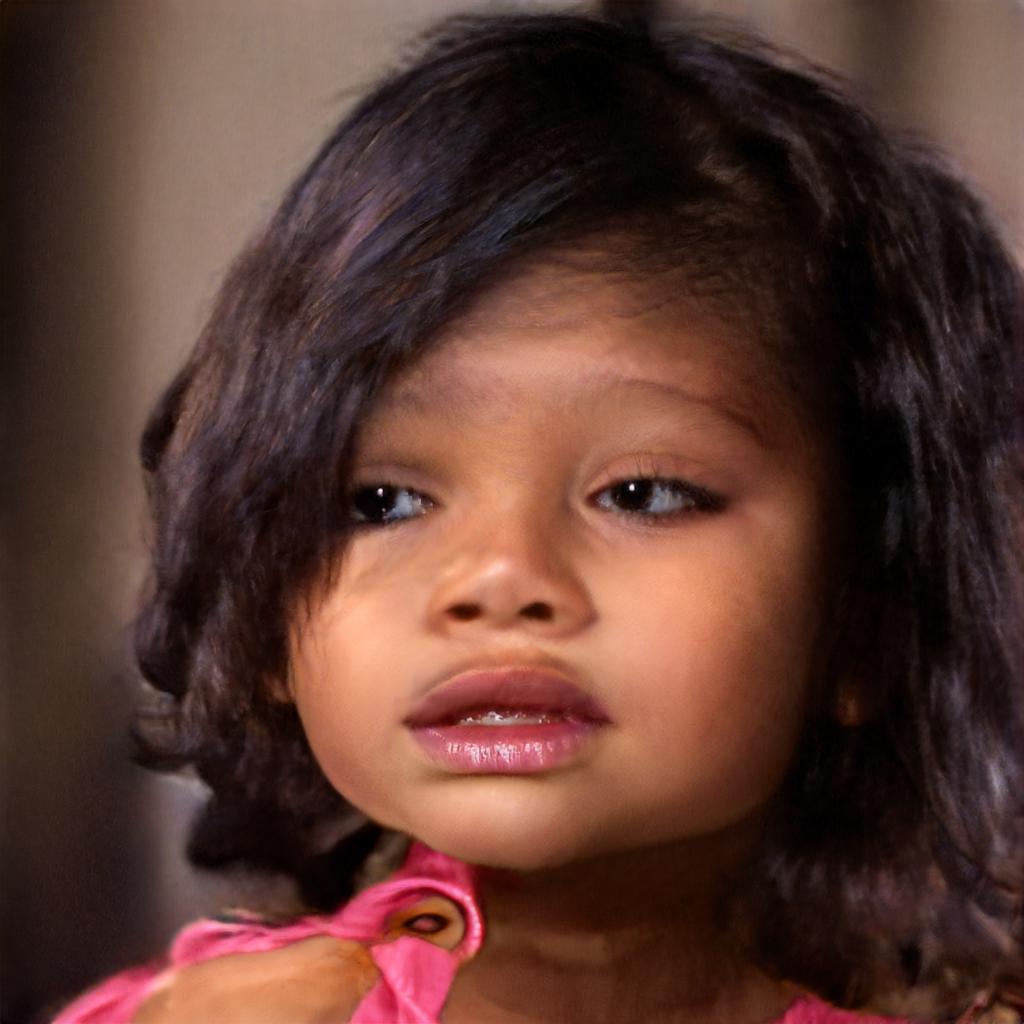}\hfill
        \\
        \includegraphics[width=\ftqc]{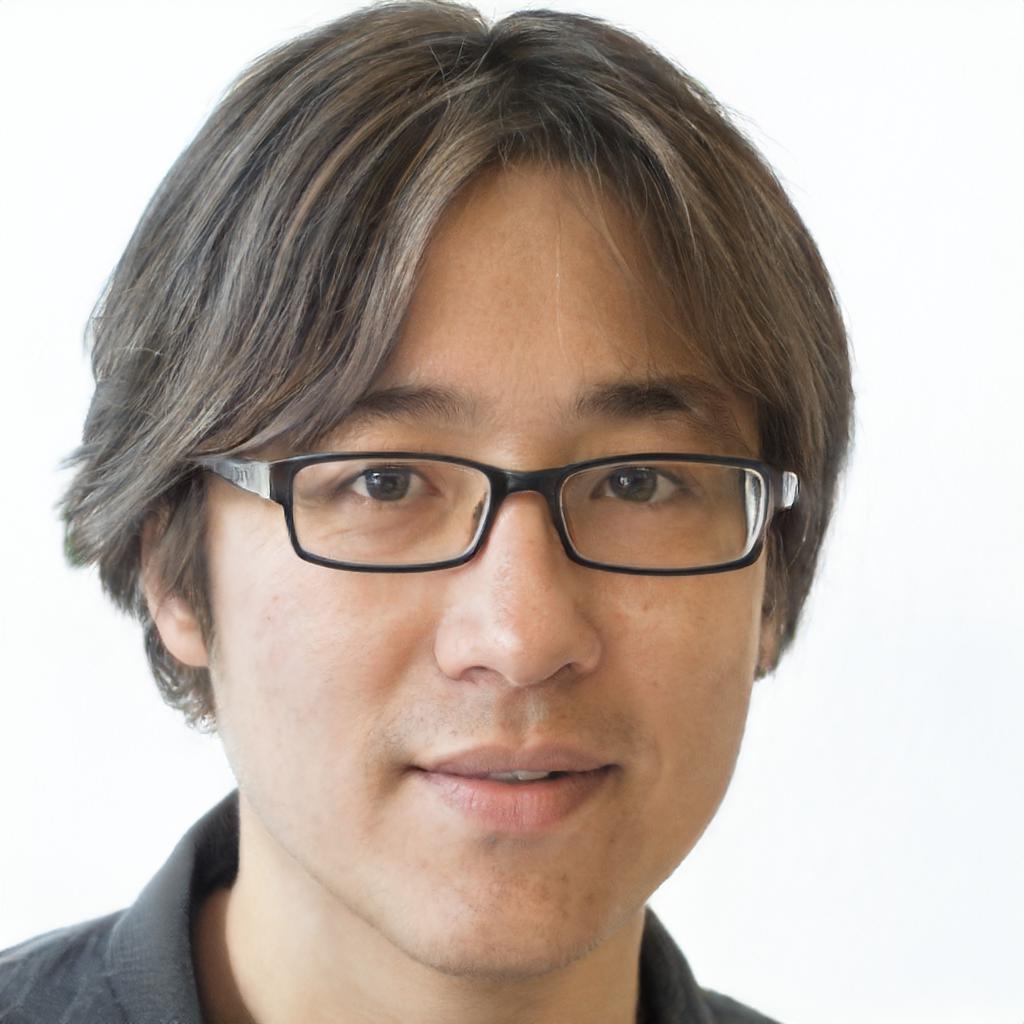}\hfill%
        \includegraphics[width=\ftqc]{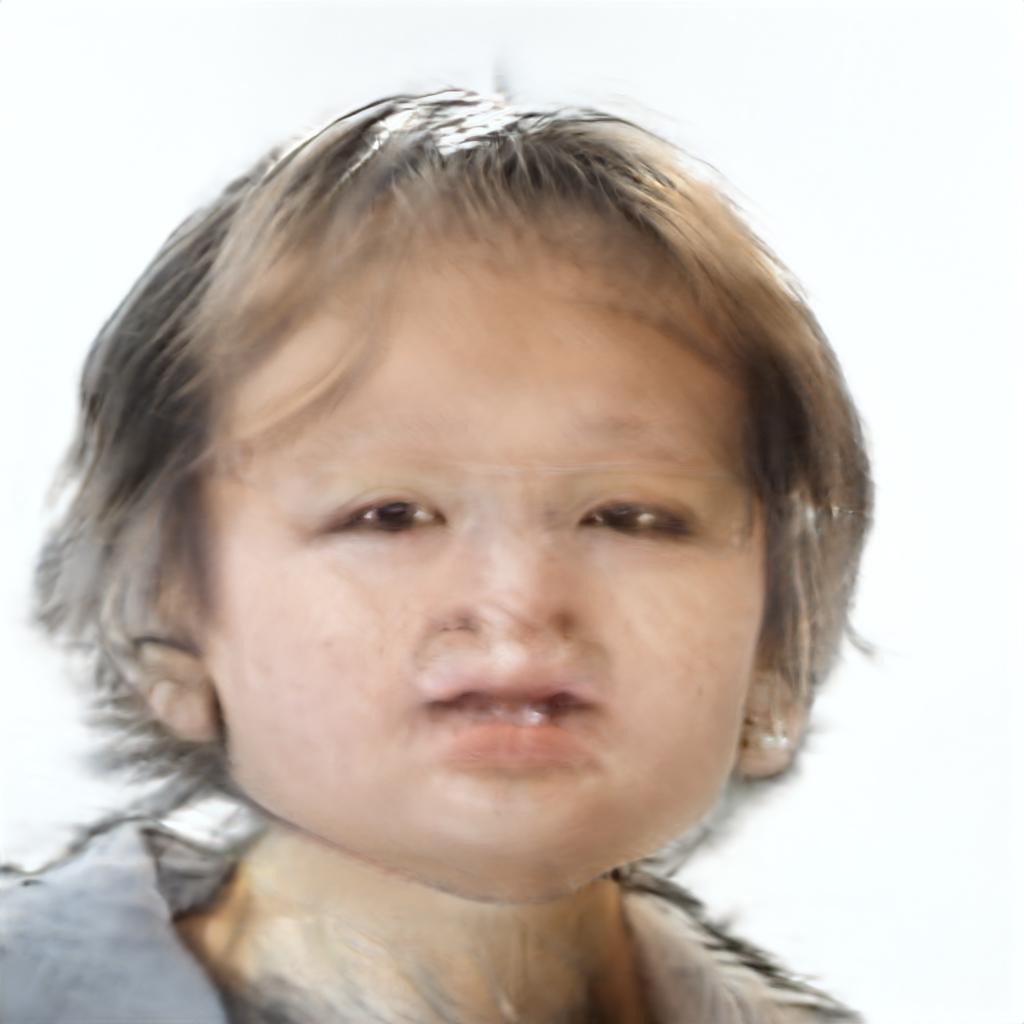}\hfill%
        \includegraphics[width=\ftqc]{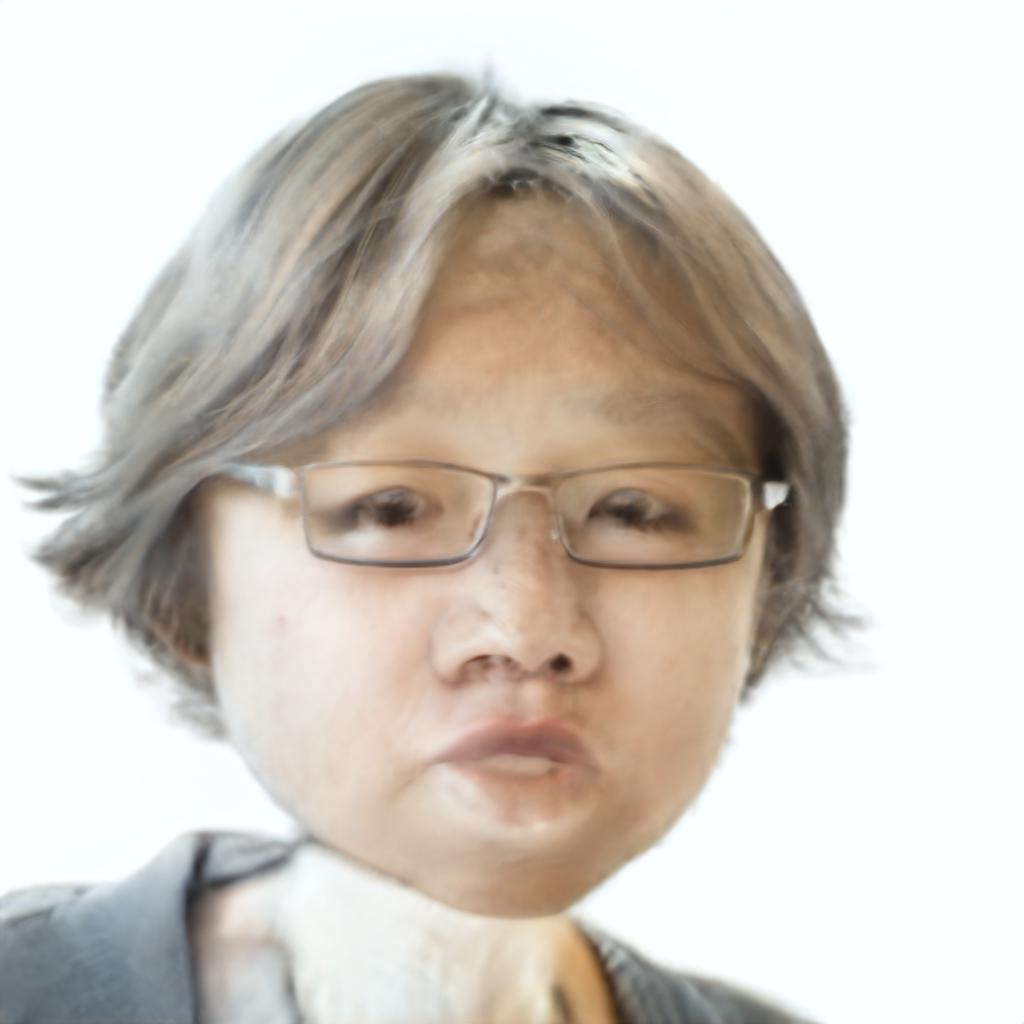}\hfill%
        \includegraphics[width=\ftqc]{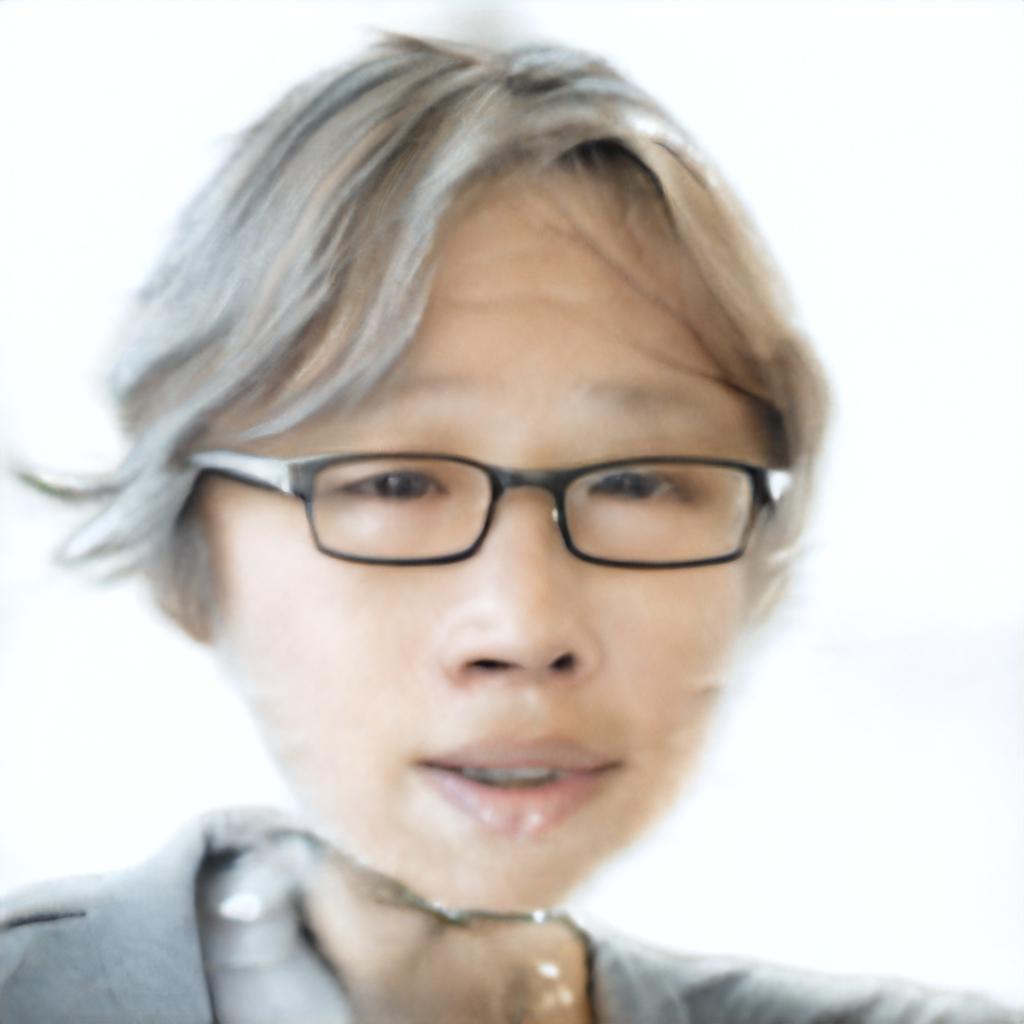}\hfill%
        \includegraphics[width=\ftqc]{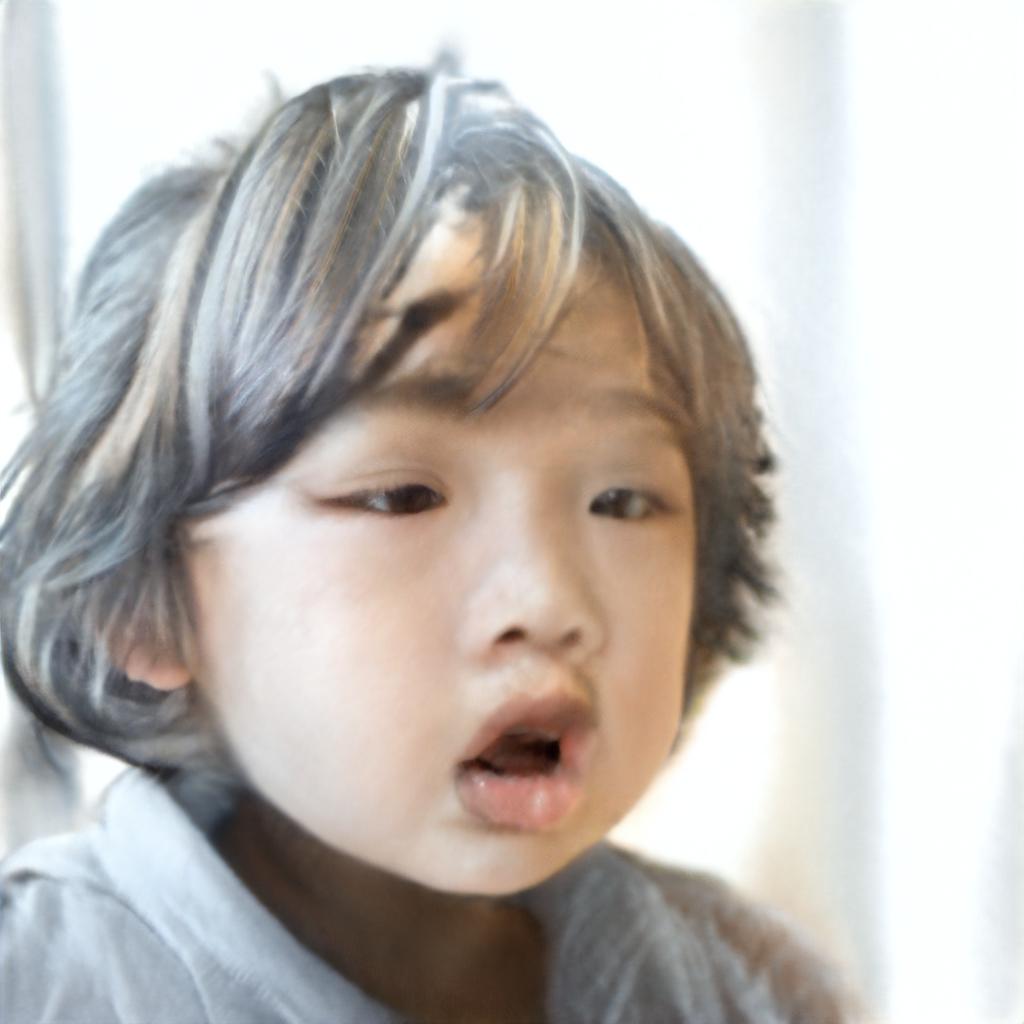}\hfill
    }
    \caption{
    \textbf{Close-domain adaptation} (FFHQ$\rightarrow$CelebA).
    Models adapted from a pretrained StyleGAN2 using $\sim$30 target images (left-most column) of \tb{(a)} CelebA ID 4978 and \tb{(b)} CelebA ID 3719.
    The proposed FSGAN generates more natural face images without noticeable artifacts.
    Comparison methods include TGAN \citep{wang2018eccv}, FD \citep{mo2020freeze}, SSGAN \citep{noguchi2019iccv}, trained with a limited number of timesteps to prevent overfitting or degradation.}
    \label{fig:personalization}
\end{figure*}

\begin{table}[H]
    \vspace{4ex}
    \centering
    \small
    \caption{
    {\bf Quantitative comparisons} in three metrics: FID \citep{heusel2017nerips}, Face Quality Index (FQI) \citep{hernandez2019faceqnet}, and sharpness \citep{kumar2012sharpness}.  
    See Fig \ref{fig:personalization} for illustrations.
    FQI and Sharpness are evaluated on 1,000 images randomly generated with the same set of seeds.
    % Compared to ``pretrain'' that has visibility to large training sets, the few-shot methods only see ~30 images.
    Bracketed/bold numbers indicated the best/second best results, respectively.}
    \vspace{-1.5ex}
    \label{tab:personaliation}
    \begin{tabu} to \textwidth {X[l 1.5] *{6}{X[c]}}
        \toprule
        & \multicolumn{3}{c}{CelebA 4978} & \multicolumn{3}{c}{CelebA 3719} \\
        \cmidrule(lr){2-4} \cmidrule(lr){5-7}
        {\bf Method} & FID & FQI & Sharpness & FID & FQI & Sharpness \\
        \midrule
        Pretrain     & --     & 0.40$\pm$0.11         & 0.91$\pm$0.06         & --       & 0.37$\pm$0.12       & 0.92$\pm$0.06 \\
        TransferGAN  & 75.41  & 0.30$\pm$0.07         & 0.61$\pm$0.05         & 178.31   & 0.26$\pm$0.09       & {\bf 0.61$\pm$0.04} \\
        FreezeD      & 75.30  & {\bf 0.33$\pm$0.09}   & 0.58$\pm$0.04         & 143.83   & {\bf 0.27$\pm$0.09} & 0.56$\pm$0.05 \\
        SSGAN        & 87.79  & 0.32$\pm$0.08         & {\bf [0.67$\pm$0.05]} & 147.14   & 0.27$\pm$0.10       & 0.58$\pm$0.05 \\
        FSGAN (ours) & 78.90  & {\bf [0.36$\pm$0.07]} & {\bf 0.65$\pm$0.05}   & 170.00   & {\bf 0.27$\pm$0.08} & {\bf [0.68$\pm$0.07]} \\
        \bottomrule
    \end{tabu}
\end{table}

\setlength{\textfloatsep}{6pt}
\begin{figure*}[t]
\begin{center}
\centering
\newlength\ftqa
\setlength\ftqa{1.9cm}
\newlength\ftqb
\setlength\ftqb{3.8cm}
\setlength\tabcolsep{0.1pt}
\newcommand{\gtop}{015} % 0, 4, 9, 15, 16, 65
\newcommand{\gbot}{051}
\newcommand{\ptop}{329} % female: 0, 329, 421, 433 (kid)
\newcommand{\pbot}{003} % male: 3, 14*, 39, 44 (kid), 70, 
\newcommand{\rtop}{502}
\newcommand{\rbot}{532}
\mpage{0.02}\hfill
\mpage{0.26}{Target Images} \hfill
\mpage{0.12}{Pretrain} \hfill
\mpage{0.11}{TGAN} \hfill %\citet{wang2018eccv}
\mpage{0.14}{FD} \hfill %\citet{mo2020freeze}
\mpage{0.12}{SSGAN} \hfill %\citet{noguchi2019iccv}
\mpage{0.12}{FSGAN}\hfill
\vspace{0.1mm}
\mpage{0.01}{\rotatebox[origin=c]{90}{\small \tb{(a)} Van Gogh (25-shot)}}
\mpage{0.26}{\includegraphics[height=\ftqb]{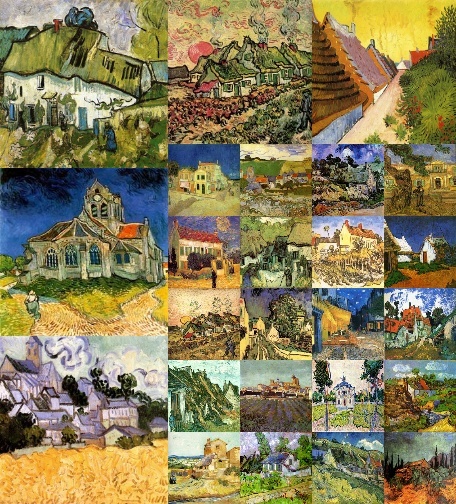}}
\mpage{0.69}{
%\\
\includegraphics[width=\ftqa]{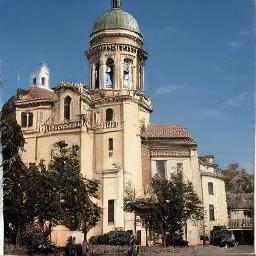}\hfill%
\includegraphics[width=\ftqa]{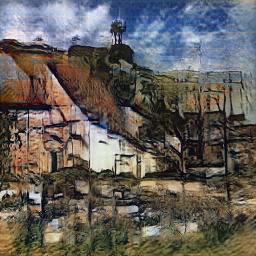}\hfill%
\includegraphics[width=\ftqa]{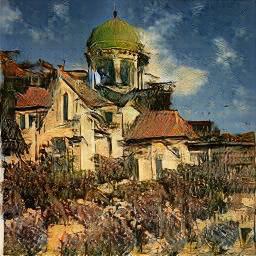}\hfill%
\includegraphics[width=\ftqa]{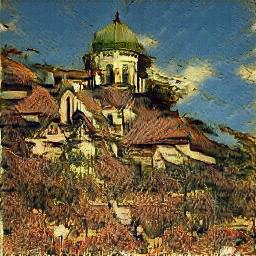}\hfill%
\includegraphics[width=\ftqa]{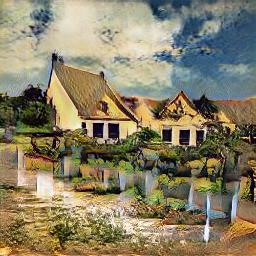}\hfill
%\\
\includegraphics[width=\ftqa]{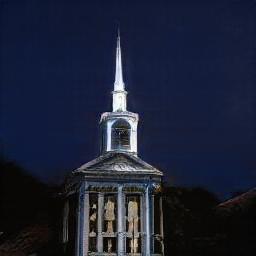}\hfill%
\includegraphics[width=\ftqa]{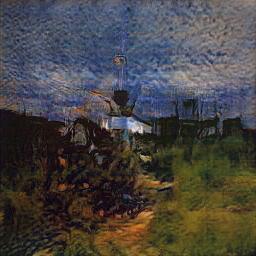}\hfill%
\includegraphics[width=\ftqa]{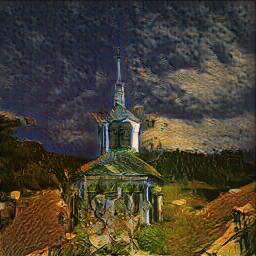}\hfill%
\includegraphics[width=\ftqa]{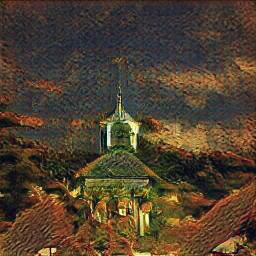}\hfill%
\includegraphics[width=\ftqa]{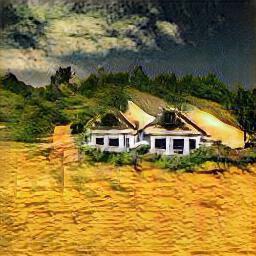}\hfill}
\mpage{0.01}{\rotatebox[origin=c]{90}{\small \tb{(b)} Portraits (25-shot)}}
\mpage{0.26}{\includegraphics[height=\ftqb]{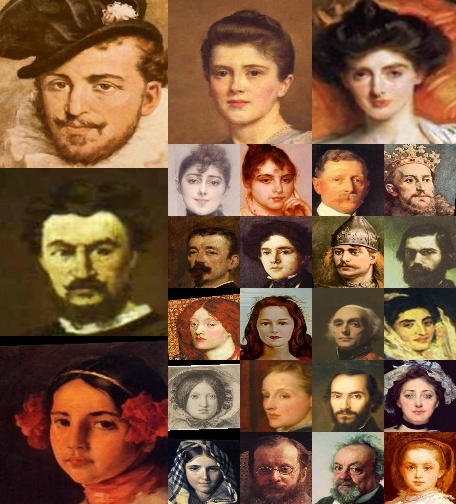}}
\mpage{0.69}{
\includegraphics[width=\ftqa]{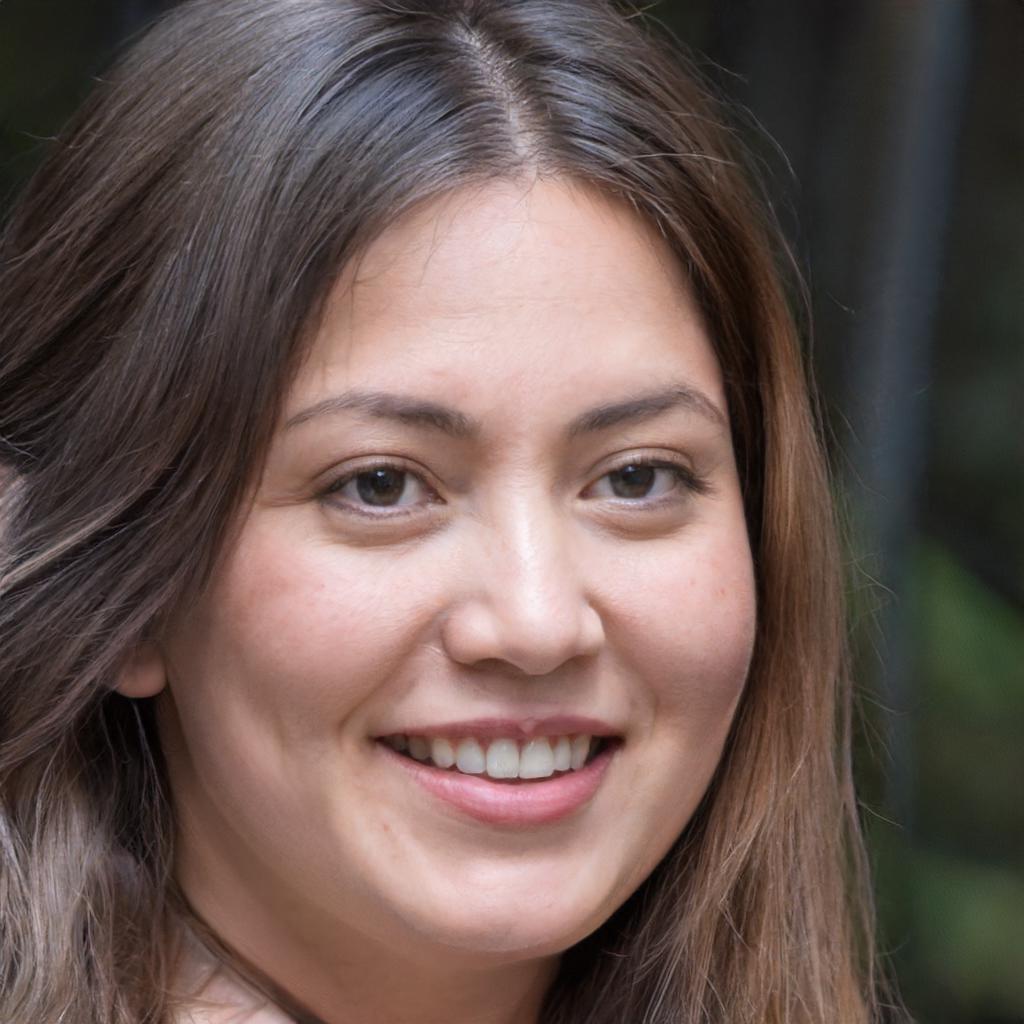}\hfill%
\includegraphics[width=\ftqa]{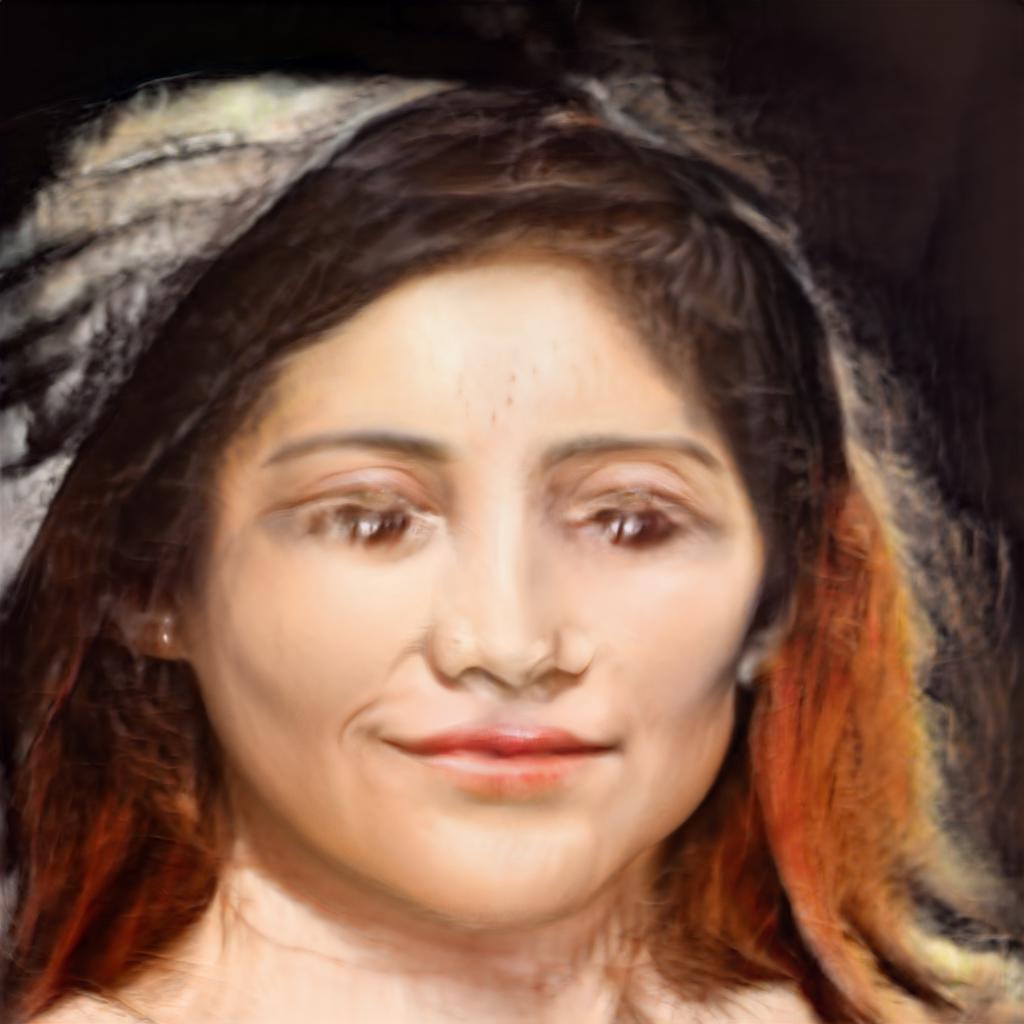}\hfill%
\includegraphics[width=\ftqa]{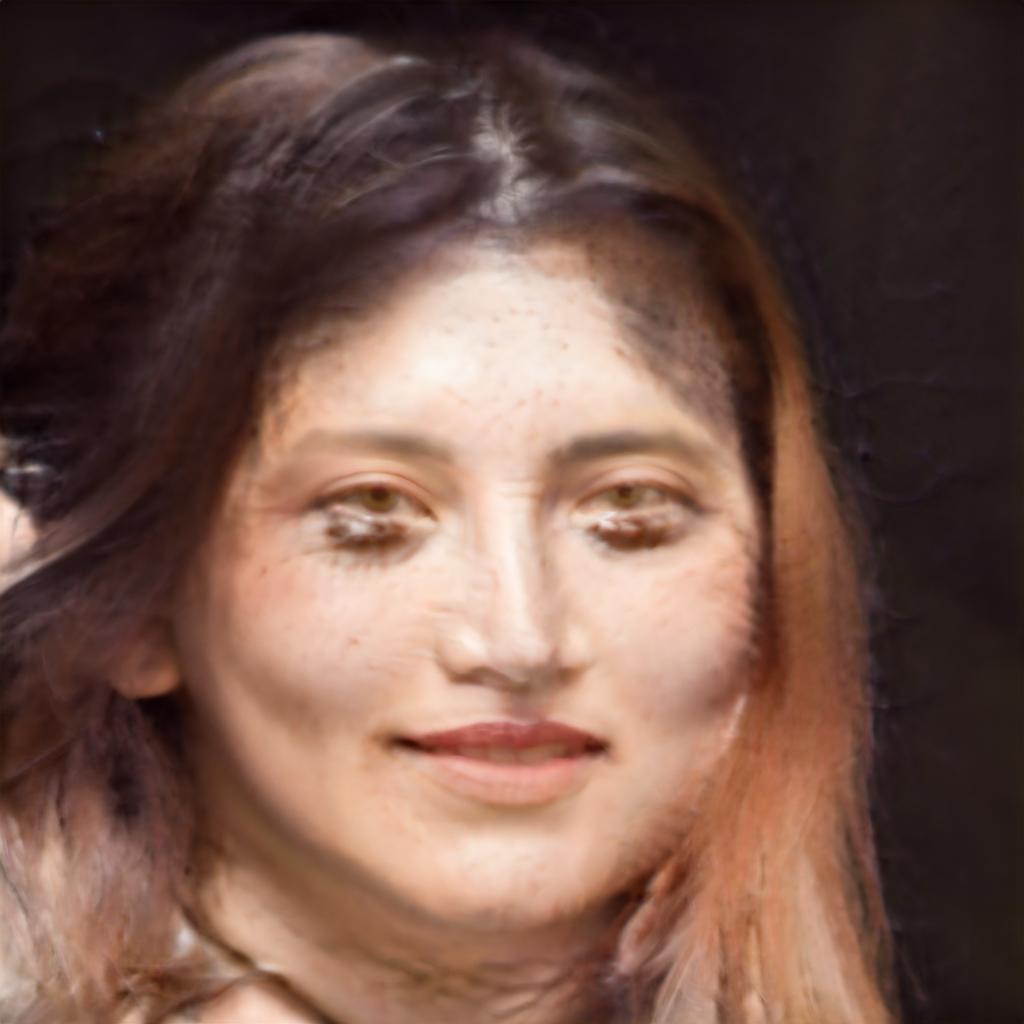}\hfill%
\includegraphics[width=\ftqa]{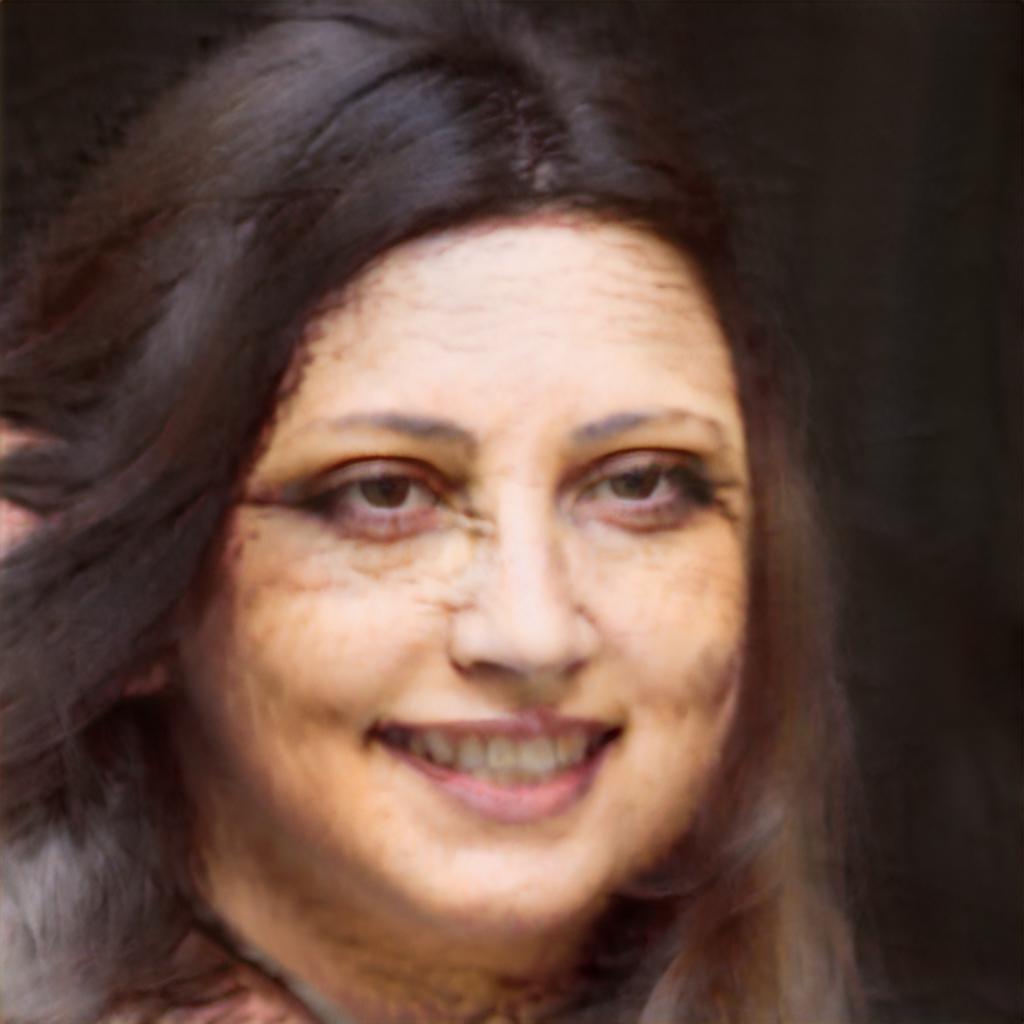}\hfill%
\includegraphics[width=\ftqa]{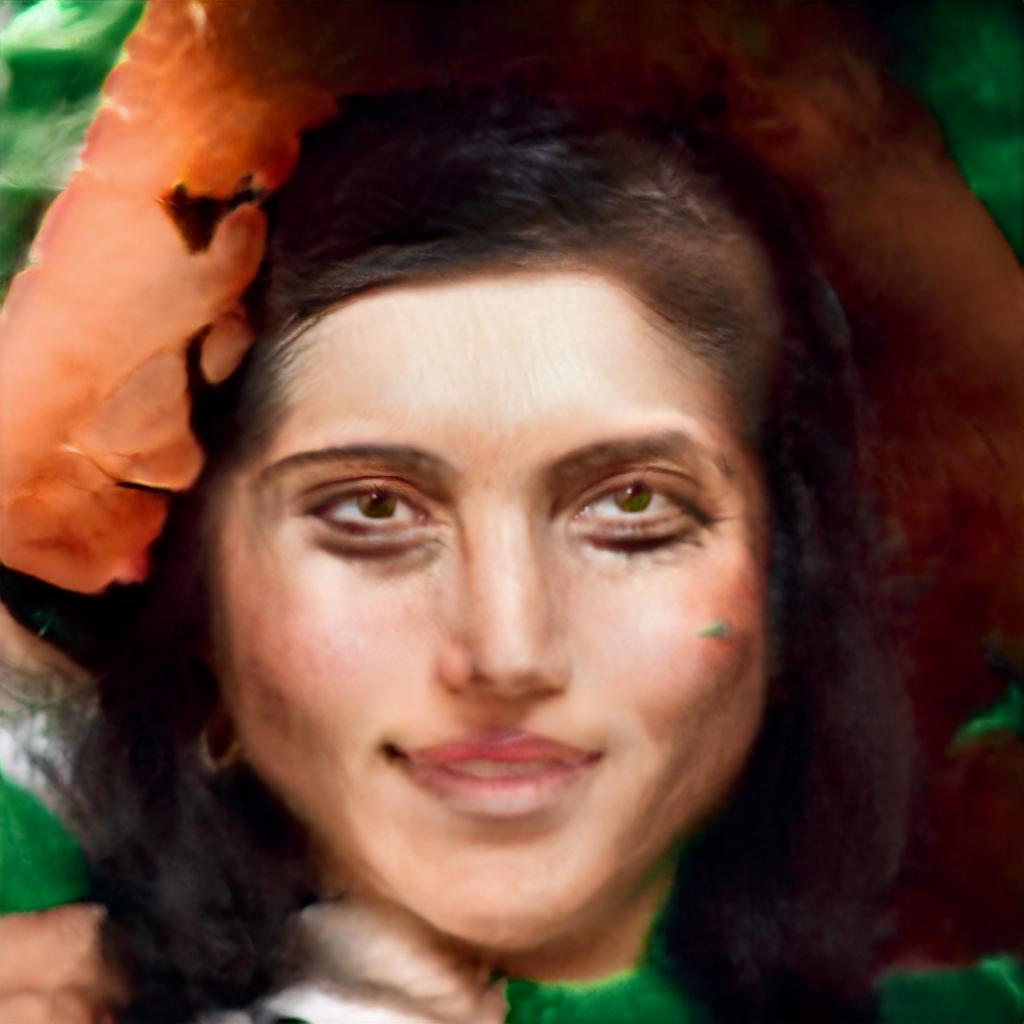}\hfill
\\
\includegraphics[width=\ftqa]{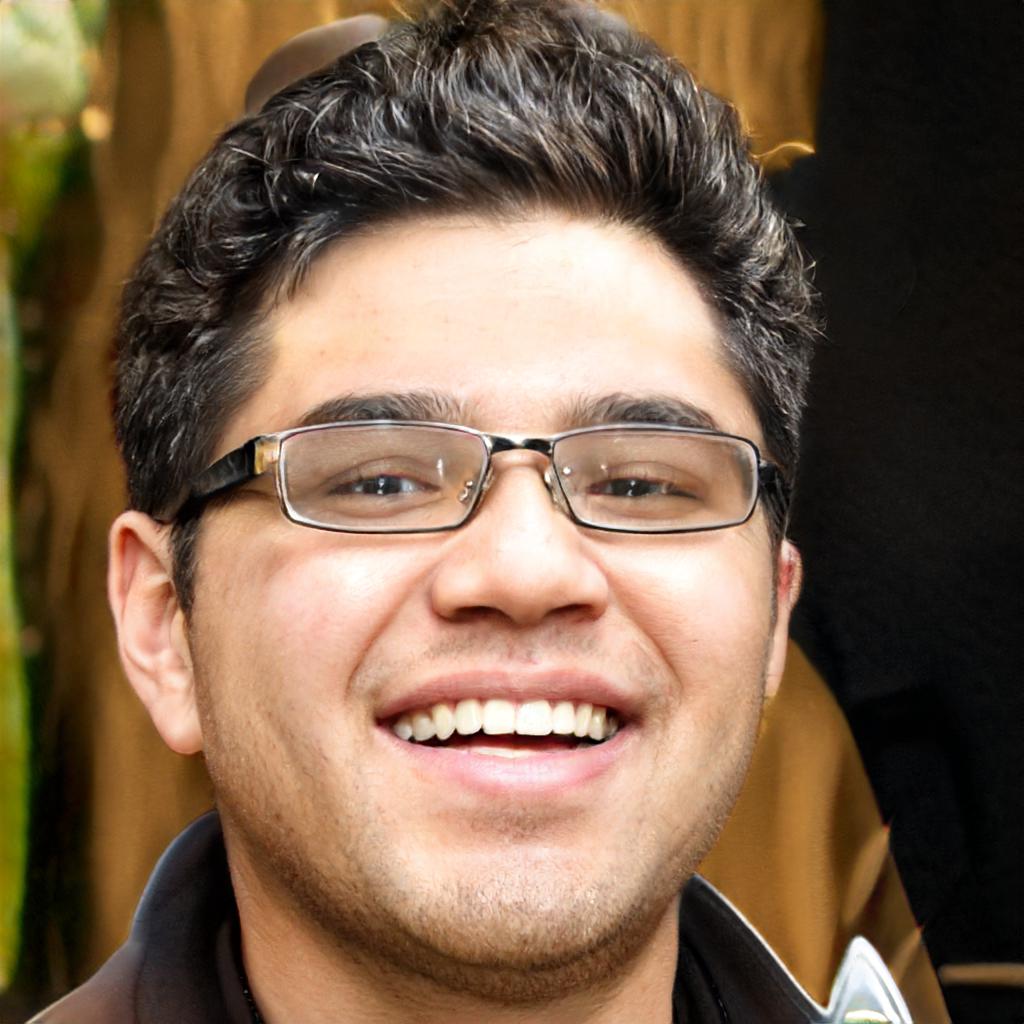}\hfill%
\includegraphics[width=\ftqa]{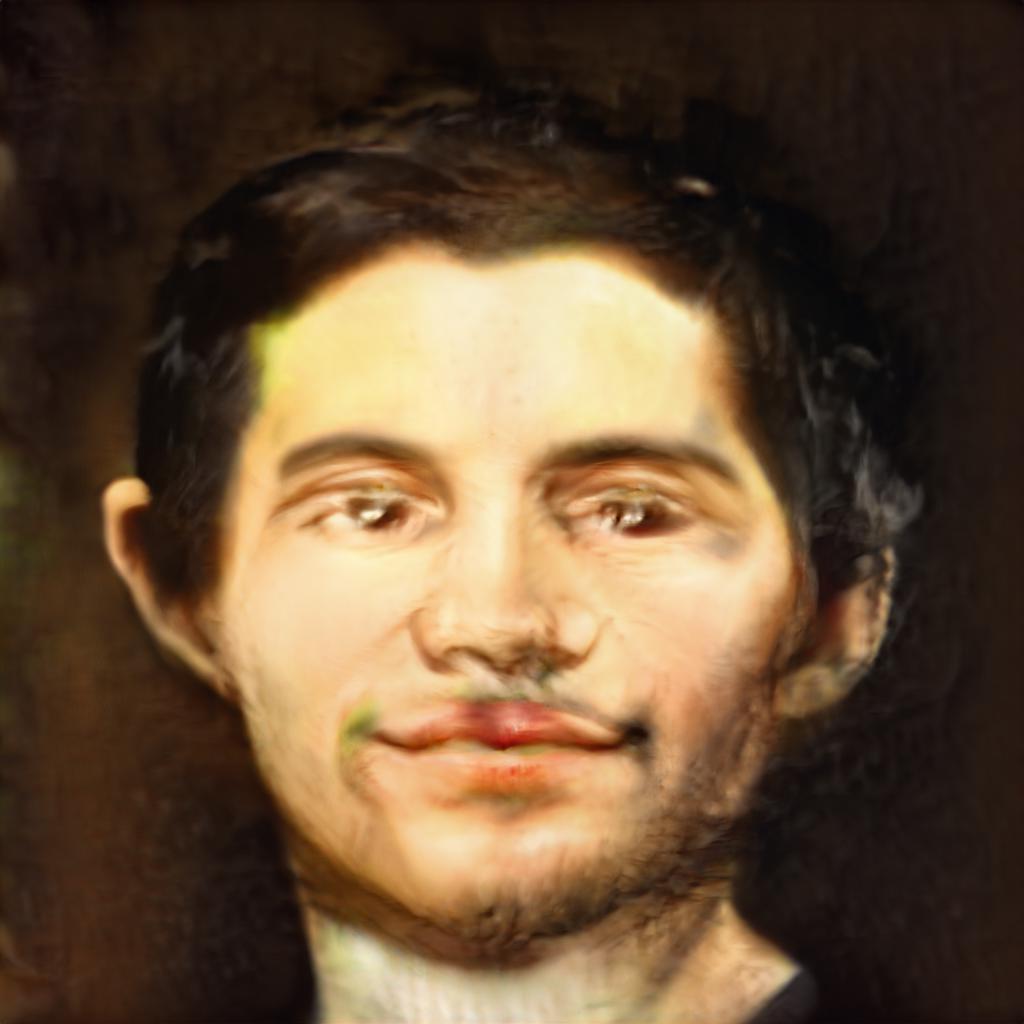}\hfill%
\includegraphics[width=\ftqa]{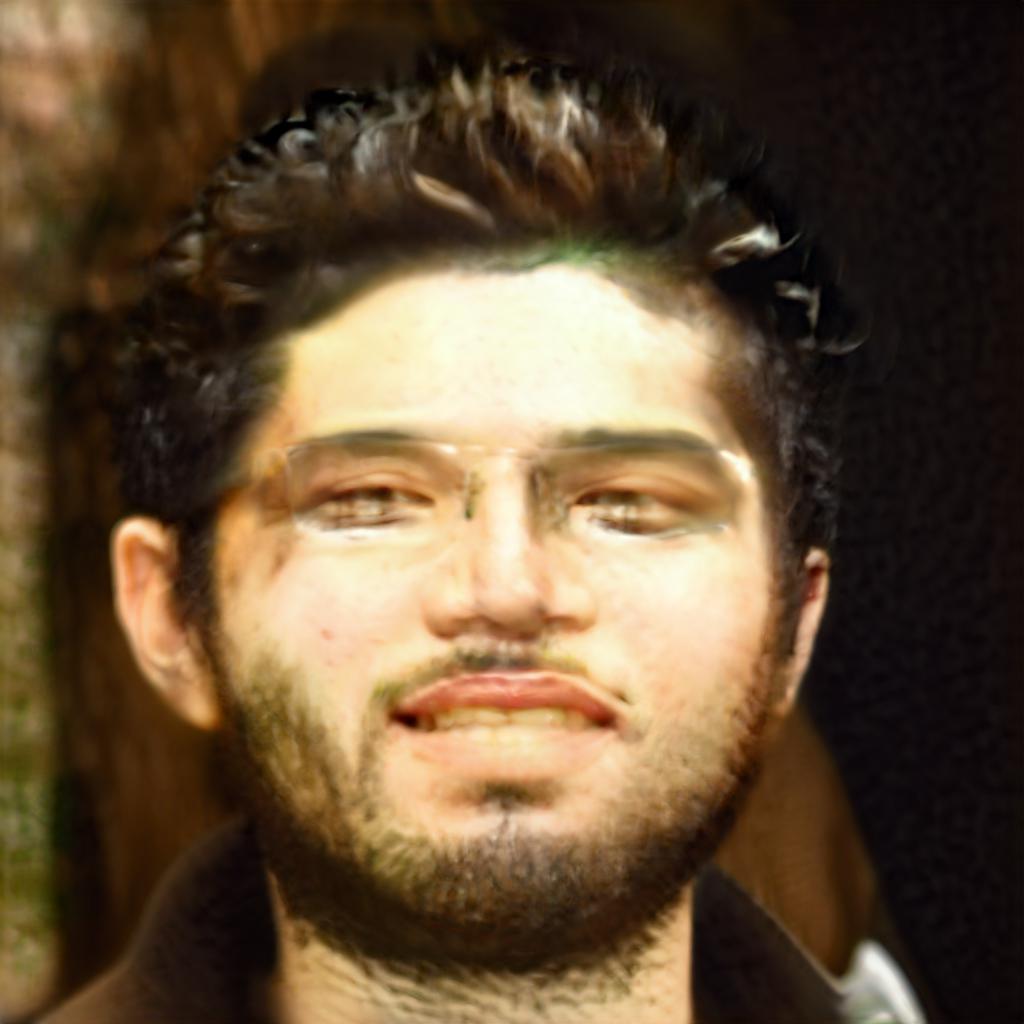}\hfill%
\includegraphics[width=\ftqa]{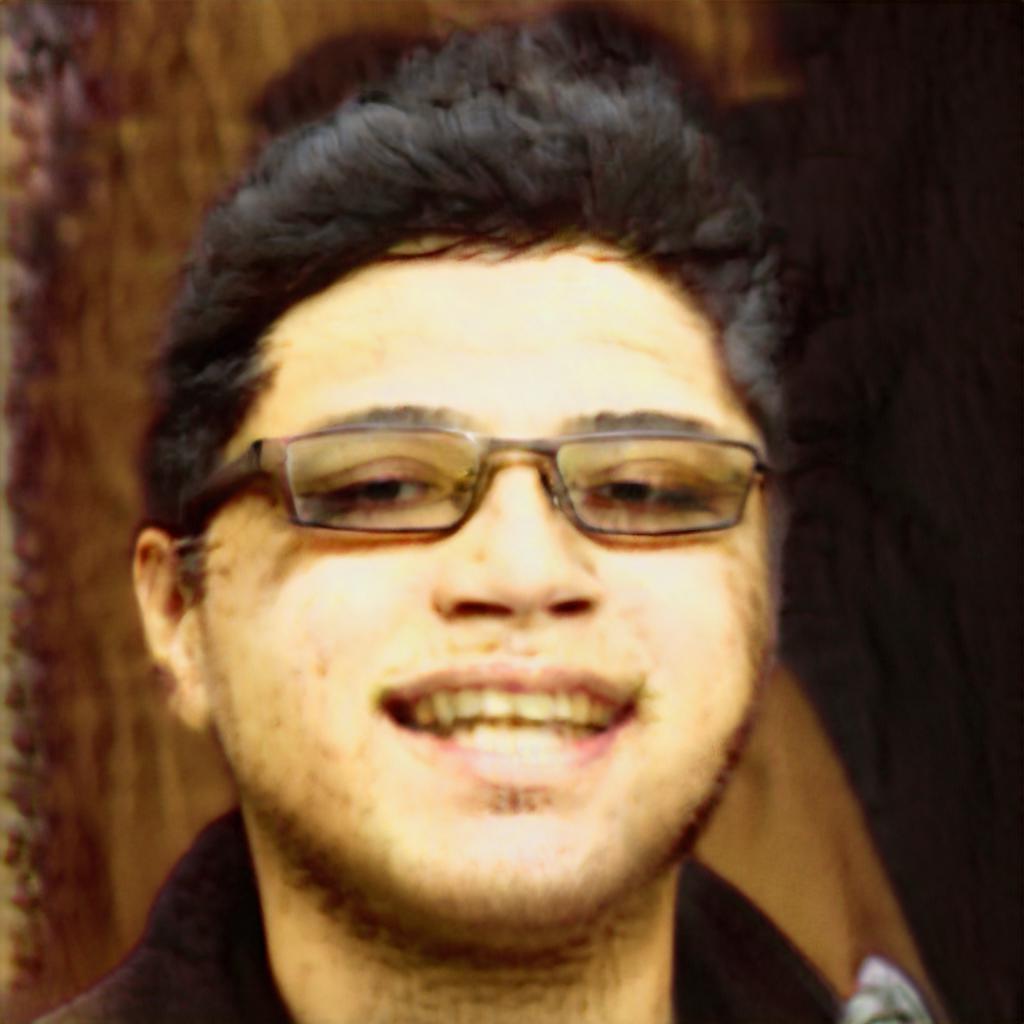}\hfill%
\includegraphics[width=\ftqa]{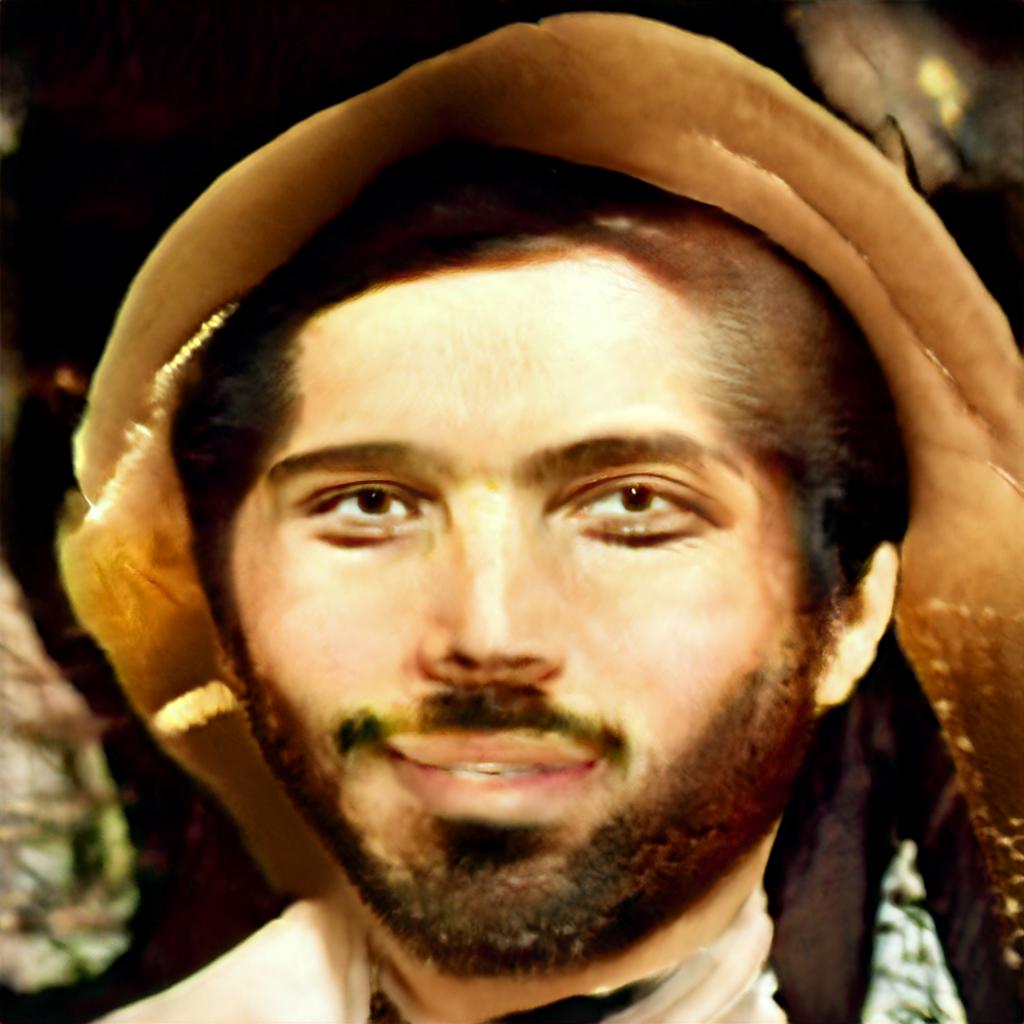}\hfill}
\mpage{0.01}{\rotatebox[origin=c]{90}{\small \tb{(c)} Rem (25-shot)}}
\mpage{0.26}{\includegraphics[height=\ftqb]{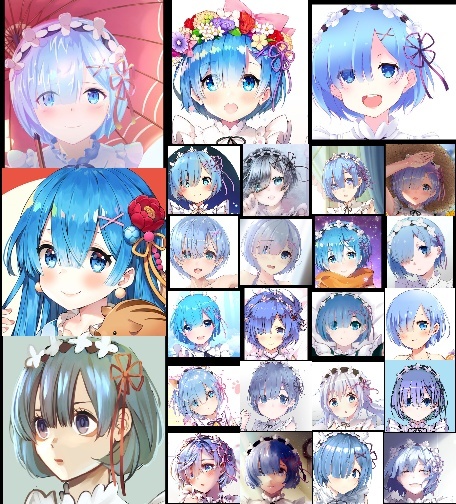}}
\mpage{0.69}{
\includegraphics[width=\ftqa]{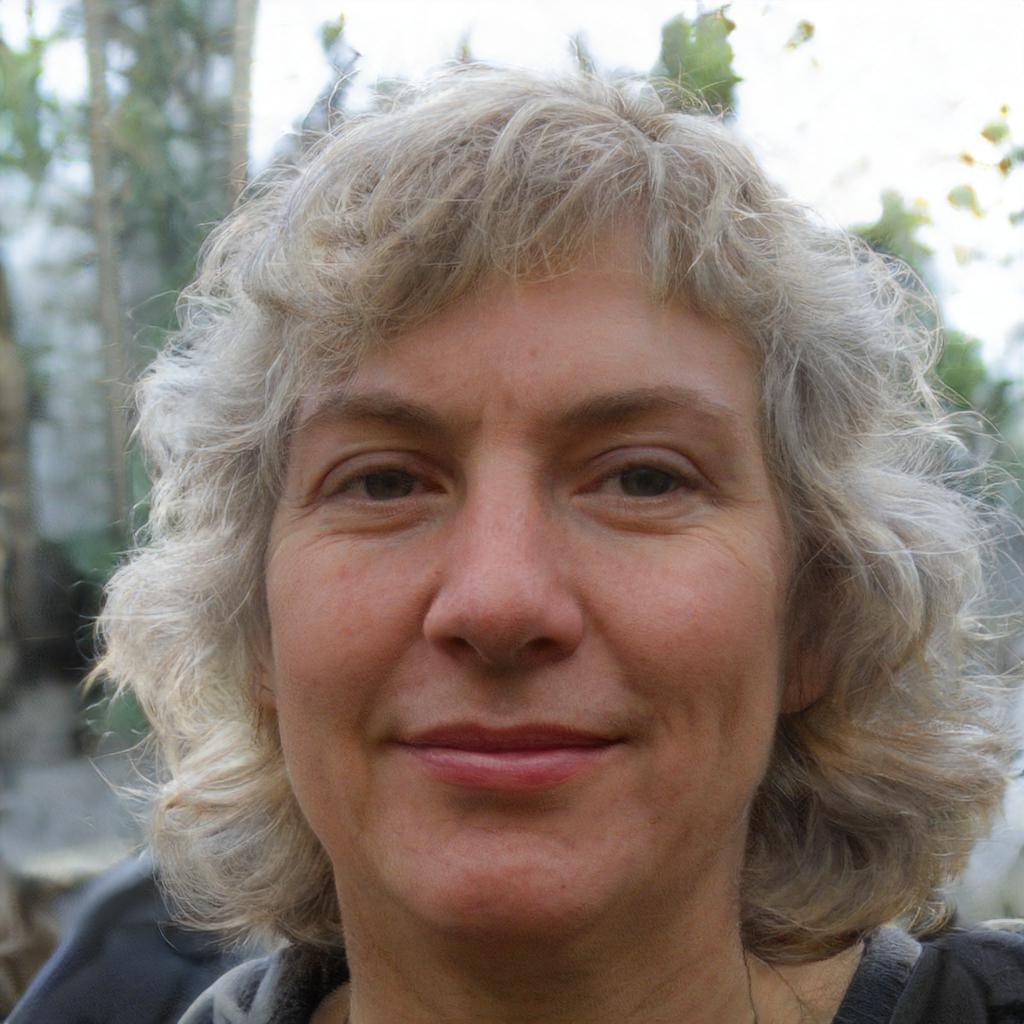}\hfill%
\includegraphics[width=\ftqa]{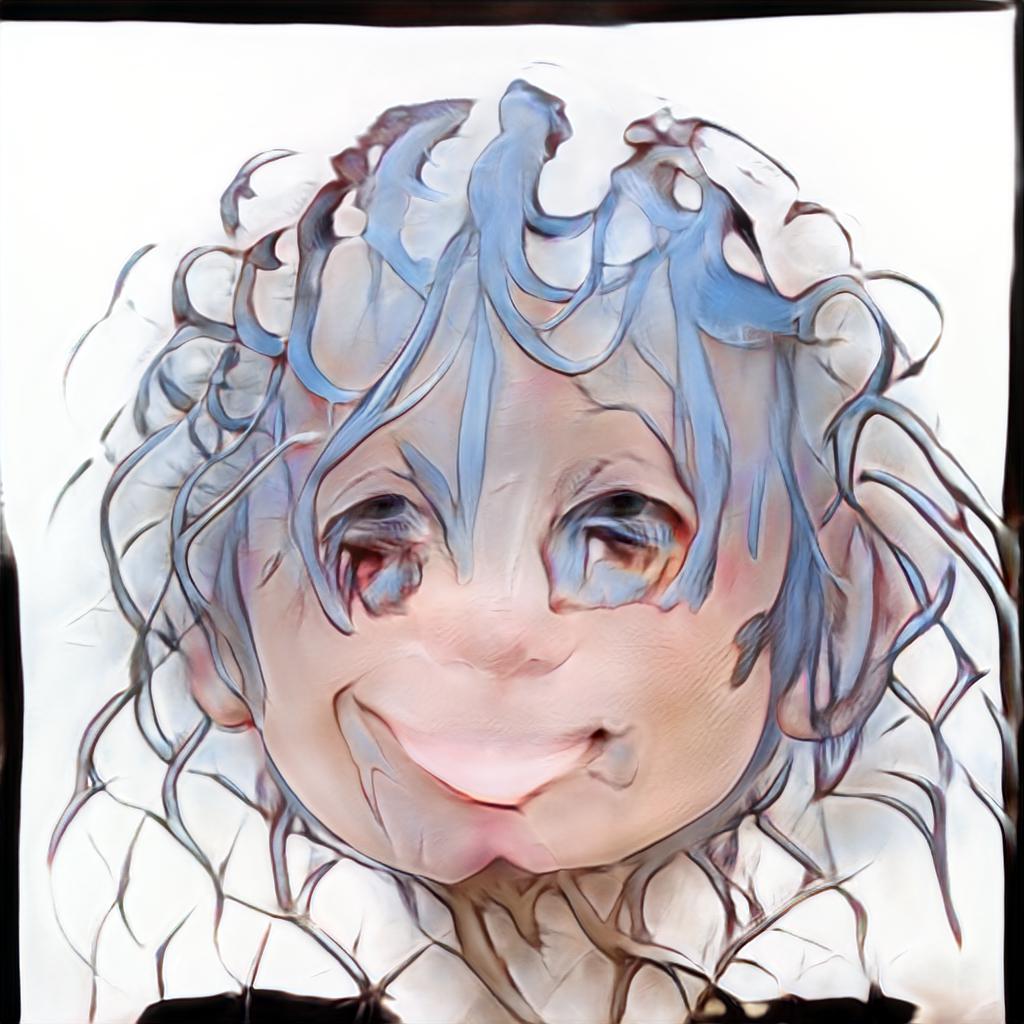}\hfill%
\includegraphics[width=\ftqa]{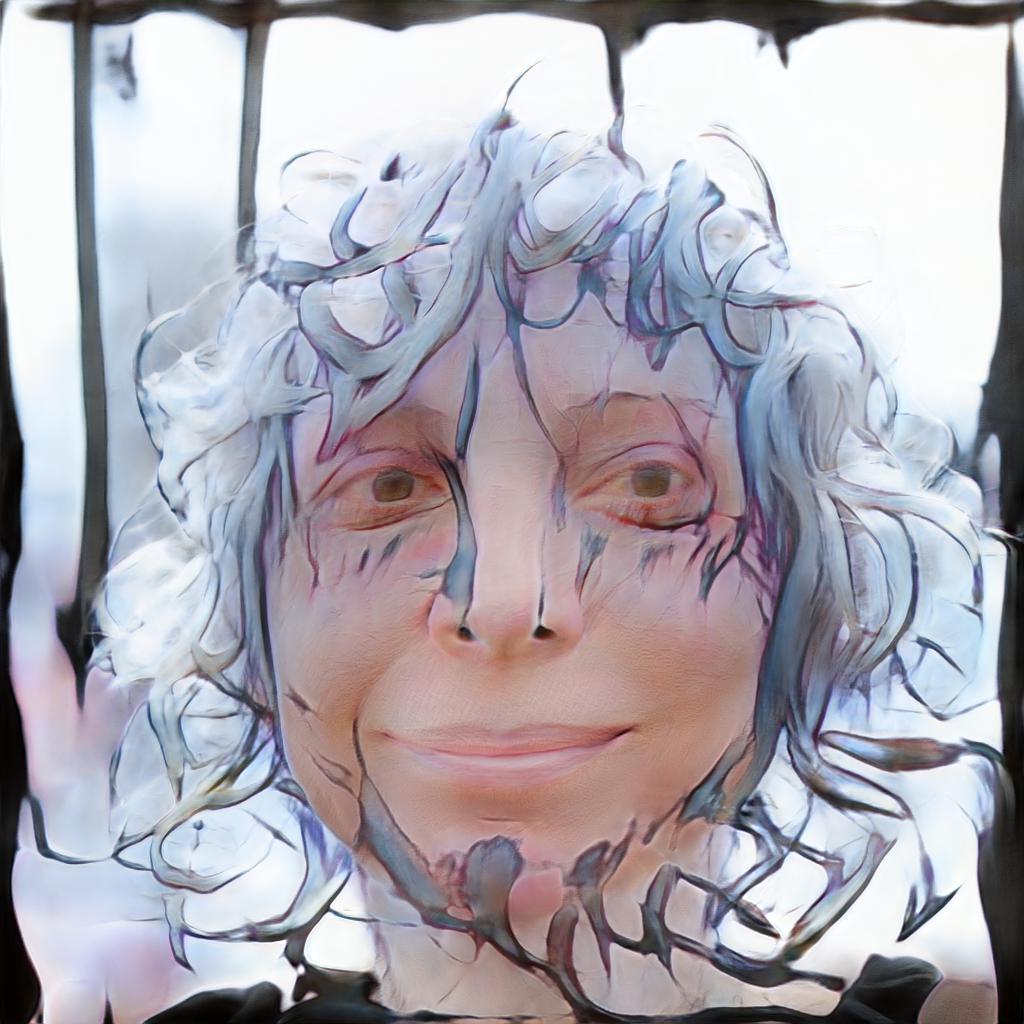}\hfill%
\includegraphics[width=\ftqa]{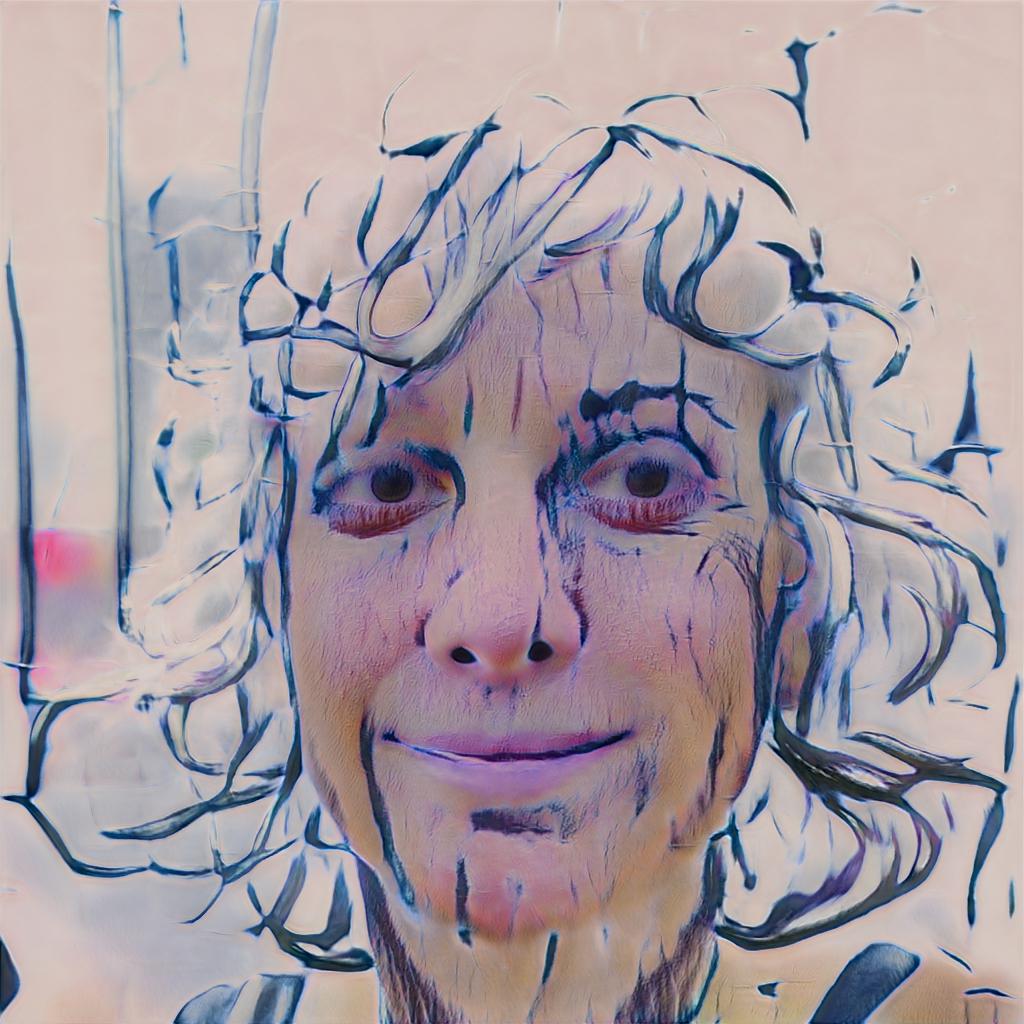}\hfill%
\includegraphics[width=\ftqa]{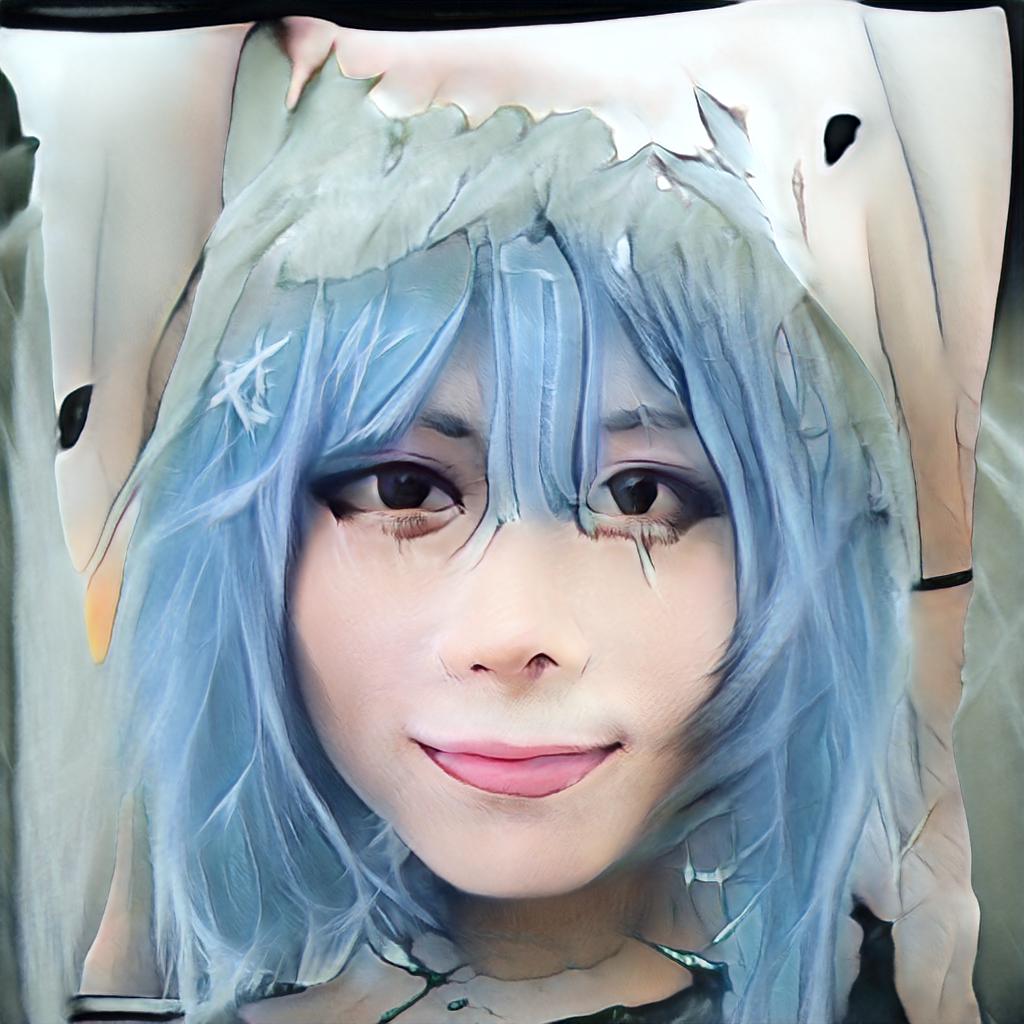}\hfill
\\
\includegraphics[width=\ftqa]{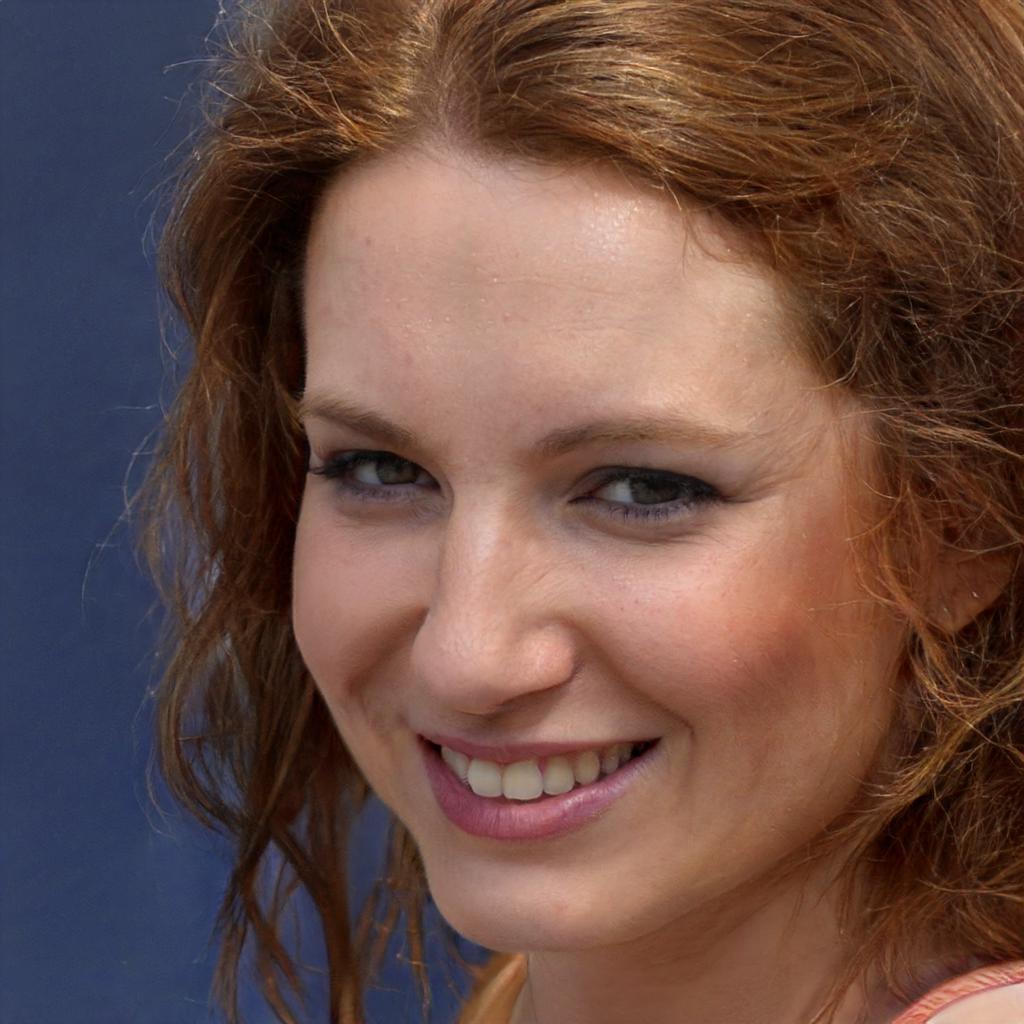}\hfill%
\includegraphics[width=\ftqa]{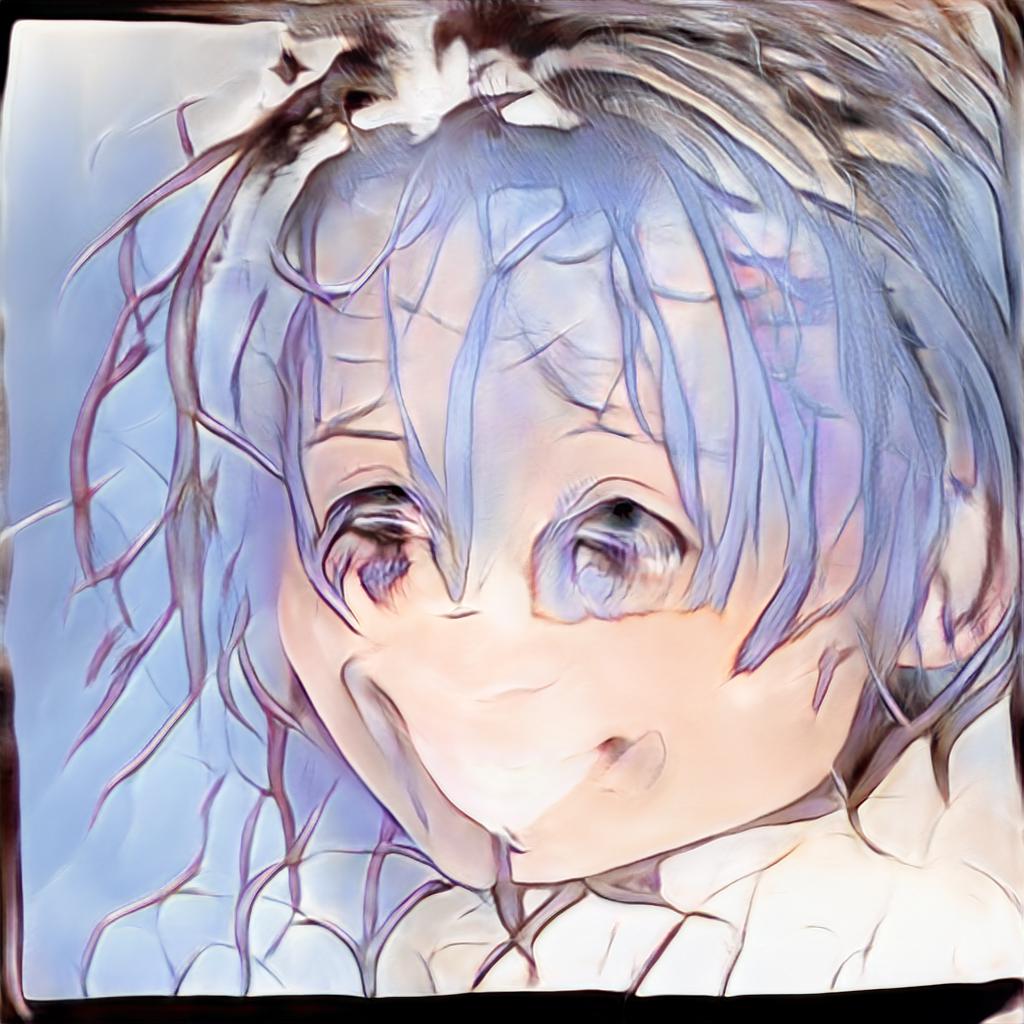}\hfill%
\includegraphics[width=\ftqa]{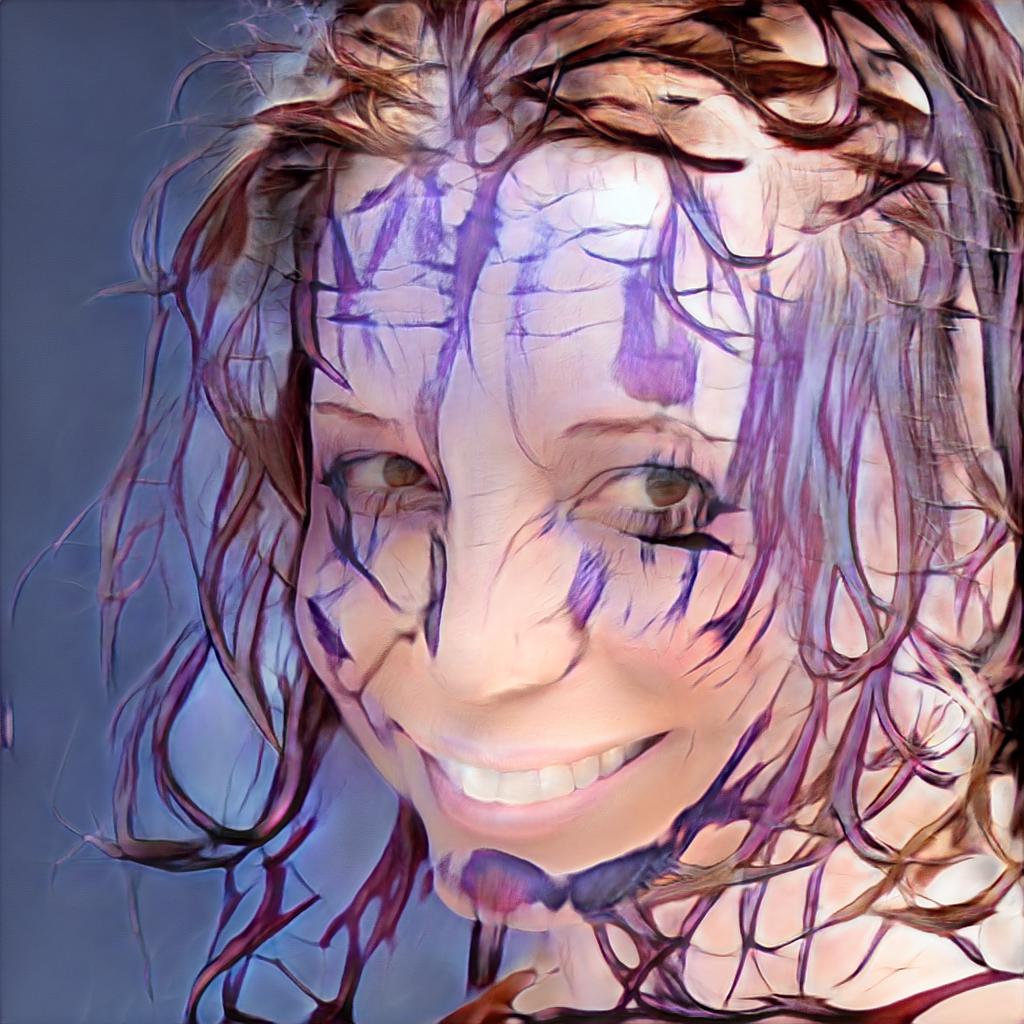}\hfill%
\includegraphics[width=\ftqa]{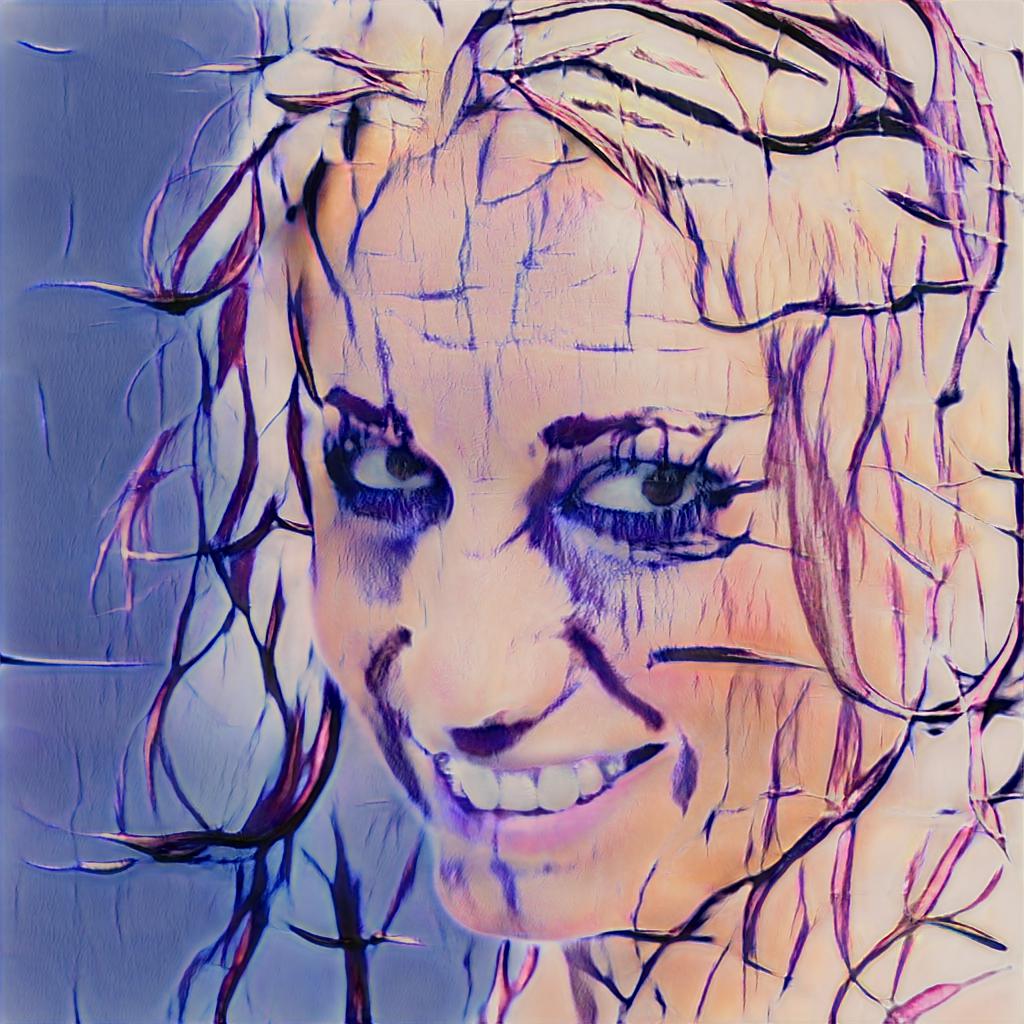}\hfill%
\includegraphics[width=\ftqa]{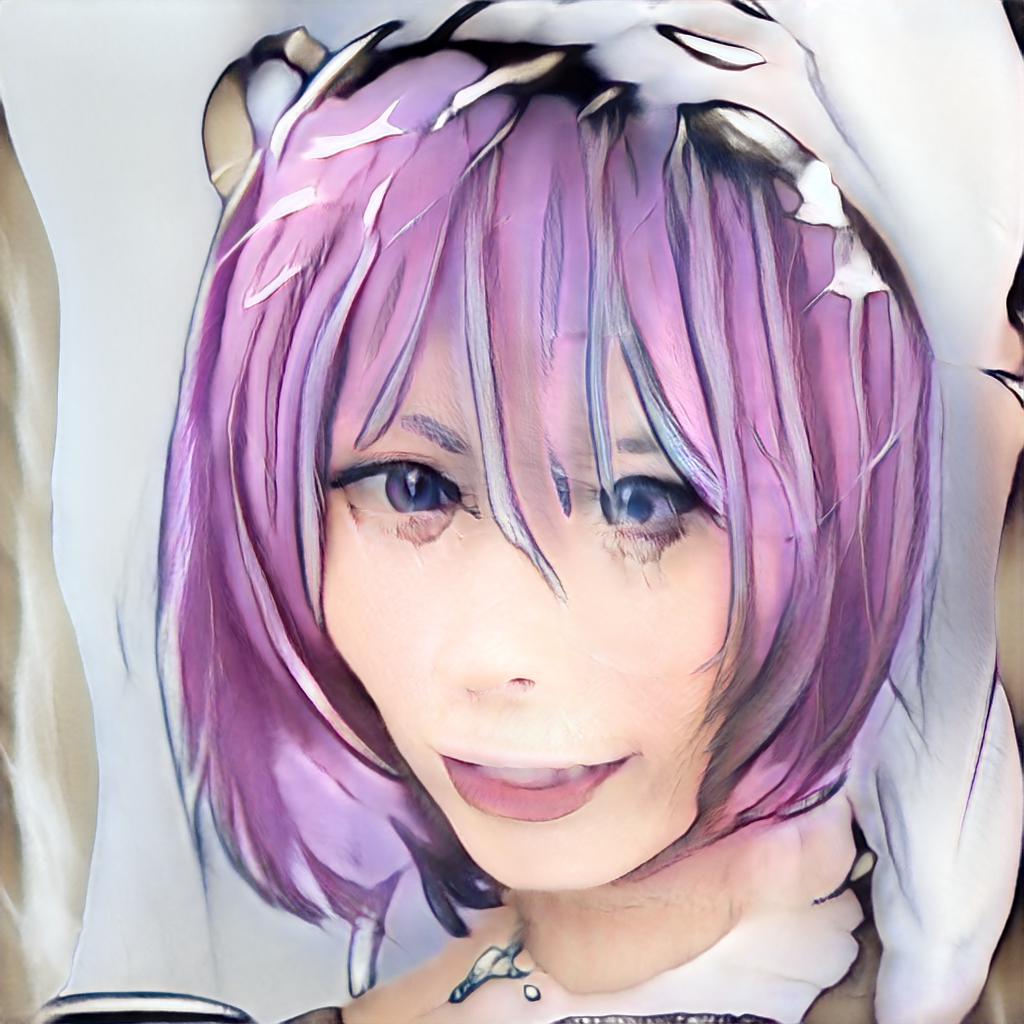}\hfill
}
\caption{
    \textbf{Far-domain adaptation} (Photo$\rightarrow$Art). 
    Comparing FSGAN with alternative GAN adaptation methods in the photo-to-art setting.
    %Comparison methods include TGAN \citep{wang2018eccv}, FD \citep{mo2020freeze}, SSGAN \citep{noguchi2019iccv}, with limited number iterations to prevent overfitting or GAN generation.
    \tb{(a)} FSGAN more effectively alters building layouts and adds landscape in the foreground to match the Van Gogh paintings, maintaining better spatial coherency. 
    \tb{(b)} FSGAN adopts features from the Portraits dataset (hats, beards, artistic backgrounds), while other methods primarily alter image textures.
    \tb{(c)} FSGAN transforms natural hair and facial features to imitate the anime target while retaining spatial consistency. Note the occurrence of pink hair in our generated images, which does not exist in the few-shot target but is visually consistent.}
\label{fig:stylization}
\end{center}
\end{figure*}

\textbf{Datasets.}
We used FFHQ \citep{karras2019style} and LSUN Churches \citep{yu2015lsun} pretrained checkpoints from StyleGAN2 \citep{karras2019corr}, and transferred to few-shot single-ID CelebA (30 or 31 images) \citep{liu2015celeba}, Portraits (5-100 images) \citep{lee2018drit}, Anime ID ``Rem'' (25 images) \footnote{\hyperlink{https://www.gwern.net/Danbooru2019}{https://www.gwern.net/Danbooru2019}}, and Van Gogh landscapes (25 images) \citep{zhu2017cyclegan}. 
We evaluate FID against a large test set (10K for CelebA) following the evaluation method of ~\citet{wang2018eccv}. 
We also evaluate face quality index~\citep{hernandez2019faceqnet} and image sharpness~\citep{kumar2012sharpness} for face domain adaptation, using 1000 images from each method generated using identical seeds.
Full few-shot target sets are shown in Figures \ref{fig:personalization} \& \ref{fig:stylization}, and we will make all few-shot sets available online.
% TODO put datasets online.

\subsection{Near-domain adaptation}
\label{sec:near}
We first show a \emph{near domain} transfer setting (adapting FFHQ to single-ID CelebA datatset~\citep{liu2015celeba}). 
As both source and target domains contain faces, the pretrained model has useful features for the transfer domain.
% an easier adaptation setting of FFHQ to single-ID CelebA, so that the pretrained checkpoint contains maximally useful features for the transfer domain.
% Compared with FFHQ, CelebA contains images  a wider face crop and centers on the eye. 
%
\figref{personalization} shows that existing GAN adaptation methods produce artifacts around the eyes/chin and low overall structural consistency.
In contrast, our method generates more natural face images with characteristics similar to the training samples (e.g., the head size, position of the faces).
% reduces head size, centers the eyes, and adjusts hair.
%
Comparing Figure \ref{fig:personalization} and Table \ref{tab:personaliation} shows that the FID correlates poorly with qualitative evaluation for this setting. 
In light of this, we report additional metrics of face quality~\citep{hernandez2019faceqnet} and sharpness~\citep{kumar2012sharpness}. 
On these metrics, our method achieves competitive performance across adaptation settings.

\subsection{Far-domain adaptation}
\label{sec:far}
We show \emph{far-domain} 25-shot transfer, where we define ``far" as differing significantly in the distribution of image features such as textures, proportions, and semantics.
1) \textit{LSUN Churches$\rightarrow$ Van Gogh paintings}: The two domains differ in the foreground, building shapes, and textural styles.
% the target domain contains structurally distorted, diverse landscapes and different building types, and differ in color and texture distributions.
2) \textit{FFHQ$\rightarrow$Art portraits}: The main differences between the two domains are low-level styles and facial features.
3) \textit{FFHQ$\rightarrow$Anime Rem ID}: A challenging setting with exaggerated facial proportions and lack of texture details.
\figref{stylization} shows visual comparisons with three state-of-the-art methods. 
We find that the proposed FSGAN can adapt the model to produce more dramatic changes to match the target distributions in terms of semantics, proportions, and textures while maintaining image quality.

\subsection{N-shot Settings}
\label{sec:nshot}

We test the sensitivity of both FSGAN (ours) and FreezeD~\citep{mo2020freeze} to differing n-shot settings and show the results in Figure \ref{fig:nshot}.
We find that FSGAN is more robust to n-shot setting compared to FreezeD.
To show this better, we compare two variations of FreezeD.
The first FreezeD variant (FD) is limited in timesteps (20K images / 16K on 5-shot) to match FSGAN and the results reported in Figures \ref{fig:personalization} \& \ref{fig:stylization}.
Limiting timesteps prevents degradation that occurs at later iterations in the few-shot settings. 
However, the time-limited FD produces low quality and limited adaptation of textures and semantic features.
The second FreezeD variant (FD-FT) is trained for longer (60K images) to demonstrate (1) degradation in fewer n-shot and (2) improvements in quality/adaptation in higher n-shot as seen in \citep{mo2020freeze}.
In contrast, our method (FSGAN) effectively transfers semantic features while preserving quality across all n-shot settings tested in Figure \ref{fig:nshot}.
We note variance across n-shot settings for all methods as the data distribution changes.

\section{Conclusions}
\label{sec:conclusions}
We presented Few-Shot GAN, a simple yet effective method for adapting a pre-trained GAN based model to a new target domain where the number of training images is scarce.
Our core idea lies in factorizing the weights of convolutional/fully-connected layers in a pretrained model using SVD to identify a semantically meaningful parameter space for adaptation. 
Our strategy preserves the capability of generating diverse and realistic samples while provides the flexibility for adapting the model to a target domain with few examples. 
We demonstrate the effectiveness of the proposed method with close-domain and far-domain adaptation experiments and across various n-shot settings. 
We show favorable results compared with existing data-efficient GAN adaptation methods.

\newpage
\bibliographystyle{iclr2020}
{\small
\bibliography{main}
}

\end{document}